\documentclass[twoside,11pt]{article}

\usepackage[preprint]{dmlr2e}

\usepackage{booktabs}
\usepackage{amsmath}
\usepackage{amsfonts}
\usepackage{amssymb}
\usepackage{array}
\usepackage{multirow}
\usepackage{adjustbox}
\usepackage{enumitem}
\usepackage{mathtools}
\usepackage{lastpage}
\usepackage{xspace}
\usepackage{tabularx}
\usepackage{makecell}
\usepackage{tikz}
\usetikzlibrary{arrows.meta,positioning,calc,fit}

\newcommand{\bench}{\textsc{MIRA-Math}\xspace}
\newcommand{\releaseurl}{\url{https://github.com/cedar-lau/mira-math}} 
\newcommand{\dataurl}{\url{https://huggingface.co/datasets/samersaabjr/MIRA-MATH/}}
\newcommand{\jsonl}{\texttt{JSONL}\xspace}

\newcommand{\budget}{B}
\newcolumntype{Y}{>{\centering\arraybackslash}X}

\dmlrheading{1}{2026}{1--\pageref{LastPage}}{Submitted 2026}{Under review}{mira-math}{Al Bateh and Saab}
\ShortHeadings{MIRA-Math}{Al Bateh and Saab}
\firstpageno{1}

\title{\bench: A Benchmark for Minimal Information Requesting and Mathematical Reasoning}

\author{%
\name Charbel Al Bateh \email charbel.albateh@lau.edu \\
\addr Department of Electrical and Computer Engineering\\
Lebanese American University\\
Byblos, Lebanon
\AND
\name Samer Saab Jr. \email samer.saabjr@lau.edu.lb \\
\addr Department of Electrical and Computer Engineering\\
Lebanese American University\\
Byblos, Lebanon
}

\editor{}

\begin{document}
\maketitle

\begin{abstract}
Mathematical reasoning benchmarks typically provide all facts needed to solve each problem, while interactive benchmarks often mix reasoning with tools, retrieval, and long-horizon dialogue. We introduce \bench, a benchmark for a narrower diagnostic capability: solving mathematical problems whose full latent state has a unique answer, but whose solver-facing view is missing exactly one necessary atomic fact. The solver must request the missing information in natural language under a strict budget and then integrate the returned fact into an exact final answer. A fixed constrained LLM responder sees only the dataset-provided atomic fact and must either \textsc{offer} the quoted fact when the request matches it, or \textsc{decline} otherwise. Thus, instance generation, typed hint specifications, validation, and final-answer verification are deterministic, while request metrics are measured under a fixed LLM-mediated responder channel. \bench contains 2{,}310 generated instances from 22 typed mathematical families spanning algebra, probability, linear systems, discrete structures, signal processing, Markov chains, circuits, interpolation, and numerical boundary-value problems. Experiments across frontier and small models show that request success and final-answer accuracy are separable: models may ask for the right fact yet fail the downstream computation, or fail before obtaining the canonical hint. We release generators, verifiers, prompts, run metadata, and dataset documentation to support reproducible evaluation of minimal information requesting in mathematical reasoning.
\end{abstract}


\section{Introduction}

Large language models are increasingly evaluated on mathematical reasoning, tool use, and interactive problem solving. Yet many high-profile math benchmarks assess only the final answer under full information \citep{cobbe2021training,hendrycks2021math,balunovic2025matharena}, while broad interactive benchmarks mix multiple sources of difficulty, including tool selection, retrieval, web navigation, external APIs, and long-context management \citep{mialon2023gaia,liu2023agentbench,qin2023toolllm}. These settings are valuable, but they make it difficult to isolate a basic bottleneck that appears whenever a model does not initially possess all relevant information: can the model recognize the missing fact, ask for that fact precisely, and use the answer correctly?

\bench is designed to isolate this capability. Each benchmark instance is generated from a complete mathematical state with a unique answer, but the solver model receives only a private view that is deliberately insufficient, missing exactly one necessary atomic fact. A fixed information-holder model receives only the missing fact and is constrained by a structured-output protocol: if the solver's request semantically matches the information it holds, it returns that fact; otherwise, it returns a declination stating that it does not have the requested information. The information holder does not solve, explain, negotiate, or volunteer extra hints. The benchmark therefore measures minimal information requesting under a controlled responder channel, not open-ended multi-agent collaboration.

This framing is intentionally data-centric. Following the emphasis on public benchmark infrastructure and responsible dataset design in recent data-centric machine-learning work \citep{oala2024dmlr,orr2024building}, \bench includes generator code, exact verifiers, canonical hint specifications, prompt templates, run metadata, and dataset documentation. The benchmark is closest to diagnostic benchmark datasets that expose hidden structure or evaluation gaps through carefully controlled data construction \citep{matsubara2024rethinking,zhang2024labelbench,zhang2024openood,telyatnikov2025topobench}, but its target capability is different: information acquisition under partial mathematical observability.

\paragraph{Contributions.} This paper makes three contributions:

\begin{itemize}
    \item We define a reproducible protocol for minimal information requesting in mathematical reasoning, with typed atomic hints, a fixed constrained LLM responder using structured \textsc{offer}/\textsc{decline} outputs, deterministic dataset validation, and exact answer checking. 
    \item We report baseline evaluations across multiple solver models and prompt regimes, over a benchmark of 2,310 instances showing that first-request success, request hit rate, and final accuracy capture distinct failure modes. 
    \item We provide dataset documentation, intended-use guidance, and a maintenance plan.
\end{itemize}

\section{Related Work}
\label{sec:related-work}

\bench is positioned at the intersection of mathematical reasoning benchmarks,
clarification and information-acquisition benchmarks, active reasoning under incomplete
information, and data-centric benchmark design. Table~\ref{tab:positioning} summarizes
the closest literature streams and the distinction between \bench and prior work.

\begin{table}[t] 
\centering 
\caption{Positioning of \bench relative to nearby benchmark streams.} 
\label{tab:positioning} 
\scriptsize 
\setlength{\tabcolsep}{3pt} 
\renewcommand{\arraystretch}{0.92} 
\begin{adjustbox}{max width=\textwidth} 
\begin{tabular}{p{0.25\textwidth}p{0.34\textwidth}p{0.35\textwidth}} 
\toprule 
\textbf{Stream} & \textbf{Typical focus} & \textbf{How \bench differs} \\ 
\midrule 
Full-information math and diagnostic benchmarks \citep{cobbe2021training,hendrycks2021math,balunovic2025matharena,matsubara2024rethinking} & Solve fully specified problems, often with exact final-answer scoring or controlled generators. & The solver's private view is intentionally incomplete; success requires requesting and then using a missing atomic fact. \\ 
Clarification and incomplete-information reasoning \citep{min-etal-2020-ambigqa,gan2024clarqllm,li2025questbench,zhou2025arbench,huang2025criticmath} & Ask useful questions or detect missing information in underspecified tasks. & \bench uses typed mathematical families, a fixed constrained information holder, and exact verification of final mathematical integration. \\ 
Agent and tool-use benchmarks \citep{mialon2023gaia,liu2023agentbench,qin2023toolllm,zhou2023webarena} & Evaluate web navigation, tools, retrieval, APIs, and long-horizon workflows. & \bench removes tools and environment management to isolate minimal mathematical information requesting. \\ 
Multi-agent communication \citep{foerster2016learning,sukhbaatar2016learning,pmlr-v97-das19a,pmlr-v119-wang20i,wu2023autogen,li2023camel,chen2023agentverse} & Study coordination and message passing under partial observability. & The responder is a fixed constrained information holder, not a strategic teammate, so failures are more localized to request precision, responder matching, or mathematical integration. \\ 
Data-centric benchmark design \citep{oala2024dmlr,orr2024building,gebru2021datasheets,mitchell2019modelcards} & Emphasize public artifacts, documentation, intended use, and reproducible evaluation. & \bench releases generators, verifiers, typed hint schemas, responder specifications, prompts, run metadata, and documentation. \\ 
\bottomrule 
\end{tabular} 
\end{adjustbox} 
\end{table}

\subsection{Final-answer mathematical reasoning is not enough}

Mathematical reasoning benchmarks such as GSM8K and MATH have been central for
measuring whether language models can produce correct final answers from complete
problem statements \citep{cobbe2021training,hendrycks2021math}. More recent evaluations
increase difficulty, reduce contamination, or track competition-level performance
\citep{balunovic2025matharena}. These benchmarks are important, but they usually assume
that the model receives all information required to solve the problem. As a result, they do
not isolate the ability to notice that a problem is underdetermined, identify the missing
fact, ask for it, and then use it correctly.

\bench targets this missing capability directly. It should therefore not be read as a
replacement for full-information mathematical benchmarks. Instead, it complements them
by turning information acquisition into an explicitly measured step. A model that is strong
on ordinary math benchmarks may still fail in \bench if it cannot identify the missing
atomic fact or if it asks for an irrelevant quantity.

\subsection{Controlled diagnostic benchmarks}

A second relevant stream uses controlled generation and exact evaluation to reveal hidden
model weaknesses. Symbolic and scientific benchmark work has shown the value of
generator-controlled tasks, exact verifiers, and diagnostic decompositions
\citep{matsubara2024rethinking,somasekharan2026cfdllmbench}. Data-centric benchmark
papers similarly emphasize that the dataset, generator, evaluation protocol, and
documentation are part of the scientific contribution
\citep{zhang2024labelbench,zhang2024openood,telyatnikov2025topobench}.

\bench follows this diagnostic tradition but uses controlled generation for a different
purpose. Each instance is generated from a latent complete mathematical state that fixes a
unique answer, while the solver's private view omits a necessary atomic fact. This
global-well-posed/local-underdetermined construction lets us evaluate not only whether the
model eventually gets the answer right, but also whether it requested the correct missing
information along the way.

\subsection{Clarification and information acquisition}

Clarification benchmarks study whether models ask useful questions when instructions,
dialogue states, or user requests are ambiguous or underspecified
\citep{min-etal-2020-ambigqa,gan2024clarqllm,zhang2025futureconversation,zhao2026askbench}.
This line of work is closely related in motivation in that practical assistants should not always
guess when required information is absent. However, many clarification settings involve
subjective ambiguity, natural human preferences, or dialogue-level uncertainty. These
properties are valuable for realism but make it difficult to assign exact credit for a
particular request.

\bench reduces this ambiguity by construction. The full latent state has a unique target answer, and the missing information is represented by typed atomic hints specified by the generator. The solver may still ask in free-form natural language, but the response channel is constrained. A fixed information-holder LLM sees only its private constraints and must choose between a structured \textsc{offer} and a structured \textsc{decline}. The accepted/declined events are therefore logged in an auditable schema, while the semantic matching decision itself remains LLM-mediated. The deterministic components are the generator invariants, canonical hint specification, instance validation, and final-answer verification.

\subsection{Active and missing-information reasoning benchmarks}

The closest recent work studies reasoning under incomplete information. QuestBench
formalizes underspecified reasoning as missing variable assignments and evaluates whether
models can identify the minimal necessary question in logic, planning, and math-derived
tasks \citep{li2025questbench}. AR-Bench broadens the setting to active reasoning, where
models must ask questions to acquire missing evidence in interactive environments
\citep{zhou2025arbench}. CRITIC-math focuses directly on incomplete mathematical
problems and evaluates whether large reasoning models proactively ask for information
rather than hallucinating an answer \citep{huang2025criticmath}.

\bench is closest to this family of benchmarks, but it makes a different design choice. QuestBench emphasizes selecting the right clarification from a finite set of options; \bench requires open-ended natural-language requests that must be precise enough to be accepted by a fixed constrained information-holder LLM. AR-Bench emphasizes broader active reasoning in interactive scenarios; \bench restricts the environment to a single controlled mathematical information channel. CRITIC-math studies whether models detect incompleteness in mathematical problems; \bench additionally evaluates whether the model identifies the exact missing atomic slot and integrates the returned value into an exact solution. These differences make \bench a narrower but more controlled diagnostic for minimal mathematical information requesting.

\subsection{Agent, tool-use, and communication benchmarks}

Broad agent benchmarks evaluate capabilities such as tool selection, API use, retrieval, web navigation, and long-horizon planning  \citep{mialon2023gaia,liu2023agentbench,qin2023toolllm,zhou2023webarena}. LLM multi-agent frameworks and multi-agent reinforcement-learning work further study how agents coordinate and communicate under partial observability \citep{foerster2016learning,sukhbaatar2016learning,pmlr-v97-das19a,pmlr-v119-wang20i,wu2023autogen,li2023camel,chen2023agentverse}. These settings are useful, but they conflate information acquisition with many other sources of difficulty.

\bench intentionally avoids this conflation. The second role is not an autonomous collaborator and does not solve, negotiate, explain, or strategically communicate. It is a fixed LLM-mediated information holder over dataset-provided atomic hints: the holder sees only its private constraints, must quote an explicit constraint when offering information, and otherwise returns a structured declination. This design makes \bench a benchmark of request precision and mathematical integration, not a benchmark of general multi-agent collaboration.

\subsection{Data-centric benchmark design}

Finally, \bench is a data-centric benchmark contribution. Work on dataset documentation,
model cards, and responsible benchmark design emphasizes the need to specify intended
uses, limitations, quality-control procedures, and maintenance plans
\citep{gebru2021datasheets,mitchell2019modelcards,orr2024building,oala2024dmlr}.
Following this view, the \bench release is intended to include generated \jsonl files,
family generators, exact verifiers, canonical hint specifications, prompt templates, run
metadata, raw logs, and dataset documentation. The benchmark's contribution is therefore
not only a collection of problems, but a reproducible evaluation protocol for a specific
capability: asking for the minimal missing mathematical fact needed to solve an
underdetermined problem.

\section{Benchmark Design}
\label{sec:design}

\subsection{Task Formalization}

A \bench instance is generated from a latent complete state $z$. The full state determines a unique target answer $y = f(z)$. The solver receives a private view $v_A = \phi_A(z)$ that omits one necessary atomic fact. The hint source stores that atomic hint $h \in H(z)$, each with a typed identifier such as a missing coefficient, boundary value, congruence, transition probability, or interpolation point. The benchmark enforces two conditions:
\begin{align}
\text{global well-posedness:} && \left|\{y: y \text{ is consistent with } v_A \cup H(z)\}\right| &= 1, \\
\text{local underdetermination:} && \left|\{y: y \text{ is consistent with } v_A\}\right| &> 1.
\end{align}

\subsection{Interaction Protocol}

At evaluation time the solver is shown its private view, the target question, and the request budget $\budget$. At each turn the solver may either issue a request or provide a final answer. Requests are free-form natural language. A fixed information-holder model receives its own private view containing the missing atomic fact and must respond using a structured schema. If the request semantically matches the fact it holds, the responder returns that fact; otherwise, it returns a declination and the solver may try again until the request budget is exhausted.

For example, in a recurrence instance the solver may see the recurrence rule and $a(0)$ but not $a(1)$. A request such as ``What is the missing value of $a(1)$?'' maps to the canonical initial-condition hint and receives the stored value. A vague request such as ``Can you help me solve the recurrence?'' is declined because it does not identify an allowed atomic hint. In a variable-slot family, such as a linear system with one missing coefficient, the request must identify the missing coefficient slot rather than merely ask for ``the missing information.''


\begin{figure}[t]
    \centering
    \includegraphics[width=\textwidth]{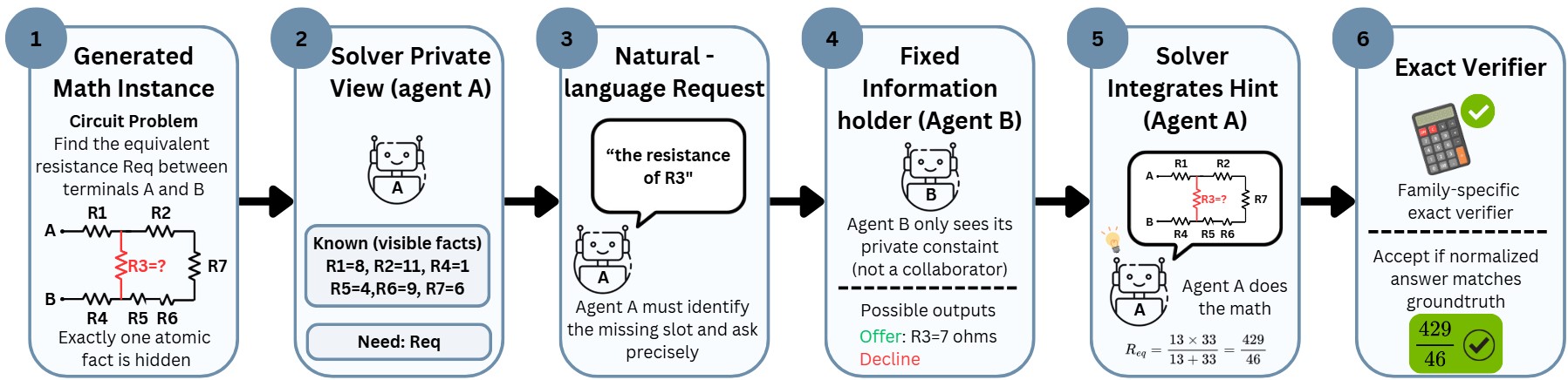}
    \caption{
Overview of the \bench protocol, illustrated with a successful
\texttt{circuit\_missing\_resistance} trace. Agent A receives a private view
with one atomic fact hidden, asks for the missing slot in natural language,
and receives either a structured \textsc{offer} or \textsc{decline} from a
fixed constrained information-holder that sees only its private constraint.
After receiving the hint, Agent A must still compute the final answer and pass
the family-specific exact verifier. 
}
    \label{fig:mira_protocol}
\end{figure}

This constrained responder channel is a central design choice. In an unconstrained two-model dialogue, the responder could solve part of the problem, volunteer extra information, negotiate, or provide explanations. \bench removes these behaviors by giving the information holder only its private constraints and requiring a structured offer-or-decline response. The resulting request metrics should be interpreted as measuring the solver's information-requesting behavior under a controlled LLM responder, not as a measure of general multi-agent teamwork or a purely deterministic request-matching oracle.

\paragraph{Responder implementation.} In the released reference runner, the information holder is implemented as a fixed LLM responder with a flat structured-output schema. Its only valid outputs are \textsc{offer}, with \texttt{has\_exact\_match=true}, a quoted private constraint, and a hint string; or \textsc{decline}, with a fixed declination message. If the structured output cannot be parsed, the runner attempts conservative raw-output extraction and otherwise records a declination. This design prevents the responder from solving the task, volunteering extra information, or giving partial clues, but it does not make semantic matching itself deterministic. Consequently, request metrics should be interpreted as solver performance under this fixed constrained responder channel. Exact mathematical correctness is still evaluated by deterministic family-specific verifiers.

\subsection{Family Types}

The 22 families are partitioned into two types. \textbf{Type A} families have a fixed hint slot: the missing information is structurally determined by the family. For example, a Bayes-rule instance always withholds the prior, and a recurrence instance always withholds one initial condition. \textbf{Type B} families have a variable hint slot: the missing position changes by instance, such as a missing transition probability, boundary value, coefficient, interpolation point, resistor, or grid cell. Type B therefore requires the solver to locate which slot is missing before phrasing the request.

This design supports diagnosis of two separate subskills. Type A primarily tests whether the model knows what kind of fact is needed and can compute after receiving it. Type B additionally tests whether the model can inspect its local view, identify the absent slot, and request that slot precisely.

\subsection{Metrics}

We report the following metrics.
\begin{itemize}[leftmargin=*] 
    \item \textbf{Accuracy}: fraction of instances whose final answer normalizes to the ground truth under type-aware comparison. 
    \item \textbf{Request hit rate}: mean per-instance ratio of accepted requests to fixed-responder messages, $\text{offers} / (\text{offers} + \text{declines})$.
    \item \textbf{First-request success}: fraction of instances in which the solver's first request receives a structured offer from the fixed responder. 
    \item \textbf{Average requests}: mean number of requests the solver issues per instance. 
    \item \textbf{Average declinations}: mean number of declinations from the fixed responder per instance. 
    \item \textbf{Average hints}: mean number of accepted offers the solver receives per instance. 
    \item \textbf{Average tokens}: approximate transcript cost, computed as mean transcript word-count × 1.3 per instance and summed across the solver and responder messages.
\end{itemize}

\paragraph{Trace-level decomposition.} 
To separate information-acquisition failures from downstream mathematical failures, we also classify each transcript using three binary indicators. Let $O_i$ indicate that instance $i$ received at least one structured \textsc{offer}; let $C_i$ indicate that an offered hint matches the instance's canonical atomic-hint specification under an offline audit of the quoted constraint and machine-readable hint fields; and let $F_i$ indicate that the final answer is accepted by the exact verifier. We report: 
\[ p_{\mathrm{no\text{-}canonical}}=\frac{1}{N}\sum_i (1-C_i), \qquad p_{\mathrm{integration\text{-}fail}}=\frac{1}{N}\sum_i C_i(1-F_i), \qquad p_{\mathrm{success}}=\frac{1}{N}\sum_i F_i. \] 

The first quantity captures failures to acquire the canonical missing fact under the fixed responder channel. The second captures cases where the missing fact was acquired but the solver did not integrate it into a correct mathematical answer. The third is the usual final-answer accuracy. We separately log raw offers $O_i$ so that any rare noncanonical offer can be distinguished from a genuine solver request success.

Accuracy measures mathematical completion, while request metrics measure information acquisition under the fixed responder channel. Their separation is important: a model may fail to elicit the canonical missing fact, may elicit it but still fail the mathematics, or may solve correctly after receiving it. The trace-level decomposition above makes these cases explicit rather than relying only on aggregate accuracy and request hit rate.

\section{Dataset Construction}
\label{sec:dataset}

\subsection{Current Release}

The current \bench release is the \emph{20/50 typed} dataset: 20 instances per difficulty level for each Type A family and 50 instances per difficulty level for each Type B family. This yields 660 Type A instances, 1,650 Type B instances, and 2,310 instances in total. Each instance stores the family name, type, difficulty, random seed, solver prompt, target question, typed atomic-hint specification, canonical hint values, exact answer, and verifier metadata.

The 22 families (described in Appendix~\ref{app:families}) cover Bayesian inference, Chinese remainder reconstruction, coordinate geometry, triangle path-sum inversion, linear systems with a separator variable, rank-one matrix completion, moment recovery, phase retrieval, piecewise functions, rank-deficient linear systems with a shared variable, recurrences, birth-death chains, series-parallel resistor networks, deconvolution, discrete tomography, eigenvector-constrained matrix questions, discrete Laplace equations, linear systems with a missing coefficient, Markov chains, polynomial interpolation, portfolio variance, and hidden Markov models.

\subsection{Generator Invariants and Quality Control}

Every generator enforces family-specific nondegeneracy conditions. Examples include rejecting singular linear systems, parallel lines, nonunique phase-retrieval instances, ambiguous tomography targets without a sufficient hint cell, and rational answers whose denominators exceed family-specific thresholds. All numeric answers are represented exactly as integers, tuples, or reduced rational values. Verifiers compare normalized exact values rather than surface strings.

For families where local underdetermination can be checked algebraically, the generator uses rank, uniqueness, or constraint-satisfaction tests. For finite discrete families, the generator can enumerate consistent completions. These checks are part of the dataset artifact, so users can regenerate the release, create larger splits, or audit any instance from its seed.

\subsection{Data Format, Access, and Maintenance}

The public release includes generated \jsonl files, family generator code, exact verifiers, prompt templates, responder specifications, model-run scripts, and raw logs. The repository and dataset pages are:
\begin{center}
Code and documentation: \releaseurl\\
Generated data: \dataurl
\end{center}
The generated benchmark data are released under CC BY 4.0, and the generator, verifier, runner, and evaluation code are released under the MIT License.

The maintenance plan is to version releases by generator version, seed set, and prompt protocol. Future releases will preserve the current 20/50 typed split as a regression suite while adding larger held-out seed sets and optional multi-hint variants. Each release will include checksums, schema documentation, and a changelog documenting changes to generators, prompts, matching rules, and verifiers.

\section{Experimental Protocol}
\label{sec:experiments}


We evaluate solver models in two prompt regimes: zero-shot (ZS) and four-shot (4S). The four-shot prompt provides examples of valid minimal requests and final-answer formatting without reusing test instances. In all reported experiments, the information-holder model is fixed to \texttt{gpt-4o-mini}; the solver model is the variable under evaluation. The responder receives only its private constraints and is prompted to return a flat structured output with one of two types: \textsc{offer}, when it judges the request to semantically match one of its private constraints, or \textsc{decline}, otherwise. A valid offer must include \texttt{has\_exact\_match=true}, the exact quoted private constraint, and the extracted hint string. A declination uses a fixed message and must not reveal what other information the responder holds. Thus the responder is fixed and constrained, but it is not a deterministic matching oracle.

The released runner stores the full transcript for each instance, including Agent A requests, Agent B offer/decline decisions, quoted constraints, extracted hints, parsing errors when present, final answers, and exact-verifier outcomes. This makes it possible to recompute aggregate metrics and the trace-level decomposition in Section~\ref{sec:results} without rerunning model calls, provided the raw transcripts are retained.

The main results report six solver models. Exact decoding settings, model identifiers, timestamps, prompts, and raw transcripts are included with the release artifacts so that future users can reproduce the runs or update the baselines as model APIs and open-weight checkpoints change.

\section{Results}
\label{sec:results}

\begin{figure}[t] 
    \centering 
    \includegraphics[width=\textwidth]{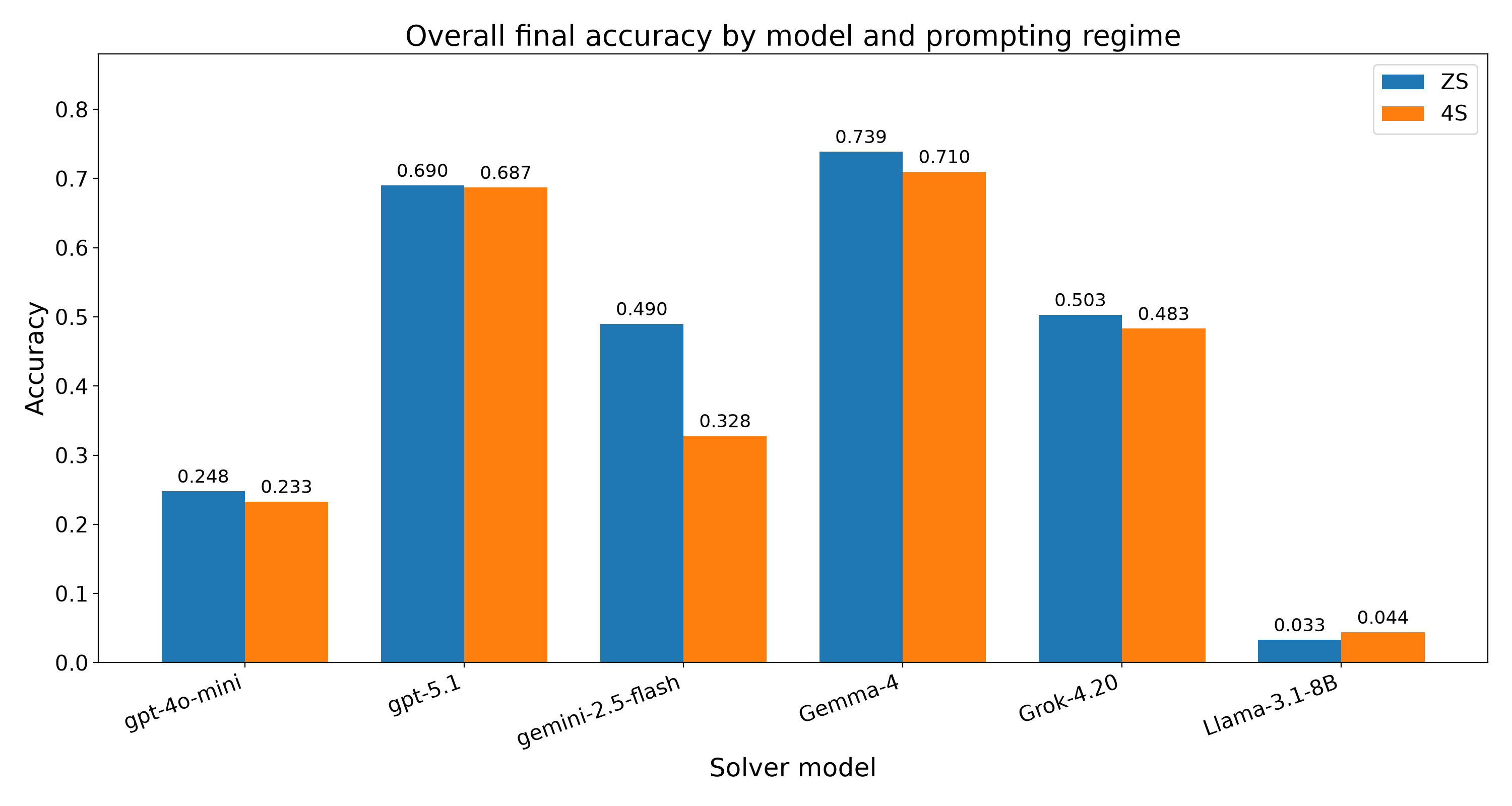} 
    \caption{ Overall final-answer accuracy by solver model and prompting regime. Bars are grouped by solver model, with separate zero-shot (ZS) and four-shot (4S) results. The full numeric counterpart, including request hit rate, first-request success, average requests, declinations, and token estimates, is retained in Appendix~\ref{app:main_numeric_tables}.} 
    \label{fig:overall_accuracy_bars} 
\end{figure}

\paragraph{Requesting and solving are distinct.} 
Figure~\ref{fig:overall_accuracy_bars} summarizes final-answer accuracy across solver models and prompting regimes, while Appendix~\ref{app:main_numeric_tables} retains the full numeric request metrics. The aggregate results show that high request performance does not necessarily imply high final-answer accuracy. For example, in the zero-shot setting, \texttt{gemini-2.5-flash} has the highest request hit rate and first-request success among the completed models, but its final accuracy remains below \texttt{gpt-5.1} and \texttt{Gemma-4}. Conversely, \texttt{Gemma-4} obtains the highest zero-shot final accuracy while not achieving the highest request metrics. This validates the benchmark's central diagnostic split: asking for the right fact and using it correctly are separable skills. 

\paragraph{Trace snapshots make the diagnostic split concrete.} To make the aggregate metrics easier to interpret, Appendix~\ref{app:trace_snapshots} gives short, offline-audited transcript snapshots for three representative outcomes: no canonical hint acquired, canonical hint acquired but final answer wrong, and full request--resolve--solve success. These snapshots are not used to estimate prevalence and should not be read as representative samples. Instead, they show what the logged metrics mean at the transcript level. For example, a no-hint trace shows a solver cycling through plausible but wrong missing rates until the request budget is exhausted; a hint-but-wrong trace shows the solver receiving the prior needed for Bayes' rule but still submitting an incorrect posterior; and a full-success trace shows the solver requesting the missing resistor value and then computing the equivalent resistance exactly. Table~\ref{tab:trace_snapshot_index} summarizes the examples and their diagnostic role.


\begin{table}[t]
\centering
\small
\setlength{\tabcolsep}{3pt}
\renewcommand{\arraystretch}{1.18}
\caption{Illustrative transcript snapshots used to interpret the aggregate metrics. These are hand-audited examples, not a frequency estimate. Full transcripts and raw logs are released with the benchmark artifacts.}
\label{tab:trace_snapshot_index}
\begin{tabularx}{\linewidth}{@{}
    l
    >{\raggedright\arraybackslash}p{0.20\linewidth}
    >{\raggedright\arraybackslash}p{0.19\linewidth}
    >{\raggedright\arraybackslash}p{0.12\linewidth}
    >{\raggedright\arraybackslash}X
@{}}
\toprule
Tag & Outcome class & Example family & Instance & Diagnostic role \\
\midrule
T1
& No canonical hint acquired
& \shortstack[l]{\texttt{birth\_death}\\\texttt{missing\_rate}} 
& \texttt{bir-000768}
& The solver recognizes that a rate is missing but repeatedly asks for the wrong rate, so the responder only declines and no final answer is submitted.
\\

T2
& Canonical hint acquired, final answer wrong
& \shortstack[l]{\texttt{bayes\_missing}\\\texttt{prior}} 
& \texttt{bay-000044}
& The solver asks for and receives the prior $P(H)$, but the downstream Bayes calculation is wrong. This is an integration failure rather than a request failure.
\\

T3
& Canonical hint acquired, final answer correct
& \shortstack[l]{\texttt{circuit\_missing}\\\texttt{resistance}} 
& \texttt{cir-000924}
& The solver identifies the missing resistor, receives it, and computes the exact equivalent resistance. This illustrates the intended successful loop.
\\
\bottomrule
\end{tabularx}
\end{table}

\paragraph{Request acquisition versus mathematical integration.} The transcript decomposition confirms that final accuracy should not be read as a pure requesting score. In the reference protocol, an instance can fail because the solver never obtains the canonical atomic hint, or because it obtains the hint but performs the subsequent computation incorrectly. We therefore report the disjoint trace-level categories in Table~\ref{tab:trace_decomposition}: no canonical hint acquired, canonical hint acquired but final answer wrong, and final answer correct. This decomposition is especially important for families such as Chinese-remainder reconstruction, phase retrieval, polynomial interpolation, and portfolio variance, where models can often obtain the relevant hint but still fail exact arithmetic, symbolic reconstruction, or answer normalization.

\begin{table}[t]
\centering
\small
\setlength{\tabcolsep}{3pt}
\renewcommand{\arraystretch}{1.15}
\caption{Trace-level decomposition of failures. ``No canonical hint'' means that the transcript did not contain an accepted offer matching the instance's canonical atomic-hint specification under the offline hint audit. ``Hint but wrong'' means that the canonical hint was acquired but the exact final-answer verifier rejected the solver's answer. ``Correct'' is the standard final-answer accuracy. All values are computed from raw transcripts; no additional model calls are required to reproduce this decomposition from the released logs.}
\label{tab:trace_decomposition}
\begin{tabularx}{\linewidth}{@{}>{\raggedright\arraybackslash}p{0.25\linewidth}cYYYY@{}}
\toprule
\makecell[l]{Agent A\\model}
& Format
& \makecell[c]{No\\canonical\\hint}
& \makecell[c]{Hint\\but\\wrong}
& Correct
& \makecell[c]{Acc.\\given\\canonical\\hint} \\
\midrule
gpt-4o-mini        & ZS & 0.213 & 0.540 & 0.248 & 0.314 \\
gpt-5.1            & ZS & 0.200 & 0.111 & 0.689 & 0.861 \\
gemini-2.5-flash   & ZS & 0.126 & 0.385 & 0.490 & 0.560 \\
Gemma-4            & ZS & 0.187 & 0.075 & 0.739 & 0.908 \\
Grok-4.20          & ZS & 0.231 & 0.266 & 0.503 & 0.654 \\
Llama-3.1-8B-Inst  & ZS & 0.393 & 0.574 & 0.033 & 0.054 \\
\midrule
gpt-4o-mini        & 4S & 0.210 & 0.557 & 0.233 & 0.295 \\
gpt-5.1            & 4S & 0.197 & 0.116 & 0.687 & 0.855 \\
gemini-2.5-flash   & 4S & 0.196 & 0.477 & 0.328 & 0.407 \\
Gemma-4            & 4S & 0.216 & 0.074 & 0.710 & 0.906 \\
Grok-4.20          & 4S & 0.250 & 0.267 & 0.483 & 0.644 \\
Llama-3.1-8B-Inst  & 4S & 0.375 & 0.581 & 0.044 & 0.070 \\
\bottomrule
\end{tabularx}
\end{table}

\begin{figure}[t] 
    \centering 
    \includegraphics[width=\textwidth]{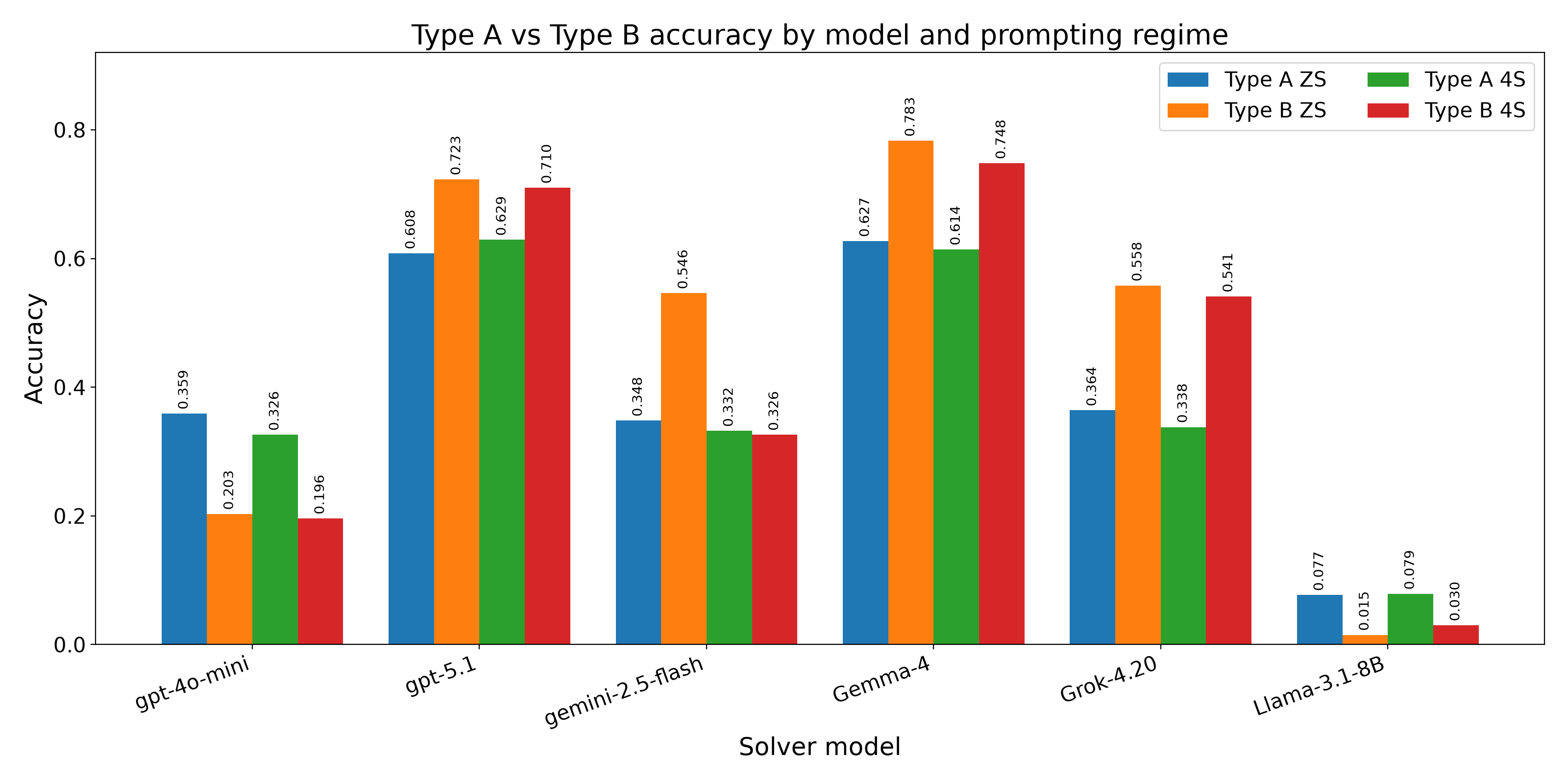} 
    \caption{ Type~A versus Type~B final-answer accuracy by solver model and prompting regime. Type~A families have a fixed missing-information slot, whereas Type~B families require instance-level localization of the missing slot. The full numeric tables for Type~A/Type~B accuracy and request quality are retained in Appendix~\ref{app:main_numeric_tables}. } 
    \label{fig:type_accuracy_bars} 
\end{figure} 

\paragraph{Variable-slot families are not uniformly harder.} Figure~\ref{fig:type_accuracy_bars} shows that Type~B families, despite requiring the solver to identify which slot is absent, are not uniformly harder in final-answer accuracy. Strong models often score higher on Type~B than on Type~A. This is not a contradiction: Type~A contains several families where the request is easy but the downstream computation is brittle, such as exact Chinese-remainder reconstruction, path-sum inversion, moment recovery, and phase retrieval. The Type~A/Type~B distinction should therefore be interpreted as a missing-slot diagnostic, not as a total difficulty ranking. Appendix~\ref{app:main_numeric_tables} gives the corresponding request-quality metrics, including hit rate, first-request success, and average number of requests by family type.

\begin{figure}[t] 
    \centering 
    \includegraphics[width=\textwidth]{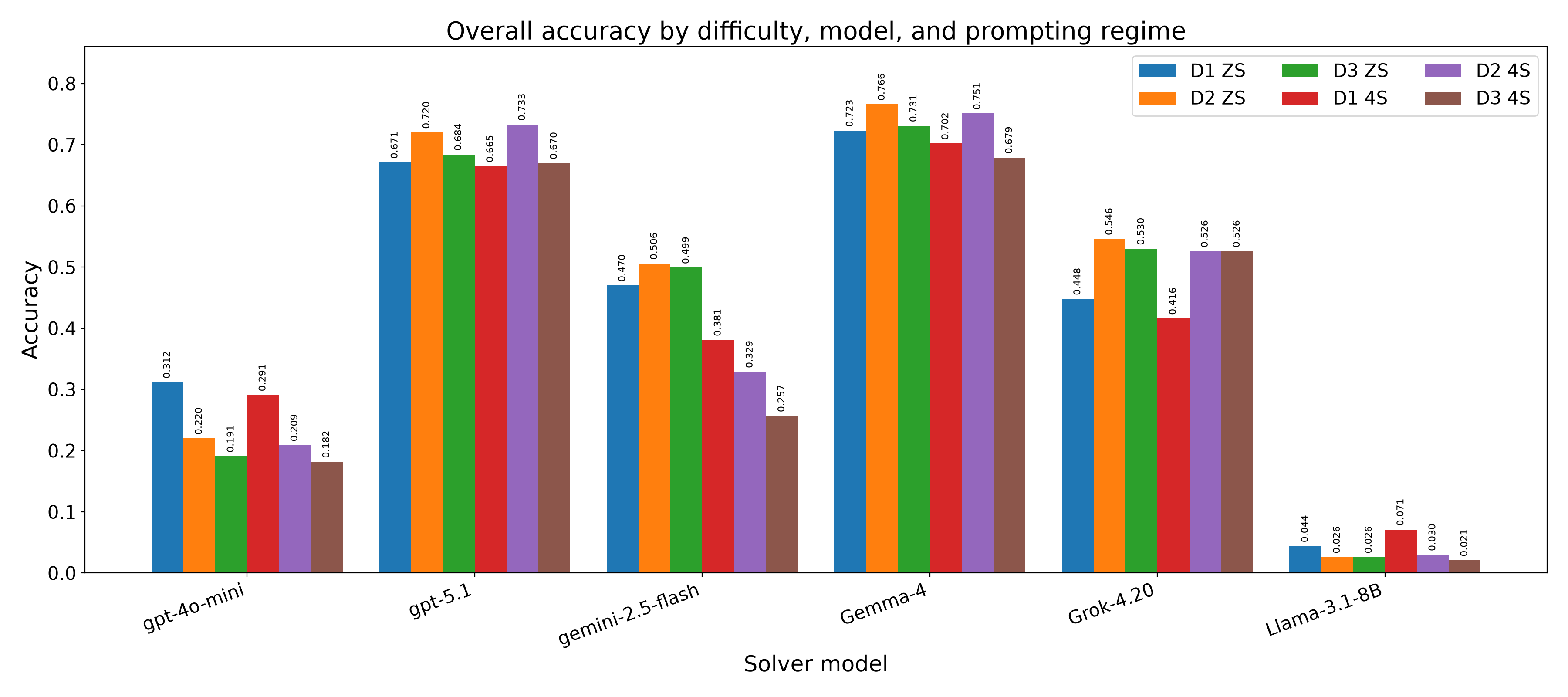} 
    \caption{ Overall final-answer accuracy by difficulty level, solver model, and prompting regime. Bars are grouped by solver model and split by difficulty level D1--D3 for zero-shot and four-shot prompting. The full numeric difficulty tables, including average request counts and the difficulty-by-family-type accuracy split, are retained in Appendix~\ref{app:main_numeric_tables}. } 
    \label{fig:difficulty_accuracy_bars} 
\end{figure} 

\paragraph{Difficulty interacts with family structure.} Figure~\ref{fig:difficulty_accuracy_bars} shows that smaller models can degrade sharply with difficulty, while stronger models are less monotonic. This reflects the benchmark design: difficulty scaling may increase coefficient ranges, dimensions, denominators, or target-index depth depending on the family. Per-family and per-type reporting remain essential, because a single aggregate score can hide whether a model fails at requesting, exact arithmetic, linear algebra, combinatorial search, or answer normalization. The full difficulty-by-type breakdown is provided in Appendix~\ref{app:main_numeric_tables}.

\paragraph{Four-shot prompting is not uniformly beneficial.} Figures~\ref{fig:overall_accuracy_bars}--\ref{fig:difficulty_accuracy_bars} show that the tested four-shot prompt does not uniformly improve final-answer accuracy. The Type~A/Type~B split helps interpret this pattern. Demonstrations can help when the missing slot is structurally stable: in Type~A families, the solver can often reuse a demonstrated pattern such as asking for a prior, initial condition, threshold, or missing equation. By contrast, Type~B families require instance-level slot localization: the solver must inspect the current problem and identify which coefficient, transition, boundary value, grid cell, resistor, interpolation point, or emission entry is absent. A small four-shot prompt necessarily covers only a few request archetypes, and may therefore induce exemplar anchoring: the solver may imitate the demonstrated request forms instead of adapting to the specific missing slot in the current instance. We treat this as a prompt-sensitivity hypothesis rather than a causal claim. The main result is that demonstrations of the interaction protocol do not automatically improve minimal-information reasoning; in some families they improve first-request compliance while leaving mathematical integration unchanged or worse. Appendix~\ref{app:trace_snapshots} includes an illustrative prompt-anchoring snapshot (T4), and Appendix~\ref{app:main_numeric_tables} gives the full per-family four-shot deltas.

Taken together, these results show that \bench should be read less as a single leaderboard and more as a diagnostic decomposition. The same aggregate accuracy can arise from different mechanisms: failure to localize the missing slot, successful acquisition followed by incorrect computation, or correct end-to-end request and solution. The trace-level logs, per-family metrics, and Type~A/Type~B split are therefore intended to be used jointly when comparing models or prompt strategies.

\section{Failure Modes and Diagnostic Use}
\label{sec:diagnostics}

The benchmark exposes at least two recurring failure modes. \emph{Request-acquisition failures} occur when the solver does not obtain the canonical missing fact under the fixed responder channel. These can happen because the solver asks for a wrong slot, asks too vaguely, or repeatedly probes quantities that are already present in its private view. \emph{Integration failures} occur when the solver receives the canonical hint but still computes, formats, or normalizes the final answer incorrectly. 

Several families make this distinction visible. In fixed-slot families such as \texttt{bayes\_missing} \texttt{\_prior}, the request can be easy while exact arithmetic remains brittle. In variable-slot families such as \texttt{birth\_death\_missing\_rate}, the solver must first localize the absent rate before it can apply a standard computation. Other families, such as \texttt{circuit\_missing\_resistance}, show the intended successful pattern: identify the missing slot, request it precisely, and integrate the returned value into the exact verifier-compatible answer. 

Appendix~\ref{app:trace_snapshots} gives compact transcript snapshots for these cases. The examples are illustrative rather than representative; aggregate claims are based on the tables above. Their purpose is to show how the same final incorrect score can correspond to different underlying failures: no canonical hint acquired, canonical hint acquired but wrong downstream computation, or successful request and solution. This is why the release includes both exact verifiers and raw interaction logs.

\section{Intended Uses and Limitations}
\label{sec:limitations}

\paragraph{Intended uses.}
\bench is intended for evaluating models, prompts, and solver loops that must operate under partial mathematical information. Appropriate uses include measuring request precision, comparing communication-efficiency strategies, auditing whether a model can identify missing facts, and studying the relationship between clarification success and final mathematical accuracy.

\paragraph{Out-of-scope uses.}
The benchmark should not be used as evidence that a model can collaborate with another autonomous agent. The second role is a fixed, constrained information holder, not a teammate. The benchmark should also not be treated as a general measure of tool use, web navigation, social reasoning, or real-world decision-making.

\paragraph{Limitations.}
The tasks are synthetic and mathematically structured. This improves controllability and exact verification but limits realism. The current release uses one required atomic hint per instance; many real tasks require multiple facts, or uncertain evidence. The typed hint schema and constrained responder protocol is useful for reproducibility but may understate the difficulty of communicating with humans or open-ended systems. The responder channel is controlled but not deterministic: in the reported experiments, a fixed LLM information holder performs the semantic offer-or-decline decision under structured-output constraints. This means request metrics can still reflect occasional responder matching errors, although the final-answer verifier and dataset validity checks are deterministic. Finally, model results can become stale as APIs and open-weight checkpoints change; the release is designed to support re-running baselines rather than treating the reported ranking as permanent.

\section{Conclusion}

\bench introduces a controlled benchmark for minimal information requesting in mathematical reasoning. Each instance is generated from a complete latent mathematical state with a unique answer, but the solver receives only a locally underdetermined view and must request exactly the missing atomic fact needed to complete the solution. This construction turns information acquisition into an explicit, measurable step rather than leaving it implicit inside final-answer accuracy.

The results show that this step is not redundant with ordinary mathematical solving. Across the tested models, request hit rate, first-request success, and final-answer accuracy diverge substantially: some models obtain the canonical hint but fail exact downstream computation, while others fail before acquiring the resolving fact. The Type~A/Type~B split and transcript-level decomposition further show that fixed missing slots, variable missing slots, arithmetic integration, and answer normalization can fail independently.

The benchmark is intentionally narrow. It does not claim to measure open-ended collaboration, tool use, or human clarification behavior. Instead, it provides a reproducible diagnostic setting with typed atomic hints, a fixed constrained LLM responder, deterministic validation, exact family-specific verifiers, and released transcripts. We hope this makes \bench useful both as a baseline suite for future models and as an analysis tool for studying when language models ask for missing mathematical information rather than guessing.

\impact{\bench has positive potential as a controlled diagnostic for safer and more transparent use of language models in settings where systems should ask for missing information instead of guessing. The main risks are overclaiming and benchmark overfitting. The benchmark does not measure open-ended collaboration, human communication, or real-world decision-making, and we state these limitations explicitly. Since the data are synthetically generated mathematical problems, privacy and consent risks are minimal. We mitigate misuse by releasing generator code, verifiers, documentation, and intended-use guidance, and by recommending held-out seed sets for future evaluations.}

\acks{The authors are responsible for the dataset design, generated instances, and evaluation protocol.}

\bibliography{ref}

\section*{Dataset Certification} 
\bench is a synthetically generated benchmark. Each instance is produced from a recorded generator family, difficulty level, random seed, typed atomic-hint specification, exact answer, and verifier metadata. The release includes generated data, generator code, exact verifiers, prompt templates, responder specifications, model-run scripts, raw interaction logs, and documentation. The dataset contains mathematical problem text and synthetic numeric values rather than personal data. Intended uses and out-of-scope uses are described in Section~\ref{sec:limitations}.

\appendix

\section{Implementation and Reproducibility Details}
\label{app:implementation_repro}

\paragraph{Released artifacts.}
The release contains the family generators, instance validator, exact family-specific solvers, prompt templates, structured-output schemas, reference LangGraph runner, scoring code, generated JSONL files, run metadata, and raw transcripts. Each generated instance records the family name, difficulty, random seed, agent-private constraints, machine-readable canonical atomic-hint specification, global solution, answer type, and verifier metadata.

\paragraph{Instance validation.}
The validator checks three properties for each instance: global uniqueness of the full latent problem, local underdetermination of each agent's private view, and sufficiency of the canonical atomic hint when applied to the solver's view. These checks are family-specific and deterministic. The final-answer verifier also uses normalized exact values rather than surface-form string matching.

\paragraph{Responder protocol.}
The reference responder is a fixed information-holder LLM. It receives only its private constraints and Agent A's current request. It must emit a flat structured object with either \textsc{offer} or \textsc{decline}. An \textsc{offer} must set \texttt{has\_exact\_match=true}, quote the exact private constraint, and provide the extracted hint. A \textsc{decline} must not reveal alternative information. If structured parsing fails, the runner attempts conservative raw-output extraction; if that also fails, the response is recorded as a declination. Therefore, semantic matching is controlled and auditable but LLM-mediated.

\paragraph{Transcript scoring.}
For each transcript, the scorer records final-answer accuracy, request attempts, offer count, declination count, first-request success, requests before first offer, token estimate, and exact expected/actual normalized answers. We additionally compute the trace-level indicators used in Table~\ref{tab:trace_decomposition}: whether a canonical hint was acquired, whether the solver produced a correct final answer after acquiring it, and whether a noncanonical offer occurred.

\paragraph{Versioning.}
To make future reruns comparable, each release records the generator version, seed set, prompt-protocol version, responder model identifier, solver model identifier, decoding settings, code commit, dataset checksum, and transcript checksum. Model rankings should be interpreted as a snapshot under these settings because API-backed model behavior can change over time.

\paragraph{Exact run configuration.}
Tables~\ref{tab:repro_settings} and~\ref{tab:repro_models} record the concrete settings used for all reported runs so that the main reproducibility facts are available without consulting the release artifacts. All runs use greedy decoding (\texttt{temperature}~$=0$) for both agents, a single pass per instance, and a fixed request seed of $1234$; residual provider-side nondeterminism may remain where the backend does not honor the seed. Models are referenced by provider alias on the dates shown; the underlying API snapshots and open-weight checkpoints are not version-pinned and may change.

\begin{table}[t]
\centering
\small
\caption{Shared run configuration, identical across all reported solver models and both prompting regimes. The request budget scales with instance difficulty $d$ as $3d$. Datasets are released as JSONL with a fixed generation seed; each instance additionally stores its own generation seed.}
\label{tab:repro_settings}
\begin{tabular}{ll}
\toprule
Setting & Value \\
\midrule
Responder (Agent B) & \texttt{gpt-4o-mini} (OpenAI) \\
Decoding temperature & $0.0$ (Agent A and Agent B) \\
Max new tokens & unset (provider/model default) \\
Request budget (max attempts) & $3 / 6 / 9$ at difficulty $1 / 2 / 3$ \\
Runs per instance & $1$ (single pass) \\
Inference-time seed & $1234$ \\
Per-instance timeout & 3600~s (parallel runner) \\
Zero-shot dataset & \texttt{family\_types\_20\_50.jsonl} (2{,}310 inst.) \\
Four-shot dataset & \texttt{family\_types\_20\_50\_few\_shot.jsonl} (2{,}310 inst.) \\
Dataset generation seed & $1234$ \\
Protocol / method & \texttt{llm} (single-shot request, then final answer) \\
\bottomrule
\end{tabular}
\end{table}

\begin{table}[t]
\centering
\small
\caption{Per-solver exact model identifiers, serving backend, and run dates. The responder (Agent~B) is \texttt{gpt-4o-mini} in every configuration; for the \texttt{gpt-4o-mini} solver, Agents~A and~B are the same model. Date ranges span the initial run and any subsequent resumption waves for instances that timed out or errored.}
\label{tab:repro_models}
\begin{adjustbox}{max width=\textwidth}
\begin{tabular}{llll}
\toprule
Solver (Agent A) & Exact model ID & Backend & Run dates (2026) \\
\midrule
gpt-4o-mini       & \texttt{gpt-4o-mini}                  & OpenAI API            & Mar 13 \\
gpt-5.1           & \texttt{gpt-5.1}                      & OpenAI API            & Mar 23 \\
gemini-2.5-flash  & \texttt{gemini-2.5-flash}             & Google GenAI / OpenRouter & Apr 09--21 \\
Gemma-4           & \texttt{google/gemma-4-31b-it}        & OpenRouter            & Apr 17--19 \\
Grok-4.20         & \texttt{x-ai/grok-4.20}               & OpenRouter            & Apr 22--26 \\
Llama-3.1-8B-Inst & \texttt{Meta-Llama-3.1-8B-Instruct} (quantized, 16k ctx) & Ollama (local) & May 07--09 \\
\bottomrule
\end{tabular}
\end{adjustbox}
\end{table}

\section{Additional Results and Full Numeric Tables} 
\label{app:additional_results} 

\subsection{Full numeric tables for main-text figures} 
\label{app:main_numeric_tables} 

Tables~\ref{tab:full_overall_metrics}--\ref{tab:full_type_request_quality} provide the full numeric values underlying Figures~\ref{fig:overall_accuracy_bars}--\ref{fig:difficulty_accuracy_bars}. The main text visualizes the central accuracy patterns for readability, while these appendix tables retain the complete request-quality and token-usage metrics.

\begin{table}[t] 
\centering 
\small 
\caption{ Full overall performance metrics on \bench. This table is the numeric counterpart of Figure~\ref{fig:overall_accuracy_bars} and includes request hit rate, first-request success, average requests, declinations, and token estimates. } 
\label{tab:full_overall_metrics} 
\begin{adjustbox}{max width=\textwidth}
\begin{tabular}{ll cccccc}
\toprule
\textbf{Agent A Model} & \textbf{Type} & \textbf{Acc} & \textbf{Hit Rate} & \textbf{1st-Req OK} & \textbf{Avg Req} & \textbf{Avg Decl} & \textbf{Avg Tokens} \\
\midrule
\texttt{gpt-4o-mini}            &\multirow{6}{*}{ZS}& 0.248 & 0.596 & 0.493 & 2.55 & 1.77 & 507 \\
\texttt{gpt-5.1}                && 0.690 & 0.701 & 0.643 & 2.19 & 1.39 & 1065 \\
\texttt{gemini-2.5-flash}       && 0.490 & 0.793 & 0.735 & 1.69 & 0.82 & 533 \\
\texttt{Gemma-4}                && 0.739 & 0.720 & 0.652 & 2.12 & 1.30 & 1099 \\
\texttt{Grok-4.20}              && 0.503 & 0.661 & 0.587 & 2.34 & 1.57 & 1536 \\
\texttt{Llama-3.1-8B-Inst}      && 0.033 & 0.433 & 0.329 & 3.35 & 2.74 & 165 \\

\midrule
\texttt{gpt-4o-mini}            &\multirow{6}{*}{4S}& 0.233 & 0.604 & 0.510 & 2.60 & 1.81 & 494 \\
\texttt{gpt-5.1}                && 0.687 & 0.719 & 0.665 & 2.09 & 1.29 & 1013 \\
\texttt{gemini-2.5-flash}       && 0.328 & 0.722 & 0.670 & 2.06 & 1.25 & 672 \\
\texttt{Gemma-4}                && 0.710 & 0.703 & 0.648 & 2.26 & 1.47 & 1107 \\
\texttt{Grok-4.20}              && 0.483 & 0.634 & 0.555 & 2.46 & 1.71 & 1513 \\
\texttt{Llama-3.1-8B-Inst}      && 0.044 & 0.450 & 0.347 & 3.35 & 2.72 & 173 \\

\bottomrule
\end{tabular}
\end{adjustbox}
\end{table} 

\begin{table}[t] 
\centering 
\small 
\caption{ Full performance by family type. Type~A denotes fixed missing-information slots and Type~B denotes variable missing-information slots. This table is the numeric counterpart to Figure~\ref{fig:type_accuracy_bars}. } 
\label{tab:full_type_accuracy} 
\begin{adjustbox}{max width=\textwidth}
\begin{tabular}{ll cc cc cc}
\toprule
& & \multicolumn{2}{c}{\textbf{Accuracy}} & \multicolumn{2}{c}{\textbf{1st-Req OK}} & \multicolumn{2}{c}{\textbf{Avg Requests}} \\
\cmidrule(lr){3-4} \cmidrule(lr){5-6} \cmidrule(lr){7-8}
\textbf{Agent A Model} & \textbf{Type} & \textbf{Type A} & \textbf{Type B} & \textbf{Type A} & \textbf{Type B} & \textbf{Type A} & \textbf{Type B} \\
\midrule
\texttt{gpt-4o-mini}            &\multirow{6}{*}{ZS}& 0.359 & 0.203 & 0.542 & 0.473 & 2.36 & 2.63 \\
\texttt{gpt-5.1}                && 0.608 & 0.723 & 0.615 & 0.655 & 2.52 & 2.06 \\
\texttt{gemini-2.5-flash}       && 0.348 & 0.546 & 0.658 & 0.765 & 2.23 & 1.48 \\
\texttt{Gemma-4}                && 0.627 & 0.783 & 0.598 & 0.673 & 2.45 & 1.98 \\
\texttt{Grok-4.20}              && 0.364 & 0.558 & 0.591 & 0.586 & 2.30 & 2.36 \\
\texttt{Llama-3.1-8B-Inst}      && 0.077 & 0.015 & 0.364 & 0.315 & 3.02 & 3.48 \\
\midrule
\texttt{gpt-4o-mini}            &\multirow{6}{*}{4S}& 0.326 & 0.196 & 0.612 & 0.470 & 2.36 & 2.70 \\
\texttt{gpt-5.1}                && 0.629 & 0.710 & 0.661 & 0.667 & 2.32 & 2.00 \\
\texttt{gemini-2.5-flash}       && 0.332 & 0.326 & 0.676 & 0.668 & 2.23 & 1.99 \\
\texttt{Gemma-4}                && 0.614 & 0.748 & 0.550 & 0.687 & 2.62 & 2.11 \\
\texttt{Grok-4.20}              && 0.338 & 0.541 & 0.656 & 0.515 & 2.30 & 2.52 \\
\texttt{Llama-3.1-8B-Inst}      && 0.079 & 0.030 & 0.547 & 0.267 & 2.94 & 3.51 \\
\bottomrule
\end{tabular}
\end{adjustbox}
\end{table} 

\begin{table}[t] 
\centering 
\small 
\caption{ Full accuracy and average-request breakdown by difficulty level. This table is the numeric counterpart to Figure~\ref{fig:difficulty_accuracy_bars}. } 
\label{tab:full_difficulty_accuracy} 
\begin{adjustbox}{max width=\textwidth}
\begin{tabular}{ll ccc ccc}
\toprule
& & \multicolumn{3}{c}{\textbf{Accuracy}} & \multicolumn{3}{c}{\textbf{Avg Requests}} \\
\cmidrule(lr){3-5} \cmidrule(lr){6-8}
\textbf{Agent A Model} & \textbf{Format} & \textbf{D1} & \textbf{D2} & \textbf{D3} & \textbf{D1} & \textbf{D2} & \textbf{D3} \\
\midrule
\texttt{gpt-4o-mini}            &\multirow{6}{*}{ZS}& 0.312 & 0.220 & 0.191 & 1.90 & 2.56 & 3.39 \\
\texttt{gpt-5.1}                && 0.671 & 0.720 & 0.684 & 1.70 & 2.24 & 2.78 \\
\texttt{gemini-2.5-flash}       && 0.470 & 0.506 & 0.499 & 1.56 & 1.70 & 1.86 \\
\texttt{Gemma-4}                && 0.723 & 0.766 & 0.731 & 1.62 & 2.16 & 2.72 \\
\texttt{Grok-4.20}              && 0.448 & 0.546 & 0.530 & 1.76 & 2.34 & 3.10 \\
\texttt{Llama-3.1-8B-Inst}      && 0.044 & 0.026 & 0.026 & 2.16 & 3.47 & 4.76 \\
\midrule
\texttt{gpt-4o-mini}            &\multirow{6}{*}{4S}& 0.291 & 0.209 & 0.182 & 1.81 & 2.69 & 3.54 \\
\texttt{gpt-5.1}                && 0.665 & 0.733 & 0.670 & 1.68 & 2.15 & 2.56 \\
\texttt{gemini-2.5-flash}       && 0.381 & 0.329 & 0.257 & 1.68 & 2.09 & 2.52 \\
\texttt{Gemma-4}                && 0.702 & 0.751 & 0.679 & 1.63 & 2.31 & 3.01 \\
\texttt{Grok-4.20}              && 0.416 & 0.526 & 0.526 & 1.81 & 2.47 & 3.28 \\
\texttt{Llama-3.1-8B-Inst}      && 0.071 & 0.030 & 0.021 & 2.09 & 3.47 & 4.88 \\
\bottomrule
\end{tabular}
\end{adjustbox} 
\end{table}

\begin{table}[t] 
\centering 
\small 
\caption{ Full accuracy split by difficulty level and family type. This table supports the discussion of how difficulty interacts with fixed-slot and variable-slot family structure. } 
\label{tab:full_difficulty_type_accuracy} 
\begin{adjustbox}{max width=\textwidth}
\begin{tabular}{ll ccc ccc}
\toprule
& & \multicolumn{3}{c}{\textbf{Type A}} & \multicolumn{3}{c}{\textbf{Type B}} \\
\cmidrule(lr){3-5} \cmidrule(lr){6-8}
\textbf{Agent A Model} & \textbf{Format} & \textbf{D1} & \textbf{D2} & \textbf{D3} & \textbf{D1} & \textbf{D2} & \textbf{D3} \\
\midrule
\texttt{gpt-4o-mini}            &\multirow{6}{*}{ZS}& 0.338 & 0.380 & 0.365 & 0.302 & 0.156 & 0.122 \\
\texttt{gpt-5.1}                && 0.512 & 0.650 & 0.690 & 0.735 & 0.748 & 0.678 \\
\texttt{gemini-2.5-flash}       && 0.277 & 0.390 & 0.400 & 0.548 & 0.552 & 0.538 \\
\texttt{Gemma-4}                && 0.519 & 0.690 & 0.705 & 0.805 & 0.796 & 0.742 \\
\texttt{Grok-4.20}              && 0.308 & 0.400 & 0.400 & 0.505 & 0.604 & 0.582 \\
\texttt{Llama-3.1-8B-Inst}      && 0.069 & 0.080 & 0.085 & 0.034 & 0.004 & 0.002 \\
\midrule
\texttt{gpt-4o-mini}            &\multirow{6}{*}{4S}& 0.285 & 0.330 & 0.375 & 0.294 & 0.160 & 0.104 \\
\texttt{gpt-5.1}                && 0.512 & 0.695 & 0.715 & 0.726 & 0.748 & 0.652 \\
\texttt{gemini-2.5-flash}       && 0.300 & 0.355 & 0.350 & 0.414 & 0.318 & 0.220 \\
\texttt{Gemma-4}                && 0.481 & 0.685 & 0.715 & 0.791 & 0.778 & 0.664 \\
\texttt{Grok-4.20}              && 0.292 & 0.365 & 0.370 & 0.466 & 0.590 & 0.588 \\
\texttt{Llama-3.1-8B-Inst}      && 0.081 & 0.085 & 0.070 & 0.068 & 0.008 & 0.002 \\
\bottomrule
\end{tabular}
\end{adjustbox}
\end{table}

\begin{table}[t] 
\centering 
\small 
\caption{ Full request-quality breakdown by family type, including hit rate, first-request success, and average request count. This table supports the claim that request acquisition and final-answer accuracy capture different failure modes. } 
\label{tab:full_type_request_quality} 
\begin{adjustbox}{max width=\textwidth}
\begin{tabular}{ll cc cc cc}
\toprule
& & \multicolumn{2}{c}{\textbf{Hit Rate}} & \multicolumn{2}{c}{\textbf{1st-Req OK}} & \multicolumn{2}{c}{\textbf{Avg Requests}} \\
\cmidrule(lr){3-4} \cmidrule(lr){5-6} \cmidrule(lr){7-8}
\textbf{Agent A Model} & \textbf{Format} & \textbf{Type A} & \textbf{Type B} & \textbf{Type A} & \textbf{Type B} & \textbf{Type A} & \textbf{Type B} \\
\midrule
\texttt{gpt-4o-mini}            &\multirow{6}{*}{ZS}& 0.653 & 0.573 & 0.542 & 0.473 & 2.36 & 2.63 \\
\texttt{gpt-5.1}                && 0.659 & 0.715 & 0.615 & 0.655 & 2.52 & 2.06 \\
\texttt{gemini-2.5-flash}       && 0.704 & 0.829 & 0.658 & 0.765 & 2.23 & 1.48 \\
\texttt{Gemma-4}                && 0.659 & 0.745 & 0.598 & 0.673 & 2.45 & 1.98 \\
\texttt{Grok-4.20}              && 0.681 & 0.653 & 0.591 & 0.586 & 2.30 & 2.36 \\
\texttt{Llama-3.1-8B-Inst}      && 0.498 & 0.407 & 0.364 & 0.315 & 3.02 & 3.48 \\
\midrule
\texttt{gpt-4o-mini}            &\multirow{6}{*}{4S}& 0.671 & 0.577 & 0.612 & 0.470 & 2.36 & 2.70 \\
\texttt{gpt-5.1}                && 0.702 & 0.726 & 0.661 & 0.667 & 2.32 & 2.00 \\
\texttt{gemini-2.5-flash}       && 0.723 & 0.722 & 0.676 & 0.668 & 2.23 & 1.99 \\
\texttt{Gemma-4}                && 0.618 & 0.737 & 0.550 & 0.687 & 2.62 & 2.11 \\
\texttt{Grok-4.20}              && 0.704 & 0.606 & 0.656 & 0.515 & 2.30 & 2.52 \\
\texttt{Llama-3.1-8B-Inst}      && 0.587 & 0.396 & 0.547 & 0.267 & 2.94 & 3.51 \\
\bottomrule
\end{tabular}
\end{adjustbox}
\end{table}

\subsection{Per-family and prompt-effect results} 

The following tables provide additional per-family result summaries and four-shot effect analyses for each reported solver model and prompt regime.



\begin{table}[t]
\centering
\caption{Per-family accuracy on the 20/50 typed dataset (\texttt{gpt-4o-mini}, 2310 instances). Families sorted by Type then name. $n$ = instances per family.}
\label{tab:per-family-acc-4omini}
\begin{adjustbox}{max width=\textwidth}
\begin{tabular}{llr cc cc}
\toprule
& & & \multicolumn{2}{c}{\textbf{zero-shot}} & \multicolumn{2}{c}{\textbf{four-shot}} \\
\cmidrule(lr){4-5} \cmidrule(lr){6-7}
\textbf{Type} & \textbf{Family} & $n$ & \textbf{Acc} & \textbf{1st-Req} & \textbf{Acc} & \textbf{1st-Req} \\
\midrule
\multirow{11}{*}{A}
 & \texttt{bayes\_missing\_prior}       &  60 & 0.750 & 1.000 & 0.800 & 1.000 \\
 & \texttt{crt\_reconstruction}         &  60 & 0.000 & 1.000 & 0.000 & 1.000 \\
 & \texttt{geometry\_coordinates}       &  60 & 0.033 & 0.500 & 0.067 & 0.567 \\
 & \texttt{graph\_path\_sums}           &  60 & 0.000 & 0.000 & 0.000 & 0.000 \\
 & \texttt{linear\_system\_separator}   &  60 & 0.933 & 0.283 & 0.783 & 0.350 \\
 & \texttt{matrix\_completion}          &  60 & 0.167 & 0.967 & 0.133 & 1.000 \\
 & \texttt{moment\_problem}             &  60 & 0.000 & 0.000 & 0.000 & 0.000 \\
 & \texttt{phase\_retrieval}            &  60 & 0.000 & 0.650 & 0.033 & 0.967 \\
 & \texttt{piecewise\_missing\_thresh.} &  60 & 0.950 & 1.000 & 0.800 & 1.000 \\
 & \texttt{rankdef\_linear\_shared}     &  60 & 0.133 & 0.350 & 0.033 & 0.300 \\
 & \texttt{recurrence\_missing\_init}   &  60 & 0.983 & 1.000 & 0.933 & 1.000 \\
\midrule
\multirow{11}{*}{B}
 & \texttt{birth\_death\_missing\_rate}  & 150 & 0.227 & 0.387 & 0.280 & 0.400 \\
 & \texttt{circuit\_missing\_resistance} & 150 & 0.560 & 1.000 & 0.540 & 0.993 \\
 & \texttt{deconvolution}               & 150 & 0.127 & 0.127 & 0.127 & 0.080 \\
 & \texttt{discrete\_tomography}        & 150 & 0.080 & 0.013 & 0.147 & 0.113 \\
 & \texttt{eigenvector\_missing\_entry}  & 150 & 0.593 & 0.547 & 0.593 & 0.300 \\
 & \texttt{laplace\_grid}               & 150 & 0.000 & 0.020 & 0.000 & 0.007 \\
 & \texttt{linear\_sys\_missing\_coeff}  & 150 & 0.093 & 0.007 & 0.140 & 0.047 \\
 & \texttt{markov\_missing\_transition}  & 150 & 0.153 & 0.207 & 0.100 & 0.193 \\
 & \texttt{poly\_interpolation}         & 150 & 0.081 & 0.819 & 0.027 & 0.780 \\
 & \texttt{portfolio\_var\_missing\_c.}  & 150 & 0.153 & 0.960 & 0.141 & 0.832 \\
 & \texttt{steady\_state\_missing\_em.}  & 150 & 0.167 & 1.000 & 0.060 & 0.993 \\
\bottomrule
\end{tabular}
\end{adjustbox}
\end{table}


\begin{table}[p]
\centering
\caption{Full per-family metrics (\texttt{gpt-4o-mini}, zero-shot, 20/50 typed dataset).}
\label{tab:per-family-full-4omini}
\begin{adjustbox}{max width=\textwidth}
\begin{tabular}{llr ccccccc}
\toprule
\textbf{Type} & \textbf{Family} & $n$ & \textbf{Acc} & \textbf{Hit Rate} & \textbf{1st-Req} & \textbf{Avg Req} & \textbf{Avg Decl} & \textbf{Avg Hints} & \textbf{Avg Tok} \\
\midrule
\multirow{11}{*}{A}
 & \texttt{bayes\_missing\_prior}       &  60 & 0.750 & 1.000 & 1.000 & 1.00 & 0.00 & 1.00 & 345 \\
 & \texttt{crt\_reconstruction}         &  60 & 0.000 & 1.000 & 1.000 & 1.00 & 0.00 & 1.00 & 312 \\
 & \texttt{geometry\_coordinates}       &  60 & 0.033 & 0.597 & 0.500 & 2.17 & 1.18 & 0.78 & 465 \\
 & \texttt{graph\_path\_sums}           &  60 & 0.000 & 0.003 & 0.000 & 6.00 & 5.98 & 0.02 & 826 \\
 & \texttt{linear\_system\_separator}   &  60 & 0.933 & 0.421 & 0.283 & 2.72 & 1.72 & 1.00 & 607 \\
 & \texttt{matrix\_completion}          &  60 & 0.167 & 0.983 & 0.967 & 1.02 & 0.02 & 1.00 & 311 \\
 & \texttt{moment\_problem}             &  60 & 0.000 & 0.000 & 0.000 & 6.00 & 6.00 & 0.00 & 908 \\
 & \texttt{phase\_retrieval}            &  60 & 0.000 & 0.650 & 0.650 & 1.35 & 0.35 & 0.65 & 497 \\
 & \texttt{piecewise\_missing\_thresh.} &  60 & 0.950 & 1.000 & 1.000 & 1.00 & 0.00 & 1.00 & 342 \\
 & \texttt{rankdef\_linear\_shared}     &  60 & 0.133 & 0.419 & 0.350 & 2.68 & 1.78 & 0.62 & 555 \\
 & \texttt{recurrence\_missing\_init}   &  60 & 0.983 & 1.000 & 1.000 & 1.00 & 0.00 & 1.00 & 358 \\
\midrule
\multirow{11}{*}{B}
 & \texttt{birth\_death\_missing\_rate}  & 150 & 0.227 & 0.405 & 0.387 & 2.79 & 1.95 & 0.65 & 566 \\
 & \texttt{circuit\_missing\_resistance} & 150 & 0.560 & 1.000 & 1.000 & 1.00 & 0.00 & 1.00 & 489 \\
 & \texttt{deconvolution}               & 150 & 0.127 & 0.133 & 0.127 & 4.23 & 3.60 & 0.34 & 560 \\
 & \texttt{discrete\_tomography}        & 150 & 0.080 & 0.087 & 0.013 & 2.91 & 2.73 & 0.16 & 424 \\
 & \texttt{eigenvector\_missing\_entry}  & 150 & 0.593 & 0.580 & 0.547 & 2.25 & 1.33 & 0.77 & 555 \\
 & \texttt{laplace\_grid}               & 150 & 0.000 & 0.102 & 0.020 & 4.41 & 3.72 & 0.53 & 597 \\
 & \texttt{linear\_sys\_missing\_coeff}  & 150 & 0.093 & 0.036 & 0.007 & 4.32 & 3.83 & 0.36 & 688 \\
 & \texttt{markov\_missing\_transition}  & 150 & 0.153 & 0.359 & 0.207 & 3.04 & 2.21 & 0.63 & 559 \\
 & \texttt{poly\_interpolation}         & 149 & 0.081 & 0.784 & 0.819 & 1.96 & 0.97 & 0.61 & 389 \\
 & \texttt{portfolio\_var\_missing\_c.}  & 150 & 0.153 & 0.960 & 0.960 & 1.04 & 0.04 & 0.96 & 535 \\
 & \texttt{steady\_state\_missing\_em.}  & 150 & 0.167 & 1.000 & 1.000 & 1.00 & 0.00 & 1.00 & 548 \\
\bottomrule
\end{tabular}
\end{adjustbox}
\end{table}


\begin{table}[p]
\centering
\caption{Full per-family metrics (\texttt{gpt-4o-mini}, four-shot, 20/50 typed dataset).}
\label{tab:per-family-full-4omini-4s}
\begin{adjustbox}{max width=\textwidth}
\begin{tabular}{llr ccccccc}
\toprule
\textbf{Type} & \textbf{Family} & $n$ & \textbf{Acc} & \textbf{Hit Rate} & \textbf{1st-Req} & \textbf{Avg Req} & \textbf{Avg Decl} & \textbf{Avg Hints} & \textbf{Avg Tok} \\
\midrule
\multirow{11}{*}{A}
 & \texttt{bayes\_missing\_prior}       &  60 & 0.800 & 1.000 & 1.000 & 1.00 & 0.00 & 1.00 & 248 \\
 & \texttt{crt\_reconstruction}         &  60 & 0.000 & 1.000 & 1.000 & 1.00 & 0.00 & 1.00 & 271 \\
 & \texttt{geometry\_coordinates}       &  60 & 0.067 & 0.658 & 0.517 & 2.07 & 1.22 & 0.85 & 358 \\
 & \texttt{graph\_path\_sums}           &  60 & 0.000 & 0.010 & 0.000 & 5.97 & 5.93 & 0.03 & 747 \\
 & \texttt{linear\_system\_separator}   &  60 & 0.783 & 0.503 & 0.233 & 2.45 & 1.45 & 1.00 & 625 \\
 & \texttt{matrix\_completion}          &  60 & 0.133 & 1.000 & 1.000 & 1.00 & 0.00 & 1.00 & 258 \\
 & \texttt{moment\_problem}             &  60 & 0.000 & 0.000 & 0.000 & 6.00 & 6.00 & 0.00 & 725 \\
 & \texttt{phase\_retrieval}            &  60 & 0.033 & 0.983 & 0.967 & 1.03 & 0.03 & 1.00 & 307 \\
 & \texttt{piecewise\_missing\_thresh.} &  60 & 0.800 & 1.000 & 1.000 & 1.00 & 0.00 & 1.00 & 185 \\
 & \texttt{rankdef\_linear\_shared}     &  60 & 0.033 & 0.228 & 0.017 & 3.43 & 2.72 & 0.72 & 778 \\
 & \texttt{recurrence\_missing\_init}   &  60 & 0.933 & 1.000 & 1.000 & 1.00 & 0.00 & 1.00 & 341 \\
\midrule
\multirow{11}{*}{B}
 & \texttt{birth\_death\_missing\_rate}  & 150 & 0.280 & 0.544 & 0.353 & 2.81 & 1.89 & 0.91 & 580 \\
 & \texttt{circuit\_missing\_resistance} & 150 & 0.540 & 0.997 & 0.993 & 1.01 & 0.01 & 1.00 & 298 \\
 & \texttt{deconvolution}                & 150 & 0.127 & 0.206 & 0.093 & 4.54 & 4.01 & 0.53 & 736 \\
 & \texttt{discrete\_tomography}         & 150 & 0.147 & 0.190 & 0.107 & 2.69 & 2.38 & 0.31 & 486 \\
 & \texttt{eigenvector\_missing\_entry}  & 150 & 0.593 & 0.640 & 0.533 & 2.69 & 1.85 & 0.85 & 513 \\
 & \texttt{laplace\_grid}                & 150 & 0.000 & 0.176 & 0.013 & 4.87 & 4.20 & 0.67 & 908 \\
 & \texttt{linear\_sys\_missing\_coeff}  & 150 & 0.140 & 0.401 & 0.260 & 4.17 & 3.53 & 0.64 & 629 \\
 & \texttt{markov\_missing\_transition}  & 150 & 0.100 & 0.529 & 0.313 & 2.87 & 2.00 & 0.87 & 572 \\
 & \texttt{poly\_interpolation}          & 150 & 0.027 & 0.728 & 0.607 & 1.83 & 0.89 & 0.95 & 392 \\
 & \texttt{portfolio\_var\_missing\_c.}  & 150 & 0.140 & 0.939 & 0.900 & 1.16 & 0.17 & 0.99 & 314 \\
 & \texttt{steady\_state\_missing\_em.}  & 150 & 0.060 & 0.996 & 0.993 & 1.01 & 0.01 & 1.00 & 234 \\
\bottomrule
\end{tabular}
\end{adjustbox}
\end{table}


\begin{table}[p]
\centering
\caption{Four-shot effect: accuracy change per family (\texttt{gpt-4o-mini}, 20/50 typed dataset). $\Delta$ = four-shot accuracy $-$ baseline accuracy. Positive values indicate improvement.}
\label{tab:fourshot-effect-4omini}
\begin{adjustbox}{max width=\textwidth}
\begin{tabular}{llr ccc ccc}
\toprule
& & & \multicolumn{3}{c}{\textbf{Accuracy}} & \multicolumn{3}{c}{\textbf{1st-Request Success}} \\
\cmidrule(lr){4-6} \cmidrule(lr){7-9}
\textbf{Type} & \textbf{Family} & $n$ & \textbf{Base} & \textbf{4S} & $\Delta$ & \textbf{Base} & \textbf{4S} & $\Delta$ \\
\midrule
\multirow{11}{*}{A}
 & \texttt{bayes\_missing\_prior}       &  60 & 0.750 & 0.800 & \textbf{+0.050} & 1.000 & 1.000 &  0.000 \\
 & \texttt{crt\_reconstruction}         &  60 & 0.000 & 0.000 &  0.000 & 1.000 & 1.000 &  0.000 \\
 & \texttt{geometry\_coordinates}       &  60 & 0.033 & 0.067 & \textbf{+0.033} & 0.500 & 0.567 & +0.067 \\
 & \texttt{graph\_path\_sums}           &  60 & 0.000 & 0.000 &  0.000 & 0.000 & 0.000 &  0.000 \\
 & \texttt{linear\_system\_separator}   &  60 & 0.933 & 0.783 & $-$0.150 & 0.283 & 0.350 & +0.067 \\
 & \texttt{matrix\_completion}          &  60 & 0.167 & 0.133 & $-$0.033 & 0.967 & 1.000 & +0.033 \\
 & \texttt{moment\_problem}             &  60 & 0.000 & 0.000 &  0.000 & 0.000 & 0.000 &  0.000 \\
 & \texttt{phase\_retrieval}            &  60 & 0.000 & 0.033 & \textbf{+0.033} & 0.650 & 0.967 & \textbf{+0.317} \\
 & \texttt{piecewise\_missing\_thresh.} &  60 & 0.950 & 0.800 & $-$0.150 & 1.000 & 1.000 &  0.000 \\
 & \texttt{rankdef\_linear\_shared}     &  60 & 0.133 & 0.033 & $-$0.100 & 0.350 & 0.300 & $-$0.050 \\
 & \texttt{recurrence\_missing\_init}   &  60 & 0.983 & 0.933 & $-$0.050 & 1.000 & 1.000 &  0.000 \\
\cmidrule(lr){2-9}
 & \textit{Type A average}              & 660 & 0.359 & 0.326 & $-$0.033 & 0.542 & 0.612 & +0.070 \\
\midrule
\multirow{11}{*}{B}
 & \texttt{birth\_death\_missing\_rate}  & 150 & 0.227 & 0.280 & \textbf{+0.053} & 0.387 & 0.400 & +0.013 \\
 & \texttt{circuit\_missing\_resistance} & 150 & 0.560 & 0.540 & $-$0.020 & 1.000 & 0.993 & $-$0.007 \\
 & \texttt{deconvolution}               & 150 & 0.127 & 0.127 &  0.000 & 0.127 & 0.080 & $-$0.047 \\
 & \texttt{discrete\_tomography}        & 150 & 0.080 & 0.147 & \textbf{+0.067} & 0.013 & 0.113 & \textbf{+0.100} \\
 & \texttt{eigenvector\_missing\_entry}  & 150 & 0.593 & 0.593 &  0.000 & 0.547 & 0.300 & $-$0.247 \\
 & \texttt{laplace\_grid}               & 150 & 0.000 & 0.000 &  0.000 & 0.020 & 0.007 & $-$0.013 \\
 & \texttt{linear\_sys\_missing\_coeff}  & 150 & 0.093 & 0.140 & \textbf{+0.047} & 0.007 & 0.047 & +0.040 \\
 & \texttt{markov\_missing\_transition}  & 150 & 0.153 & 0.100 & $-$0.053 & 0.207 & 0.193 & $-$0.013 \\
 & \texttt{poly\_interpolation}         & 150 & 0.081 & 0.027 & $-$0.054 & 0.819 & 0.780 & $-$0.039 \\
 & \texttt{portfolio\_var\_missing\_c.}  & 150 & 0.153 & 0.141 & $-$0.013 & 0.960 & 0.832 & $-$0.128 \\
 & \texttt{steady\_state\_missing\_em.}  & 150 & 0.167 & 0.060 & $-$0.107 & 1.000 & 0.993 & $-$0.007 \\
\cmidrule(lr){2-9}
 & \textit{Type B average}             & 1650 & 0.203 & 0.196 & $-$0.007 & 0.473 & 0.470 & $-$0.003 \\
\midrule
 & \textbf{Overall}                    & 2310 & \textbf{0.248} & \textbf{0.233} & $-$0.015 & 0.493 & 0.510 & +0.017 \\
\bottomrule
\end{tabular}
\end{adjustbox}
\end{table}
 


\begin{table}[p]
\centering
\caption{Per-family accuracy on the 20/50 typed dataset (\texttt{gpt-5.1}). Families sorted by Type then name. $n$ = instances per family.}
\label{tab:per-family-acc-51}
\begin{adjustbox}{max width=\textwidth}
\begin{tabular}{llr cc cc}
\toprule
& & & \multicolumn{2}{c}{\textbf{zero-shot}} & \multicolumn{2}{c}{\textbf{four-shot}} \\
\cmidrule(lr){4-5} \cmidrule(lr){6-7}
\textbf{Type} & \textbf{Family} & $n$ & \textbf{Acc} & \textbf{1st-Req} & \textbf{Acc} & \textbf{1st-Req} \\
\midrule
\multirow{11}{*}{A}
 & \texttt{bayes\_missing\_prior}       &  60 & 1.000 & 1.000 & 1.000 & 1.000 \\
 & \texttt{crt\_reconstruction}         &  60 & 0.000 & 1.000 & 0.000 & 1.000 \\
 & \texttt{geometry\_coordinates}       &  60 & 0.117 & 0.067 & 0.317 & 0.267 \\
 & \texttt{graph\_path\_sums}           &  60 & 0.017 & 0.017 & 0.050 & 0.083 \\
 & \texttt{linear\_system\_separator}   &  60 & 1.000 & 0.133 & 1.000 & 1.000 \\
 & \texttt{matrix\_completion}          &  60 & 0.700 & 0.667 & 0.717 & 0.683 \\
 & \texttt{moment\_problem}             &  60 & 0.000 & 0.000 & 0.000 & 0.017 \\
 & \texttt{phase\_retrieval}            &  60 & 0.933 & 1.000 & 0.983 & 1.000 \\
 & \texttt{piecewise\_missing\_thresh.} &  60 & 1.000 & 1.000 & 1.000 & 1.000 \\
 & \texttt{rankdef\_linear\_shared}     &  60 & 0.917 & 0.883 & 0.850 & 0.217 \\
 & \texttt{recurrence\_missing\_init}   &  60 & 1.000 & 1.000 & 1.000 & 1.000 \\
\midrule
\multirow{11}{*}{B}
 & \texttt{birth\_death\_missing\_rate}  & 150 & 0.980 & 0.967 & 0.953 & 0.920 \\
 & \texttt{circuit\_missing\_resistance} & 150 & 1.000 & 1.000 & 0.980 & 1.000 \\
 & \texttt{deconvolution}               & 150 & 0.153 & 0.033 & 0.153 & 0.107 \\
 & \texttt{discrete\_tomography}        & 150 & 0.567 & 0.233 & 0.513 & 0.173 \\
 & \texttt{eigenvector\_missing\_entry}  & 150 & 0.933 & 0.853 & 0.960 & 0.927 \\
 & \texttt{laplace\_grid}               & 150 & 0.227 & 0.533 & 0.227 & 0.520 \\
 & \texttt{linear\_sys\_missing\_coeff}  & 150 & 0.693 & 0.520 & 0.573 & 0.427 \\
 & \texttt{markov\_missing\_transition}  & 150 & 0.833 & 0.640 & 0.847 & 0.647 \\
 & \texttt{poly\_interpolation}         & 150 & 0.807 & 0.467 & 0.847 & 0.660 \\
 & \texttt{portfolio\_var\_missing\_c.}  & 150 & 0.960 & 0.980 & 0.973 & 0.993 \\
 & \texttt{steady\_state\_missing\_em.}  & 150 & 0.800 & 0.973 & 0.787 & 0.967 \\
\bottomrule
\end{tabular}
\end{adjustbox}
\end{table}


\begin{table}[p]
\centering
\caption{Full per-family metrics (\texttt{gpt-5.1}, zero-shot, 20/50 typed dataset).}
\label{tab:per-family-full-51}
\begin{adjustbox}{max width=\textwidth}
\begin{tabular}{llr ccccccc}
\toprule
\textbf{Type} & \textbf{Family} & $n$ & \textbf{Acc} & \textbf{Hit Rate} & \textbf{1st-Req} & \textbf{Avg Req} & \textbf{Avg Decl} & \textbf{Avg Hints} & \textbf{Avg Tok} \\
\midrule
\multirow{11}{*}{A}
 & \texttt{bayes\_missing\_prior}       &  60 & 1.000 & 1.000 & 1.000 & 1.00 & 0.00 & 1.00 &  392 \\
 & \texttt{crt\_reconstruction}         &  60 & 0.000 & 1.000 & 1.000 & 1.00 & 0.00 & 1.00 &  791 \\
 & \texttt{geometry\_coordinates}       &  60 & 0.117 & 0.166 & 0.067 & 4.72 & 4.37 & 0.35 & 1161 \\
 & \texttt{graph\_path\_sums}           &  60 & 0.017 & 0.031 & 0.017 & 5.75 & 5.70 & 0.05 & 1211 \\
 & \texttt{linear\_system\_separator}   &  60 & 1.000 & 0.425 & 0.133 & 2.72 & 1.72 & 1.00 & 1035 \\
 & \texttt{matrix\_completion}          &  60 & 0.700 & 0.718 & 0.667 & 2.22 & 1.37 & 0.85 &  774 \\
 & \texttt{moment\_problem}             &  60 & 0.000 & 0.002 & 0.000 & 5.97 & 5.95 & 0.02 & 1154 \\
 & \texttt{phase\_retrieval}            &  60 & 0.933 & 1.000 & 1.000 & 1.00 & 0.00 & 1.00 &  957 \\
 & \texttt{piecewise\_missing\_thresh.} &  60 & 1.000 & 1.000 & 1.000 & 1.00 & 0.00 & 1.00 &  451 \\
 & \texttt{rankdef\_linear\_shared}     &  60 & 0.917 & 0.903 & 0.883 & 1.37 & 0.43 & 0.93 &  998 \\
 & \texttt{recurrence\_missing\_init}   &  60 & 1.000 & 1.000 & 1.000 & 1.00 & 0.00 & 1.00 &  619 \\
\midrule
\multirow{11}{*}{B}
 & \texttt{birth\_death\_missing\_rate}  & 150 & 0.980 & 0.974 & 0.967 & 1.09 & 0.10 & 0.99 &  804 \\
 & \texttt{circuit\_missing\_resistance} & 150 & 1.000 & 1.000 & 1.000 & 1.00 & 0.00 & 1.00 &  561 \\
 & \texttt{deconvolution}               & 150 & 0.153 & 0.066 & 0.033 & 5.59 & 5.40 & 0.19 & 1250 \\
 & \texttt{discrete\_tomography}        & 150 & 0.567 & 0.366 & 0.233 & 2.41 & 1.84 & 0.57 &  847 \\
 & \texttt{eigenvector\_missing\_entry}  & 150 & 0.933 & 0.875 & 0.853 & 1.52 & 0.59 & 0.93 &  675 \\
 & \texttt{laplace\_grid}               & 150 & 0.227 & 0.607 & 0.533 & 2.62 & 1.85 & 0.77 & 1470 \\
 & \texttt{linear\_sys\_missing\_coeff}  & 150 & 0.693 & 0.622 & 0.520 & 2.35 & 1.55 & 0.79 & 1513 \\
 & \texttt{markov\_missing\_transition}  & 150 & 0.833 & 0.810 & 0.640 & 1.47 & 0.48 & 0.99 & 1752 \\
 & \texttt{poly\_interpolation}         & 150 & 0.807 & 0.604 & 0.467 & 2.47 & 1.61 & 0.85 & 1486 \\
 & \texttt{portfolio\_var\_missing\_c.}  & 150 & 0.960 & 0.986 & 0.980 & 1.05 & 0.06 & 0.99 &  732 \\
 & \texttt{steady\_state\_missing\_em.}  & 150 & 0.800 & 0.982 & 0.973 & 1.07 & 0.07 & 1.00 & 1501 \\
\bottomrule
\end{tabular}
\end{adjustbox}
\end{table}


\begin{table}[p]
\centering
\caption{Full per-family metrics (\texttt{gpt-5.1}, four-shot, 20/50 typed dataset).}
\label{tab:per-family-full-51-4s}
\begin{adjustbox}{max width=\textwidth}
\begin{tabular}{llr ccccccc}
\toprule
\textbf{Type} & \textbf{Family} & $n$ & \textbf{Acc} & \textbf{Hit Rate} & \textbf{1st-Req} & \textbf{Avg Req} & \textbf{Avg Decl} & \textbf{Avg Hints} & \textbf{Avg Tok} \\
\midrule
\multirow{11}{*}{A}
 & \texttt{bayes\_missing\_prior}       &  60 & 1.000 & 1.000 & 1.000 & 1.00 & 0.00 & 1.00 & 404 \\
 & \texttt{crt\_reconstruction}         &  60 & 0.000 & 1.000 & 1.000 & 1.00 & 0.00 & 1.00 & 783 \\
 & \texttt{geometry\_coordinates}       &  60 & 0.317 & 0.349 & 0.267 & 3.87 & 3.32 & 0.55 & 1084 \\
 & \texttt{graph\_path\_sums}           &  60 & 0.050 & 0.094 & 0.083 & 5.37 & 5.25 & 0.12 & 1149 \\
 & \texttt{linear\_system\_separator}   &  60 & 1.000 & 1.000 & 1.000 & 1.00 & 0.00 & 1.00 & 676 \\
 & \texttt{matrix\_completion}          &  60 & 0.717 & 0.708 & 0.683 & 2.33 & 1.57 & 0.77 & 721 \\
 & \texttt{moment\_problem}             &  60 & 0.000 & 0.022 & 0.017 & 5.87 & 5.82 & 0.05 & 1165 \\
 & \texttt{phase\_retrieval}            &  60 & 0.983 & 1.000 & 1.000 & 1.00 & 0.00 & 1.00 & 1081 \\
 & \texttt{piecewise\_missing\_thresh.} &  60 & 1.000 & 1.000 & 1.000 & 1.00 & 0.00 & 1.00 & 453 \\
 & \texttt{rankdef\_linear\_shared}     &  60 & 0.850 & 0.548 & 0.217 & 2.07 & 1.17 & 0.90 & 1149 \\
 & \texttt{recurrence\_missing\_init}   &  60 & 1.000 & 1.000 & 1.000 & 1.00 & 0.00 & 1.00 & 529 \\
\midrule
\multirow{11}{*}{B}
 & \texttt{birth\_death\_missing\_rate}  & 150 & 0.953 & 0.939 & 0.920 & 1.23 & 0.27 & 0.96 & 816 \\
 & \texttt{circuit\_missing\_resistance} & 150 & 0.980 & 1.000 & 1.000 & 1.00 & 0.00 & 1.00 & 554 \\
 & \texttt{deconvolution}                & 150 & 0.153 & 0.131 & 0.107 & 5.30 & 5.07 & 0.23 & 1223 \\
 & \texttt{discrete\_tomography}         & 150 & 0.513 & 0.316 & 0.173 & 2.48 & 1.97 & 0.51 & 828 \\
 & \texttt{eigenvector\_missing\_entry}  & 150 & 0.960 & 0.938 & 0.927 & 1.31 & 0.35 & 0.96 & 624 \\
 & \texttt{laplace\_grid}                & 150 & 0.227 & 0.594 & 0.513 & 2.59 & 1.83 & 0.76 & 1421 \\
 & \texttt{linear\_sys\_missing\_coeff}  & 150 & 0.573 & 0.531 & 0.427 & 2.72 & 2.01 & 0.71 & 1421 \\
 & \texttt{markov\_missing\_transition}  & 150 & 0.847 & 0.818 & 0.647 & 1.42 & 0.43 & 0.99 & 1622 \\
 & \texttt{poly\_interpolation}          & 150 & 0.847 & 0.740 & 0.660 & 1.87 & 0.98 & 0.89 & 1338 \\
 & \texttt{portfolio\_var\_missing\_c.}  & 150 & 0.973 & 0.996 & 0.993 & 1.01 & 0.01 & 1.00 & 743 \\
 & \texttt{steady\_state\_missing\_em.}  & 150 & 0.787 & 0.981 & 0.967 & 1.05 & 0.05 & 1.00 & 1337 \\
\bottomrule
\end{tabular}
\end{adjustbox}
\end{table}


\begin{table}[p]
\centering
\caption{Four-shot effect: accuracy change per family (\texttt{gpt-5.1}, 20/50 typed dataset). $\Delta$ = four-shot accuracy $-$ baseline accuracy.}
\label{tab:fourshot-effect-51}
\begin{adjustbox}{max width=\textwidth}
\begin{tabular}{llr ccc ccc}
\toprule
& & & \multicolumn{3}{c}{\textbf{Accuracy}} & \multicolumn{3}{c}{\textbf{1st-Request Success}} \\
\cmidrule(lr){4-6} \cmidrule(lr){7-9}
\textbf{Type} & \textbf{Family} & $n$ & \textbf{Base} & \textbf{4S} & $\Delta$ & \textbf{Base} & \textbf{4S} & $\Delta$ \\
\midrule
\multirow{11}{*}{A}
 & \texttt{bayes\_missing\_prior}       &  60 & 1.000 & 1.000 &  0.000 & 1.000 & 1.000 &  0.000 \\
 & \texttt{crt\_reconstruction}         &  60 & 0.000 & 0.000 &  0.000 & 1.000 & 1.000 &  0.000 \\
 & \texttt{geometry\_coordinates}       &  60 & 0.117 & 0.317 & \textbf{+0.200} & 0.067 & 0.267 & \textbf{+0.200} \\
 & \texttt{graph\_path\_sums}           &  60 & 0.017 & 0.050 & +0.033 & 0.017 & 0.083 & +0.067 \\
 & \texttt{linear\_system\_separator}   &  60 & 1.000 & 1.000 &  0.000 & 0.133 & 1.000 & \textbf{+0.867} \\
 & \texttt{matrix\_completion}          &  60 & 0.700 & 0.717 & +0.017 & 0.667 & 0.683 & +0.017 \\
 & \texttt{moment\_problem}             &  60 & 0.000 & 0.000 &  0.000 & 0.000 & 0.017 & +0.017 \\
 & \texttt{phase\_retrieval}            &  60 & 0.933 & 0.983 & +0.050 & 1.000 & 1.000 &  0.000 \\
 & \texttt{piecewise\_missing\_thresh.} &  60 & 1.000 & 1.000 &  0.000 & 1.000 & 1.000 &  0.000 \\
 & \texttt{rankdef\_linear\_shared}     &  60 & 0.917 & 0.850 & $-$0.067 & 0.883 & 0.217 & $-$0.667 \\
 & \texttt{recurrence\_missing\_init}   &  60 & 1.000 & 1.000 &  0.000 & 1.000 & 1.000 &  0.000 \\
\cmidrule(lr){2-9}
 & \textit{Type A average}              & 660 & 0.608 & 0.629 & +0.021 & 0.615 & 0.661 & +0.046 \\
\midrule
\multirow{11}{*}{B}
 & \texttt{birth\_death\_missing\_rate}  & 150 & 0.980 & 0.953 & $-$0.027 & 0.967 & 0.920 & $-$0.047 \\
 & \texttt{circuit\_missing\_resistance} & 150 & 1.000 & 0.980 & $-$0.020 & 1.000 & 1.000 &  0.000 \\
 & \texttt{deconvolution}               & 150 & 0.153 & 0.153 &  0.000 & 0.033 & 0.107 & +0.073 \\
 & \texttt{discrete\_tomography}        & 150 & 0.567 & 0.513 & $-$0.053 & 0.233 & 0.173 & $-$0.060 \\
 & \texttt{eigenvector\_missing\_entry}  & 150 & 0.933 & 0.960 & +0.027 & 0.853 & 0.927 & +0.073 \\
 & \texttt{laplace\_grid}               & 150 & 0.227 & 0.227 &  0.000 & 0.533 & 0.520 & $-$0.013 \\
 & \texttt{linear\_sys\_missing\_coeff}  & 150 & 0.693 & 0.573 & $-$0.120 & 0.520 & 0.427 & $-$0.093 \\
 & \texttt{markov\_missing\_transition}  & 150 & 0.833 & 0.847 & +0.013 & 0.640 & 0.647 & +0.007 \\
 & \texttt{poly\_interpolation}         & 150 & 0.807 & 0.847 & +0.040 & 0.467 & 0.660 & \textbf{+0.193} \\
 & \texttt{portfolio\_var\_missing\_c.}  & 150 & 0.960 & 0.973 & +0.013 & 0.980 & 0.993 & +0.013 \\
 & \texttt{steady\_state\_missing\_em.}  & 150 & 0.800 & 0.787 & $-$0.013 & 0.973 & 0.967 & $-$0.007 \\
\cmidrule(lr){2-9}
 & \textit{Type B average}             & 1650 & 0.723 & 0.710 & $-$0.013 & 0.655 & 0.667 & +0.013 \\
\midrule
 & \textbf{Overall}                    & 2310 & \textbf{0.690} & \textbf{0.687} & $-$0.003 & 0.643 & 0.665 & +0.022 \\
\bottomrule
\end{tabular}
\end{adjustbox}
\end{table}
 


\begin{table}[t]
\centering
\caption{Per-family accuracy on the 20/50 typed dataset (\texttt{gemini-2.5-flash}, 2310 instances). Families sorted by Type then name. $n$ = instances per family.}
\label{tab:per-family-acc-gemini25flash}
\begin{adjustbox}{max width=\textwidth}
\begin{tabular}{llr cc cc}
\toprule
& & & \multicolumn{2}{c}{\textbf{zero-shot}} & \multicolumn{2}{c}{\textbf{four-shot}} \\
\cmidrule(lr){4-5} \cmidrule(lr){6-7}
\textbf{Type} & \textbf{Family} & $n$ & \textbf{Acc} & \textbf{1st-Req} & \textbf{Acc} & \textbf{1st-Req} \\
\midrule
\multirow{11}{*}{A}
 & \texttt{bayes\_missing\_prior}       &  60 & 0.983 & 1.000 & 0.867 & 1.000 \\
 & \texttt{crt\_reconstruction}         &  60 & 0.000 & 1.000 & 0.000 & 1.000 \\
 & \texttt{geometry\_coordinates}       &  60 & 0.633 & 0.517 & 0.300 & 0.550 \\
 & \texttt{graph\_path\_sums}           &  60 & 0.000 & 0.267 & 0.000 & 0.050 \\
 & \texttt{linear\_system\_separator}   &  60 & 0.000 & 0.033 & 0.000 & 0.050 \\
 & \texttt{matrix\_completion}          &  60 & 0.233 & 0.417 & 0.567 & 0.850 \\
 & \texttt{moment\_problem}             &  60 & 0.000 & 0.033 & 0.000 & 0.017 \\
 & \texttt{phase\_retrieval}            &  60 & 0.000 & 0.983 & 0.000 & 1.000 \\
 & \texttt{piecewise\_missing\_thresh.} &  60 & 1.000 & 1.000 & 0.950 & 1.000 \\
 & \texttt{rankdef\_linear\_shared}     &  60 & 0.000 & 0.983 & 0.000 & 0.917 \\
 & \texttt{recurrence\_missing\_init}   &  60 & 0.983 & 1.000 & 0.967 & 1.000 \\
\midrule
\multirow{11}{*}{B}
 & \texttt{birth\_death\_missing\_rate}  & 150 & 0.640 & 0.987 & 0.267 & 0.807 \\
 & \texttt{circuit\_missing\_resistance} & 150 & 0.827 & 1.000 & 0.573 & 0.993 \\
 & \texttt{deconvolution}               & 150 & 0.000 & 0.240 & 0.000 & 0.027 \\
 & \texttt{discrete\_tomography}        & 150 & 0.473 & 0.220 & 0.300 & 0.113 \\
 & \texttt{eigenvector\_missing\_entry}  & 150 & 0.940 & 0.980 & 0.667 & 0.840 \\
 & \texttt{laplace\_grid}               & 150 & 0.067 & 0.540 & 0.000 & 0.527 \\
 & \texttt{linear\_sys\_missing\_coeff}  & 150 & 0.193 & 0.867 & 0.107 & 0.513 \\
 & \texttt{markov\_missing\_transition}  & 150 & 0.393 & 0.607 & 0.427 & 0.567 \\
 & \texttt{poly\_interpolation}         & 150 & 0.873 & 0.980 & 0.213 & 0.967 \\
 & \texttt{portfolio\_var\_missing\_c.}  & 150 & 0.860 & 1.000 & 0.547 & 1.000 \\
 & \texttt{steady\_state\_missing\_em.}  & 150 & 0.740 & 1.000 & 0.487 & 0.993 \\
\bottomrule
\end{tabular}
\end{adjustbox}
\end{table}


\begin{table}[p]
\centering
\caption{Full per-family metrics (\texttt{gemini-2.5-flash}, zero-shot, 20/50 typed dataset).}
\label{tab:per-family-full-gemini25flash}
\begin{adjustbox}{max width=\textwidth}
\begin{tabular}{llr ccccccc}
\toprule
\textbf{Type} & \textbf{Family} & $n$ & \textbf{Acc} & \textbf{Hit Rate} & \textbf{1st-Req} & \textbf{Avg Req} & \textbf{Avg Decl} & \textbf{Avg Hints} & \textbf{Avg Tok} \\
\midrule
\multirow{11}{*}{A}
 & \texttt{bayes\_missing\_prior}       &  60 & 0.983 & 1.000 & 1.000 & 1.00 & 0.00 & 1.00 & 285 \\
 & \texttt{crt\_reconstruction}         &  60 & 0.000 & 1.000 & 1.000 & 1.00 & 0.00 & 1.00 & 335 \\
 & \texttt{geometry\_coordinates}       &  60 & 0.633 & 0.617 & 0.517 & 2.37 & 1.60 & 0.77 & 424 \\
 & \texttt{graph\_path\_sums}           &  60 & 0.000 & 0.267 & 0.267 & 4.52 & 4.25 & 0.27 & 641 \\
 & \texttt{linear\_system\_separator}   &  60 & 0.000 & 0.356 & 0.033 & 2.93 & 1.93 & 1.00 & 516 \\
 & \texttt{matrix\_completion}          &  60 & 0.233 & 0.467 & 0.417 & 2.87 & 2.25 & 0.62 & 445 \\
 & \texttt{moment\_problem}             &  60 & 0.000 & 0.054 & 0.033 & 5.78 & 5.68 & 0.10 & 671 \\
 & \texttt{phase\_retrieval}            &  60 & 0.000 & 0.992 & 0.983 & 1.02 & 0.02 & 1.00 & 762 \\
 & \texttt{piecewise\_missing\_thresh.} &  60 & 1.000 & 1.000 & 1.000 & 1.00 & 0.00 & 1.00 & 216 \\
 & \texttt{rankdef\_linear\_shared}     &  60 & 0.000 & 0.988 & 0.983 & 1.05 & 0.05 & 1.00 & 497 \\
 & \texttt{recurrence\_missing\_init}   &  60 & 0.983 & 1.000 & 1.000 & 1.00 & 0.00 & 1.00 & 273 \\
\midrule
\multirow{11}{*}{B}
 & \texttt{birth\_death\_missing\_rate}  & 150 & 0.640 & 0.990 & 0.987 & 1.02 & 0.03 & 0.99 & 451 \\
 & \texttt{circuit\_missing\_resistance} & 150 & 0.827 & 1.000 & 1.000 & 1.00 & 0.00 & 1.00 & 311 \\
 & \texttt{deconvolution}               & 150 & 0.000 & 0.426 & 0.240 & 2.91 & 2.24 & 0.67 & 487 \\
 & \texttt{discrete\_tomography}        & 150 & 0.473 & 0.381 & 0.220 & 2.34 & 1.75 & 0.59 & 540 \\
 & \texttt{eigenvector\_missing\_entry}  & 150 & 0.940 & 0.988 & 0.980 & 1.03 & 0.03 & 1.00 & 245 \\
 & \texttt{laplace\_grid}               & 150 & 0.067 & 0.617 & 0.540 & 2.39 & 1.67 & 0.72 & 1048 \\
 & \texttt{linear\_sys\_missing\_coeff}  & 150 & 0.193 & 0.928 & 0.867 & 1.17 & 0.17 & 1.00 & 617 \\
 & \texttt{markov\_missing\_transition}  & 150 & 0.393 & 0.800 & 0.607 & 1.38 & 0.39 & 0.99 & 830 \\
 & \texttt{poly\_interpolation}         & 150 & 0.873 & 0.986 & 0.980 & 1.03 & 0.04 & 0.99 & 671 \\
 & \texttt{portfolio\_var\_missing\_c.}  & 150 & 0.860 & 1.000 & 1.000 & 1.00 & 0.00 & 1.00 & 441 \\
 & \texttt{steady\_state\_missing\_em.}  & 150 & 0.740 & 1.000 & 1.000 & 1.00 & 0.00 & 1.00 & 547 \\
\bottomrule
\end{tabular}
\end{adjustbox}
\end{table}


\begin{table}[p]
\centering
\caption{Full per-family metrics (\texttt{gemini-2.5-flash}, four-shot, 20/50 typed dataset).}
\label{tab:per-family-full-gemini25flash-4s}
\begin{adjustbox}{max width=\textwidth}
\begin{tabular}{llr ccccccc}
\toprule
\textbf{Type} & \textbf{Family} & $n$ & \textbf{Acc} & \textbf{Hit Rate} & \textbf{1st-Req} & \textbf{Avg Req} & \textbf{Avg Decl} & \textbf{Avg Hints} & \textbf{Avg Tok} \\
\midrule
\multirow{11}{*}{A}
 & \texttt{bayes\_missing\_prior}       &  60 & 0.867 & 1.000 & 1.000 & 1.00 & 0.00 & 1.00 & 261 \\
 & \texttt{crt\_reconstruction}         &  60 & 0.000 & 1.000 & 1.000 & 1.00 & 0.00 & 1.00 & 501 \\
 & \texttt{geometry\_coordinates}       &  60 & 0.300 & 0.699 & 0.550 & 2.22 & 1.33 & 0.88 & 406 \\
 & \texttt{graph\_path\_sums}           &  60 & 0.000 & 0.050 & 0.050 & 5.75 & 5.70 & 0.05 & 658 \\
 & \texttt{linear\_system\_separator}   &  60 & 0.000 & 0.354 & 0.050 & 2.92 & 1.95 & 0.97 & 544 \\
 & \texttt{matrix\_completion}          &  60 & 0.567 & 0.879 & 0.850 & 1.55 & 0.57 & 0.98 & 268 \\
 & \texttt{moment\_problem}             &  60 & 0.000 & 0.017 & 0.017 & 5.97 & 5.95 & 0.02 & 676 \\
 & \texttt{phase\_retrieval}            &  60 & 0.000 & 1.000 & 1.000 & 1.00 & 0.00 & 1.00 & 894 \\
 & \texttt{piecewise\_missing\_thresh.} &  60 & 0.950 & 1.000 & 1.000 & 1.00 & 0.00 & 1.00 & 223 \\
 & \texttt{rankdef\_linear\_shared}     &  60 & 0.000 & 0.953 & 0.917 & 1.15 & 0.15 & 1.00 & 567 \\
 & \texttt{recurrence\_missing\_init}   &  60 & 0.967 & 1.000 & 1.000 & 1.00 & 0.00 & 1.00 & 270 \\
\midrule
\multirow{11}{*}{B}
 & \texttt{birth\_death\_missing\_rate}  & 150 & 0.267 & 0.884 & 0.807 & 1.27 & 0.29 & 0.98 & 476 \\
 & \texttt{circuit\_missing\_resistance} & 150 & 0.573 & 0.997 & 0.993 & 1.01 & 0.01 & 1.00 & 289 \\
 & \texttt{deconvolution}                & 150 & 0.000 & 0.071 & 0.027 & 5.43 & 5.15 & 0.28 & 724 \\
 & \texttt{discrete\_tomography}         & 150 & 0.300 & 0.210 & 0.113 & 2.63 & 2.30 & 0.33 & 422 \\
 & \texttt{eigenvector\_missing\_entry}  & 150 & 0.667 & 0.893 & 0.840 & 1.46 & 0.51 & 0.95 & 332 \\
 & \texttt{laplace\_grid}                & 150 & 0.000 & 0.578 & 0.527 & 2.78 & 2.12 & 0.66 & 1813 \\
 & \texttt{linear\_sys\_missing\_coeff}  & 150 & 0.107 & 0.563 & 0.513 & 2.73 & 2.08 & 0.65 & 1149 \\
 & \texttt{markov\_missing\_transition}  & 150 & 0.427 & 0.773 & 0.567 & 1.48 & 0.49 & 0.99 & 1039 \\
 & \texttt{poly\_interpolation}          & 150 & 0.213 & 0.973 & 0.967 & 1.09 & 0.11 & 0.98 & 887 \\
 & \texttt{portfolio\_var\_missing\_c.}  & 150 & 0.547 & 1.000 & 1.000 & 1.00 & 0.00 & 1.00 & 470 \\
 & \texttt{steady\_state\_missing\_em.}  & 150 & 0.487 & 0.997 & 0.993 & 1.01 & 0.01 & 1.00 & 634 \\
\bottomrule
\end{tabular}
\end{adjustbox}
\end{table}


\begin{table}[p]
\centering
\caption{Four-shot effect: accuracy change per family (\texttt{gemini-2.5-flash}, 20/50 typed dataset). $\Delta$ = four-shot accuracy $-$ baseline accuracy. Positive values indicate improvement.}
\label{tab:fourshot-effect-gemini25flash}
\begin{adjustbox}{max width=\textwidth}
\begin{tabular}{llr ccc ccc}
\toprule
& & & \multicolumn{3}{c}{\textbf{Accuracy}} & \multicolumn{3}{c}{\textbf{1st-Request Success}} \\
\cmidrule(lr){4-6} \cmidrule(lr){7-9}
\textbf{Type} & \textbf{Family} & $n$ & \textbf{Base} & \textbf{4S} & $\Delta$ & \textbf{Base} & \textbf{4S} & $\Delta$ \\
\midrule
\multirow{11}{*}{A}
 & \texttt{bayes\_missing\_prior}       &  60 & 0.983 & 0.867 & $-$0.117 & 1.000 & 1.000 &  0.000 \\
 & \texttt{crt\_reconstruction}         &  60 & 0.000 & 0.000 &  0.000 & 1.000 & 1.000 &  0.000 \\
 & \texttt{geometry\_coordinates}       &  60 & 0.633 & 0.300 & $-$0.333 & 0.517 & 0.550 & +0.033 \\
 & \texttt{graph\_path\_sums}           &  60 & 0.000 & 0.000 &  0.000 & 0.267 & 0.050 & $-$0.217 \\
 & \texttt{linear\_system\_separator}   &  60 & 0.000 & 0.000 &  0.000 & 0.033 & 0.050 & +0.017 \\
 & \texttt{matrix\_completion}          &  60 & 0.233 & 0.567 & \textbf{+0.333} & 0.417 & 0.850 & \textbf{+0.433} \\
 & \texttt{moment\_problem}             &  60 & 0.000 & 0.000 &  0.000 & 0.033 & 0.017 & $-$0.017 \\
 & \texttt{phase\_retrieval}            &  60 & 0.000 & 0.000 &  0.000 & 0.983 & 1.000 & +0.017 \\
 & \texttt{piecewise\_missing\_thresh.} &  60 & 1.000 & 0.950 & $-$0.050 & 1.000 & 1.000 &  0.000 \\
 & \texttt{rankdef\_linear\_shared}     &  60 & 0.000 & 0.000 &  0.000 & 0.983 & 0.917 & $-$0.067 \\
 & \texttt{recurrence\_missing\_init}   &  60 & 0.983 & 0.967 & $-$0.017 & 1.000 & 1.000 &  0.000 \\
\cmidrule(lr){2-9}
 & \textit{Type A average}              & 660 & 0.348 & 0.332 & $-$0.017 & 0.658 & 0.676 & +0.018 \\
\midrule
\multirow{11}{*}{B}
 & \texttt{birth\_death\_missing\_rate}  & 150 & 0.640 & 0.267 & $-$0.373 & 0.987 & 0.807 & $-$0.180 \\
 & \texttt{circuit\_missing\_resistance} & 150 & 0.827 & 0.573 & $-$0.253 & 1.000 & 0.993 & $-$0.007 \\
 & \texttt{deconvolution}               & 150 & 0.000 & 0.000 &  0.000 & 0.240 & 0.027 & $-$0.213 \\
 & \texttt{discrete\_tomography}        & 150 & 0.473 & 0.300 & $-$0.173 & 0.220 & 0.113 & $-$0.107 \\
 & \texttt{eigenvector\_missing\_entry}  & 150 & 0.940 & 0.667 & $-$0.273 & 0.980 & 0.840 & $-$0.140 \\
 & \texttt{laplace\_grid}               & 150 & 0.067 & 0.000 & $-$0.067 & 0.540 & 0.527 & $-$0.013 \\
 & \texttt{linear\_sys\_missing\_coeff}  & 150 & 0.193 & 0.107 & $-$0.087 & 0.867 & 0.513 & $-$0.353 \\
 & \texttt{markov\_missing\_transition}  & 150 & 0.393 & 0.427 & +0.033 & 0.607 & 0.567 & $-$0.040 \\
 & \texttt{poly\_interpolation}         & 150 & 0.873 & 0.213 & $-$0.660 & 0.980 & 0.967 & $-$0.013 \\
 & \texttt{portfolio\_var\_missing\_c.}  & 150 & 0.860 & 0.547 & $-$0.313 & 1.000 & 1.000 &  0.000 \\
 & \texttt{steady\_state\_missing\_em.}  & 150 & 0.740 & 0.487 & $-$0.253 & 1.000 & 0.993 & $-$0.007 \\
\cmidrule(lr){2-9}
 & \textit{Type B average}             & 1650 & 0.546 & 0.326 & $-$0.220 & 0.765 & 0.668 & $-$0.098 \\
\midrule
 & \textbf{Overall}                    & 2310 & \textbf{0.490} & \textbf{0.328} & $-$0.162 & 0.735 & 0.670 & $-$0.065 \\
\bottomrule
\end{tabular}
\end{adjustbox}
\end{table}
 



\begin{table}[t]
\centering
\caption{Per-family accuracy on the 20/50 typed dataset (\texttt{Gemma-4}, 2310 instances). Families sorted by Type then name. $n$ = instances per family.}
\label{tab:per-family-acc-gemma4}
\begin{adjustbox}{max width=\textwidth}
\begin{tabular}{llr cc cc}
\toprule
& & & \multicolumn{2}{c}{\textbf{zero-shot}} & \multicolumn{2}{c}{\textbf{four-shot}} \\
\cmidrule(lr){4-5} \cmidrule(lr){6-7}
\textbf{Type} & \textbf{Family} & $n$ & \textbf{Acc} & \textbf{1st-Req} & \textbf{Acc} & \textbf{1st-Req} \\
\midrule
\multirow{11}{*}{A}
 & \texttt{bayes\_missing\_prior}       &  60 & 1.000 & 1.000 & 0.983 & 1.000 \\
 & \texttt{crt\_reconstruction}         &  60 & 0.000 & 1.000 & 0.000 & 1.000 \\
 & \texttt{geometry\_coordinates}       &  60 & 0.200 & 0.083 & 0.133 & 0.117 \\
 & \texttt{graph\_path\_sums}           &  60 & 0.000 & 0.000 & 0.000 & 0.000 \\
 & \texttt{linear\_system\_separator}   &  60 & 1.000 & 0.067 & 0.950 & 0.367 \\
 & \texttt{matrix\_completion}          &  60 & 0.950 & 0.883 & 0.867 & 0.583 \\
 & \texttt{moment\_problem}             &  60 & 0.017 & 0.200 & 0.033 & 0.000 \\
 & \texttt{phase\_retrieval}            &  60 & 0.850 & 1.000 & 0.983 & 0.983 \\
 & \texttt{piecewise\_missing\_thresh.} &  60 & 1.000 & 1.000 & 1.000 & 1.000 \\
 & \texttt{rankdef\_linear\_shared}     &  60 & 0.883 & 0.350 & 0.817 & 0.000 \\
 & \texttt{recurrence\_missing\_init}   &  60 & 1.000 & 1.000 & 0.983 & 1.000 \\
\midrule
\multirow{11}{*}{B}
 & \texttt{birth\_death\_missing\_rate}  & 150 & 0.980 & 0.947 & 0.920 & 0.907 \\
 & \texttt{circuit\_missing\_resistance} & 150 & 1.000 & 1.000 & 0.987 & 1.000 \\
 & \texttt{deconvolution}               & 150 & 0.413 & 0.233 & 0.320 & 0.060 \\
 & \texttt{discrete\_tomography}        & 150 & 0.947 & 0.620 & 0.927 & 0.580 \\
 & \texttt{eigenvector\_missing\_entry}  & 150 & 0.887 & 0.853 & 0.847 & 0.800 \\
 & \texttt{laplace\_grid}               & 150 & 0.147 & 0.320 & 0.107 & 0.247 \\
 & \texttt{linear\_sys\_missing\_coeff}  & 150 & 0.773 & 0.647 & 0.667 & 0.553 \\
 & \texttt{markov\_missing\_transition}  & 150 & 0.893 & 0.607 & 0.873 & 0.740 \\
 & \texttt{poly\_interpolation}         & 150 & 0.700 & 0.200 & 0.773 & 0.727 \\
 & \texttt{portfolio\_var\_missing\_c.}  & 150 & 0.960 & 1.000 & 0.980 & 1.000 \\
 & \texttt{steady\_state\_missing\_em.}  & 150 & 0.913 & 0.973 & 0.833 & 0.940 \\
\bottomrule
\end{tabular}
\end{adjustbox}
\end{table}


\begin{table}[p]
\centering
\caption{Full per-family metrics (\texttt{Gemma-4}, zero-shot, 20/50 typed dataset).}
\label{tab:per-family-full-gemma4}
\begin{adjustbox}{max width=\textwidth}
\begin{tabular}{llr ccccccc}
\toprule
\textbf{Type} & \textbf{Family} & $n$ & \textbf{Acc} & \textbf{Hit Rate} & \textbf{1st-Req} & \textbf{Avg Req} & \textbf{Avg Decl} & \textbf{Avg Hints} & \textbf{Avg Tok} \\
\midrule
\multirow{11}{*}{A}
 & \texttt{bayes\_missing\_prior}       &  60 & 1.000 & 1.000 & 1.000 & 1.00 & 0.00 & 1.00 & 314 \\
 & \texttt{crt\_reconstruction}         &  60 & 0.000 & 1.000 & 1.000 & 1.00 & 0.00 & 1.00 & 460 \\
 & \texttt{geometry\_coordinates}       &  60 & 0.200 & 0.105 & 0.083 & 5.18 & 4.98 & 0.20 & 893 \\
 & \texttt{graph\_path\_sums}           &  60 & 0.000 & 0.000 & 0.000 & 6.00 & 6.00 & 0.00 & 2019 \\
 & \texttt{linear\_system\_separator}   &  60 & 1.000 & 0.408 & 0.067 & 2.68 & 1.68 & 1.00 & 588 \\
 & \texttt{matrix\_completion}          &  60 & 0.950 & 0.907 & 0.883 & 1.30 & 0.33 & 0.97 & 480 \\
 & \texttt{moment\_problem}             &  60 & 0.017 & 0.214 & 0.200 & 4.72 & 4.47 & 0.25 & 1771 \\
 & \texttt{phase\_retrieval}            &  60 & 0.850 & 1.000 & 1.000 & 1.00 & 0.00 & 1.00 & 712 \\
 & \texttt{piecewise\_missing\_thresh.} &  60 & 1.000 & 1.000 & 1.000 & 1.00 & 0.00 & 1.00 & 288 \\
 & \texttt{rankdef\_linear\_shared}     &  60 & 0.883 & 0.617 & 0.350 & 2.07 & 1.18 & 0.88 & 740 \\
 & \texttt{recurrence\_missing\_init}   &  60 & 1.000 & 1.000 & 1.000 & 1.00 & 0.00 & 1.00 & 323 \\
\midrule
\multirow{11}{*}{B}
 & \texttt{birth\_death\_missing\_rate}  & 150 & 0.980 & 0.963 & 0.947 & 1.09 & 0.11 & 0.98 & 561 \\
 & \texttt{circuit\_missing\_resistance} & 150 & 1.000 & 1.000 & 1.000 & 1.00 & 0.00 & 1.00 & 404 \\
 & \texttt{deconvolution}               & 150 & 0.413 & 0.299 & 0.233 & 3.85 & 3.42 & 0.43 & 2866 \\
 & \texttt{discrete\_tomography}        & 150 & 0.947 & 0.764 & 0.620 & 1.55 & 0.60 & 0.95 & 839 \\
 & \texttt{eigenvector\_missing\_entry}  & 150 & 0.887 & 0.874 & 0.853 & 1.59 & 0.69 & 0.91 & 633 \\
 & \texttt{laplace\_grid}               & 150 & 0.147 & 0.336 & 0.320 & 4.39 & 4.03 & 0.36 & 3704 \\
 & \texttt{linear\_sys\_missing\_coeff}  & 150 & 0.773 & 0.740 & 0.647 & 1.94 & 1.11 & 0.83 & 1044 \\
 & \texttt{markov\_missing\_transition}  & 150 & 0.893 & 0.802 & 0.607 & 1.40 & 0.40 & 1.00 & 914 \\
 & \texttt{poly\_interpolation}         & 150 & 0.700 & 0.424 & 0.200 & 2.97 & 2.21 & 0.75 & 1082 \\
 & \texttt{portfolio\_var\_missing\_c.}  & 150 & 0.960 & 1.000 & 1.000 & 1.00 & 0.00 & 1.00 & 605 \\
 & \texttt{steady\_state\_missing\_em.}  & 150 & 0.913 & 0.987 & 0.973 & 1.03 & 0.03 & 1.00 & 844 \\
\bottomrule
\end{tabular}
\end{adjustbox}
\end{table}


\begin{table}[p]
\centering
\caption{Full per-family metrics (\texttt{Gemma-4}, four-shot, 20/50 typed dataset).}
\label{tab:per-family-full-gemma4-4s}
\begin{adjustbox}{max width=\textwidth}
\begin{tabular}{llr ccccccc}
\toprule
\textbf{Type} & \textbf{Family} & $n$ & \textbf{Acc} & \textbf{Hit Rate} & \textbf{1st-Req} & \textbf{Avg Req} & \textbf{Avg Decl} & \textbf{Avg Hints} & \textbf{Avg Tok} \\
\midrule
\multirow{11}{*}{A}
 & \texttt{bayes\_missing\_prior}       &  60 & 0.983 & 1.000 & 1.000 & 1.00 & 0.00 & 1.00 & 309 \\
 & \texttt{crt\_reconstruction}         &  60 & 0.000 & 1.000 & 1.000 & 1.00 & 0.00 & 1.00 & 455 \\
 & \texttt{geometry\_coordinates}       &  60 & 0.133 & 0.121 & 0.117 & 5.18 & 5.05 & 0.13 & 816 \\
 & \texttt{graph\_path\_sums}           &  60 & 0.000 & 0.000 & 0.000 & 6.00 & 6.00 & 0.00 & 1953 \\
 & \texttt{linear\_system\_separator}   &  60 & 0.950 & 0.600 & 0.367 & 2.13 & 1.13 & 1.00 & 534 \\
 & \texttt{matrix\_completion}          &  60 & 0.867 & 0.655 & 0.583 & 2.42 & 1.52 & 0.90 & 746 \\
 & \texttt{moment\_problem}             &  60 & 0.033 & 0.017 & 0.000 & 5.78 & 5.72 & 0.07 & 2049 \\
 & \texttt{phase\_retrieval}            &  60 & 0.983 & 0.992 & 0.983 & 1.02 & 0.02 & 1.00 & 850 \\
 & \texttt{piecewise\_missing\_thresh.} &  60 & 1.000 & 1.000 & 1.000 & 1.00 & 0.00 & 1.00 & 265 \\
 & \texttt{rankdef\_linear\_shared}     &  60 & 0.817 & 0.414 & 0.000 & 2.33 & 1.50 & 0.83 & 659 \\
 & \texttt{recurrence\_missing\_init}   &  60 & 0.983 & 1.000 & 1.000 & 1.00 & 0.00 & 1.00 & 312 \\
\midrule
\multirow{11}{*}{B}
 & \texttt{birth\_death\_missing\_rate}  & 150 & 0.920 & 0.916 & 0.907 & 1.47 & 0.54 & 0.93 & 700 \\
 & \texttt{circuit\_missing\_resistance} & 150 & 0.987 & 1.000 & 1.000 & 1.00 & 0.00 & 1.00 & 406 \\
 & \texttt{deconvolution}                & 150 & 0.320 & 0.145 & 0.060 & 4.83 & 4.49 & 0.34 & 2986 \\
 & \texttt{discrete\_tomography}         & 150 & 0.927 & 0.734 & 0.580 & 1.62 & 0.69 & 0.93 & 802 \\
 & \texttt{eigenvector\_missing\_entry}  & 150 & 0.847 & 0.818 & 0.800 & 1.88 & 1.02 & 0.86 & 728 \\
 & \texttt{laplace\_grid}                & 150 & 0.107 & 0.256 & 0.247 & 4.63 & 4.35 & 0.27 & 3385 \\
 & \texttt{linear\_sys\_missing\_coeff}  & 150 & 0.667 & 0.629 & 0.553 & 2.62 & 1.91 & 0.71 & 1327 \\
 & \texttt{markov\_missing\_transition}  & 150 & 0.873 & 0.867 & 0.740 & 1.29 & 0.29 & 1.00 & 930 \\
 & \texttt{poly\_interpolation}          & 150 & 0.773 & 0.775 & 0.727 & 1.81 & 0.97 & 0.84 & 783 \\
 & \texttt{portfolio\_var\_missing\_c.}  & 150 & 0.980 & 1.000 & 1.000 & 1.00 & 0.00 & 1.00 & 571 \\
 & \texttt{steady\_state\_missing\_em.}  & 150 & 0.833 & 0.969 & 0.940 & 1.07 & 0.07 & 1.00 & 848 \\
\bottomrule
\end{tabular}
\end{adjustbox}
\end{table}


\begin{table}[p]
\centering
\caption{Four-shot effect: accuracy change per family (\texttt{Gemma-4}, 20/50 typed dataset). $\Delta$ = four-shot accuracy $-$ baseline accuracy. Positive values indicate improvement.}
\label{tab:fourshot-effect-gemma4}
\begin{adjustbox}{max width=\textwidth}
\begin{tabular}{llr ccc ccc}
\toprule
& & & \multicolumn{3}{c}{\textbf{Accuracy}} & \multicolumn{3}{c}{\textbf{1st-Request Success}} \\
\cmidrule(lr){4-6} \cmidrule(lr){7-9}
\textbf{Type} & \textbf{Family} & $n$ & \textbf{Base} & \textbf{4S} & $\Delta$ & \textbf{Base} & \textbf{4S} & $\Delta$ \\
\midrule
\multirow{11}{*}{A}
 & \texttt{bayes\_missing\_prior}       &  60 & 1.000 & 0.983 & $-$0.017 & 1.000 & 1.000 &  0.000 \\
 & \texttt{crt\_reconstruction}         &  60 & 0.000 & 0.000 &  0.000 & 1.000 & 1.000 &  0.000 \\
 & \texttt{geometry\_coordinates}       &  60 & 0.200 & 0.133 & $-$0.067 & 0.083 & 0.117 & +0.033 \\
 & \texttt{graph\_path\_sums}           &  60 & 0.000 & 0.000 &  0.000 & 0.000 & 0.000 &  0.000 \\
 & \texttt{linear\_system\_separator}   &  60 & 1.000 & 0.950 & $-$0.050 & 0.067 & 0.367 & \textbf{+0.300} \\
 & \texttt{matrix\_completion}          &  60 & 0.950 & 0.867 & $-$0.083 & 0.883 & 0.583 & $-$0.300 \\
 & \texttt{moment\_problem}             &  60 & 0.017 & 0.033 & +0.017 & 0.200 & 0.000 & $-$0.200 \\
 & \texttt{phase\_retrieval}            &  60 & 0.850 & 0.983 & \textbf{+0.133} & 1.000 & 0.983 & $-$0.017 \\
 & \texttt{piecewise\_missing\_thresh.} &  60 & 1.000 & 1.000 &  0.000 & 1.000 & 1.000 &  0.000 \\
 & \texttt{rankdef\_linear\_shared}     &  60 & 0.883 & 0.817 & $-$0.067 & 0.350 & 0.000 & $-$0.350 \\
 & \texttt{recurrence\_missing\_init}   &  60 & 1.000 & 0.983 & $-$0.017 & 1.000 & 1.000 &  0.000 \\
\cmidrule(lr){2-9}
 & \textit{Type A average}              & 660 & 0.627 & 0.614 & $-$0.014 & 0.598 & 0.550 & $-$0.048 \\
\midrule
\multirow{11}{*}{B}
 & \texttt{birth\_death\_missing\_rate}  & 150 & 0.980 & 0.920 & $-$0.060 & 0.947 & 0.907 & $-$0.040 \\
 & \texttt{circuit\_missing\_resistance} & 150 & 1.000 & 0.987 & $-$0.013 & 1.000 & 1.000 &  0.000 \\
 & \texttt{deconvolution}               & 150 & 0.413 & 0.320 & $-$0.093 & 0.233 & 0.060 & $-$0.173 \\
 & \texttt{discrete\_tomography}        & 150 & 0.947 & 0.927 & $-$0.020 & 0.620 & 0.580 & $-$0.040 \\
 & \texttt{eigenvector\_missing\_entry}  & 150 & 0.887 & 0.847 & $-$0.040 & 0.853 & 0.800 & $-$0.053 \\
 & \texttt{laplace\_grid}               & 150 & 0.147 & 0.107 & $-$0.040 & 0.320 & 0.247 & $-$0.073 \\
 & \texttt{linear\_sys\_missing\_coeff}  & 150 & 0.773 & 0.667 & $-$0.107 & 0.647 & 0.553 & $-$0.093 \\
 & \texttt{markov\_missing\_transition}  & 150 & 0.893 & 0.873 & $-$0.020 & 0.607 & 0.740 & \textbf{+0.133} \\
 & \texttt{poly\_interpolation}         & 150 & 0.700 & 0.773 & \textbf{+0.073} & 0.200 & 0.727 & \textbf{+0.527} \\
 & \texttt{portfolio\_var\_missing\_c.}  & 150 & 0.960 & 0.980 & +0.020 & 1.000 & 1.000 &  0.000 \\
 & \texttt{steady\_state\_missing\_em.}  & 150 & 0.913 & 0.833 & $-$0.080 & 0.973 & 0.940 & $-$0.033 \\
\cmidrule(lr){2-9}
 & \textit{Type B average}             & 1650 & 0.783 & 0.748 & $-$0.035 & 0.673 & 0.687 & +0.014 \\
\midrule
 & \textbf{Overall}                    & 2310 & \textbf{0.739} & \textbf{0.710} & $-$0.029 & 0.652 & 0.648 & $-$0.004 \\
\bottomrule
\end{tabular}
\end{adjustbox}
\end{table}
 


\begin{table}[t]
\centering
\caption{Per-family accuracy on the 20/50 typed dataset (\texttt{Grok-4.20}, 2310 instances). Families sorted by Type then name. $n$ = instances per family.}
\label{tab:per-family-acc-grok}
\begin{adjustbox}{max width=\textwidth}
\begin{tabular}{llr cc cc}
\toprule
& & & \multicolumn{2}{c}{\textbf{zero-shot}} & \multicolumn{2}{c}{\textbf{four-shot}} \\
\cmidrule(lr){4-5} \cmidrule(lr){6-7}
\textbf{Type} & \textbf{Family} & $n$ & \textbf{Acc} & \textbf{1st-Req} & \textbf{Acc} & \textbf{1st-Req} \\
\midrule
\multirow{11}{*}{A}
 & \texttt{bayes\_missing\_prior}         &  60 & 0.900 & 1.000 & 0.967 & 1.000 \\
 & \texttt{crt\_reconstruction}           &  60 & 0.000 & 1.000 & 0.017 & 1.000 \\
 & \texttt{geometry\_coordinates}         &  60 & 0.217 & 0.133 & 0.150 & 0.217 \\
 & \texttt{graph\_path\_sums}             &  60 & 0.000 & 0.117 & 0.017 & 0.083 \\
 & \texttt{linear\_system\_separator}     &  60 & 0.000 & 0.317 & 0.050 & 0.983 \\
 & \texttt{matrix\_completion}            &  60 & 0.283 & 0.933 & 0.317 & 0.917 \\
 & \texttt{moment\_problem}               &  60 & 0.000 & 0.000 & 0.000 & 0.017 \\
 & \texttt{phase\_retrieval}              &  60 & 0.567 & 1.000 & 0.467 & 1.000 \\
 & \texttt{piecewise\_missing\_thresh.}   &  60 & 1.000 & 1.000 & 0.717 & 1.000 \\
 & \texttt{rankdef\_linear\_shared}       &  60 & 0.233 & 0.000 & 0.167 & 0.000 \\
 & \texttt{recurrence\_missing\_init}     &  60 & 0.800 & 1.000 & 0.850 & 1.000 \\
\midrule
\multirow{11}{*}{B}
 & \texttt{birth\_death\_missing\_rate}   & 150 & 0.893 & 0.660 & 0.833 & 0.373 \\
 & \texttt{circuit\_missing\_resistance}  & 150 & 0.993 & 1.000 & 0.960 & 1.000 \\
 & \texttt{deconvolution}                 & 150 & 0.047 & 0.093 & 0.020 & 0.093 \\
 & \texttt{discrete\_tomography}          & 150 & 0.107 & 0.167 & 0.073 & 0.020 \\
 & \texttt{eigenvector\_missing\_entry}   & 150 & 0.527 & 0.787 & 0.640 & 0.640 \\
 & \texttt{laplace\_grid}                 & 150 & 0.080 & 0.213 & 0.120 & 0.267 \\
 & \texttt{linear\_sys\_missing\_coeff}   & 150 & 0.653 & 0.293 & 0.633 & 0.233 \\
 & \texttt{markov\_missing\_transition}   & 150 & 0.807 & 0.527 & 0.707 & 0.573 \\
 & \texttt{poly\_interpolation}           & 150 & 0.420 & 0.713 & 0.327 & 0.500 \\
 & \texttt{portfolio\_var\_missing\_c.}   & 150 & 0.920 & 0.993 & 0.920 & 0.987 \\
 & \texttt{steady\_state\_missing\_em.}   & 150 & 0.693 & 1.000 & 0.713 & 0.980 \\
\bottomrule
\end{tabular}
\end{adjustbox}
\end{table}


\begin{table}[p]
\centering
\caption{Full per-family metrics (\texttt{Grok-4.20}, zero-shot, 20/50 typed dataset).}
\label{tab:per-family-full-grok}
\begin{adjustbox}{max width=\textwidth}
\begin{tabular}{llr ccccccc}
\toprule
\textbf{Type} & \textbf{Family} & $n$ & \textbf{Acc} & \textbf{Hit Rate} & \textbf{1st-Req} & \textbf{Avg Req} & \textbf{Avg Decl} & \textbf{Avg Hints} & \textbf{Avg Tok} \\
\midrule
\multirow{11}{*}{A}
 & \texttt{bayes\_missing\_prior}         &  60 & 0.900 & 1.000 & 1.000 & 1.00 & 0.00 & 1.00 & 267 \\
 & \texttt{crt\_reconstruction}           &  60 & 0.000 & 1.000 & 1.000 & 1.00 & 0.00 & 1.00 & 401 \\
 & \texttt{geometry\_coordinates}         &  60 & 0.217 & 0.467 & 0.133 & 2.67 & 1.87 & 0.80 & 5595 \\
 & \texttt{graph\_path\_sums}             &  60 & 0.000 & 0.119 & 0.117 & 5.18 & 5.05 & 0.13 & 2802 \\
 & \texttt{linear\_system\_separator}     &  60 & 0.000 & 0.553 & 0.317 & 2.32 & 1.32 & 1.00 & 636 \\
 & \texttt{matrix\_completion}            &  60 & 0.283 & 0.938 & 0.933 & 1.25 & 0.30 & 0.95 & 330 \\
 & \texttt{moment\_problem}               &  60 & 0.000 & 0.000 & 0.000 & 6.00 & 6.00 & 0.00 & 1791 \\
 & \texttt{phase\_retrieval}              &  60 & 0.567 & 1.000 & 1.000 & 1.00 & 0.00 & 1.00 & 1563 \\
 & \texttt{piecewise\_missing\_thresh.}   &  60 & 1.000 & 1.000 & 1.000 & 1.00 & 0.00 & 1.00 & 288 \\
 & \texttt{rankdef\_linear\_shared}       &  60 & 0.233 & 0.419 & 0.000 & 2.90 & 2.00 & 0.90 & 961 \\
 & \texttt{recurrence\_missing\_init}     &  60 & 0.800 & 1.000 & 1.000 & 1.00 & 0.00 & 1.00 & 325 \\
\midrule
\multirow{11}{*}{B}
 & \texttt{birth\_death\_missing\_rate}   & 150 & 0.893 & 0.765 & 0.660 & 1.74 & 0.79 & 0.95 & 1304 \\
 & \texttt{circuit\_missing\_resistance}  & 150 & 0.993 & 1.000 & 1.000 & 1.00 & 0.00 & 1.00 & 478 \\
 & \texttt{deconvolution}                 & 150 & 0.047 & 0.104 & 0.093 & 5.35 & 5.23 & 0.13 & 3273 \\
 & \texttt{discrete\_tomography}          & 150 & 0.107 & 0.220 & 0.167 & 2.61 & 2.31 & 0.30 & 697 \\
 & \texttt{eigenvector\_missing\_entry}   & 150 & 0.527 & 0.810 & 0.787 & 2.04 & 1.18 & 0.86 & 1602 \\
 & \texttt{laplace\_grid}                 & 150 & 0.080 & 0.325 & 0.213 & 4.20 & 3.71 & 0.49 & 3024 \\
 & \texttt{linear\_sys\_missing\_coeff}   & 150 & 0.653 & 0.429 & 0.293 & 3.56 & 2.88 & 0.68 & 2460 \\
 & \texttt{markov\_missing\_transition}   & 150 & 0.807 & 0.727 & 0.527 & 1.82 & 0.87 & 0.95 & 2088 \\
 & \texttt{poly\_interpolation}           & 150 & 0.420 & 0.807 & 0.713 & 1.60 & 0.64 & 0.96 & 892 \\
 & \texttt{portfolio\_var\_missing\_c.}   & 150 & 0.920 & 0.995 & 0.993 & 1.03 & 0.03 & 1.00 & 594 \\
 & \texttt{steady\_state\_missing\_em.}   & 150 & 0.693 & 1.000 & 1.000 & 1.00 & 0.00 & 1.00 & 1262 \\
\bottomrule
\end{tabular}
\end{adjustbox}
\end{table}


\begin{table}[p]
\centering
\caption{Full per-family metrics (\texttt{Grok-4.20}, four-shot, 20/50 typed dataset).}
\label{tab:per-family-full-grok-4s}
\begin{adjustbox}{max width=\textwidth}
\begin{tabular}{llr ccccccc}
\toprule
\textbf{Type} & \textbf{Family} & $n$ & \textbf{Acc} & \textbf{Hit Rate} & \textbf{1st-Req} & \textbf{Avg Req} & \textbf{Avg Decl} & \textbf{Avg Hints} & \textbf{Avg Tok} \\
\midrule
\multirow{11}{*}{A}
 & \texttt{bayes\_missing\_prior}         &  60 & 0.967 & 1.000 & 1.000 & 1.00 & 0.00 & 1.00 & 266 \\
 & \texttt{crt\_reconstruction}           &  60 & 0.017 & 1.000 & 1.000 & 1.00 & 0.00 & 1.00 & 433 \\
 & \texttt{geometry\_coordinates}         &  60 & 0.150 & 0.567 & 0.217 & 2.27 & 1.35 & 0.92 & 1563 \\
 & \texttt{graph\_path\_sums}             &  60 & 0.017 & 0.103 & 0.083 & 5.32 & 5.18 & 0.13 & 3235 \\
 & \texttt{linear\_system\_separator}     &  60 & 0.050 & 0.989 & 0.983 & 1.03 & 0.03 & 1.00 & 475 \\
 & \texttt{matrix\_completion}            &  60 & 0.317 & 0.929 & 0.917 & 1.37 & 0.42 & 0.95 & 485 \\
 & \texttt{moment\_problem}               &  60 & 0.000 & 0.028 & 0.017 & 5.87 & 5.82 & 0.05 & 2008 \\
 & \texttt{phase\_retrieval}              &  60 & 0.467 & 1.000 & 1.000 & 1.00 & 0.00 & 1.00 & 1039 \\
 & \texttt{piecewise\_missing\_thresh.}   &  60 & 0.717 & 1.000 & 1.000 & 1.00 & 0.00 & 1.00 & 305 \\
 & \texttt{rankdef\_linear\_shared}       &  60 & 0.167 & 0.128 & 0.000 & 4.43 & 3.83 & 0.60 & 945 \\
 & \texttt{recurrence\_missing\_init}     &  60 & 0.850 & 1.000 & 1.000 & 1.00 & 0.00 & 1.00 & 383 \\
\midrule
\multirow{11}{*}{B}
 & \texttt{birth\_death\_missing\_rate}   & 150 & 0.833 & 0.583 & 0.373 & 2.32 & 1.43 & 0.89 & 2332 \\
 & \texttt{circuit\_missing\_resistance}  & 150 & 0.960 & 1.000 & 1.000 & 1.00 & 0.00 & 1.00 & 481 \\
 & \texttt{deconvolution}                 & 150 & 0.020 & 0.093 & 0.093 & 5.51 & 5.42 & 0.09 & 1937 \\
 & \texttt{discrete\_tomography}          & 150 & 0.073 & 0.081 & 0.020 & 2.90 & 2.73 & 0.17 & 333 \\
 & \texttt{eigenvector\_missing\_entry}   & 150 & 0.640 & 0.711 & 0.640 & 2.56 & 1.73 & 0.83 & 2335 \\
 & \texttt{laplace\_grid}                 & 150 & 0.120 & 0.373 & 0.267 & 4.21 & 3.69 & 0.52 & 3747 \\
 & \texttt{linear\_sys\_missing\_coeff}   & 150 & 0.633 & 0.403 & 0.233 & 3.53 & 2.85 & 0.68 & 2720 \\
 & \texttt{markov\_missing\_transition}   & 150 & 0.707 & 0.761 & 0.573 & 1.66 & 0.69 & 0.97 & 1923 \\
 & \texttt{poly\_interpolation}           & 150 & 0.327 & 0.675 & 0.500 & 1.95 & 1.02 & 0.93 & 1027 \\
 & \texttt{portfolio\_var\_missing\_c.}   & 150 & 0.920 & 0.990 & 0.987 & 1.04 & 0.04 & 1.00 & 594 \\
 & \texttt{steady\_state\_missing\_em.}   & 150 & 0.713 & 0.990 & 0.980 & 1.02 & 0.02 & 1.00 & 1424 \\
\bottomrule
\end{tabular}
\end{adjustbox}
\end{table}


\begin{table}[p]
\centering
\caption{Four-shot effect: accuracy change per family (\texttt{Grok-4.20}, 20/50 typed dataset). $\Delta$ = four-shot accuracy $-$ baseline accuracy. Positive values indicate improvement.}
\label{tab:fourshot-effect-grok}
\begin{adjustbox}{max width=\textwidth}
\begin{tabular}{llr ccc ccc}
\toprule
& & & \multicolumn{3}{c}{\textbf{Accuracy}} & \multicolumn{3}{c}{\textbf{1st-Request Success}} \\
\cmidrule(lr){4-6} \cmidrule(lr){7-9}
\textbf{Type} & \textbf{Family} & $n$ & \textbf{Base} & \textbf{4S} & $\Delta$ & \textbf{Base} & \textbf{4S} & $\Delta$ \\
\midrule
\multirow{11}{*}{A}
 & \texttt{bayes\_missing\_prior}         &  60 & 0.900 & 0.967 & +0.067 & 1.000 & 1.000 &  0.000 \\
 & \texttt{crt\_reconstruction}           &  60 & 0.000 & 0.017 & +0.017 & 1.000 & 1.000 &  0.000 \\
 & \texttt{geometry\_coordinates}         &  60 & 0.217 & 0.150 & $-$0.067 & 0.133 & 0.217 & +0.083 \\
 & \texttt{graph\_path\_sums}             &  60 & 0.000 & 0.017 & +0.017 & 0.117 & 0.083 & $-$0.033 \\
 & \texttt{linear\_system\_separator}     &  60 & 0.000 & 0.050 & +0.050 & 0.317 & 0.983 & \textbf{+0.667} \\
 & \texttt{matrix\_completion}            &  60 & 0.283 & 0.317 & +0.033 & 0.933 & 0.917 & $-$0.017 \\
 & \texttt{moment\_problem}               &  60 & 0.000 & 0.000 &  0.000 & 0.000 & 0.017 & +0.017 \\
 & \texttt{phase\_retrieval}              &  60 & 0.567 & 0.467 & $-$0.100 & 1.000 & 1.000 &  0.000 \\
 & \texttt{piecewise\_missing\_thresh.}   &  60 & 1.000 & 0.717 & \textbf{$-$0.283} & 1.000 & 1.000 &  0.000 \\
 & \texttt{rankdef\_linear\_shared}       &  60 & 0.233 & 0.167 & $-$0.067 & 0.000 & 0.000 &  0.000 \\
 & \texttt{recurrence\_missing\_init}     &  60 & 0.800 & 0.850 & +0.050 & 1.000 & 1.000 &  0.000 \\
\cmidrule(lr){2-9}
 & \textit{Type A average}              &  660 & 0.364 & 0.338 & $-$0.026 & 0.591 & 0.656 & +0.065 \\
\midrule
\multirow{11}{*}{B}
 & \texttt{birth\_death\_missing\_rate}   & 150 & 0.893 & 0.833 & $-$0.060 & 0.660 & 0.373 & \textbf{$-$0.287} \\
 & \texttt{circuit\_missing\_resistance}  & 150 & 0.993 & 0.960 & $-$0.033 & 1.000 & 1.000 &  0.000 \\
 & \texttt{deconvolution}                 & 150 & 0.047 & 0.020 & $-$0.027 & 0.093 & 0.093 &  0.000 \\
 & \texttt{discrete\_tomography}          & 150 & 0.107 & 0.073 & $-$0.033 & 0.167 & 0.020 & \textbf{$-$0.147} \\
 & \texttt{eigenvector\_missing\_entry}   & 150 & 0.527 & 0.640 & \textbf{+0.113} & 0.787 & 0.640 & \textbf{$-$0.147} \\
 & \texttt{laplace\_grid}                 & 150 & 0.080 & 0.120 & +0.040 & 0.213 & 0.267 & +0.053 \\
 & \texttt{linear\_sys\_missing\_coeff}   & 150 & 0.653 & 0.633 & $-$0.020 & 0.293 & 0.233 & $-$0.060 \\
 & \texttt{markov\_missing\_transition}   & 150 & 0.807 & 0.707 & $-$0.100 & 0.527 & 0.573 & +0.047 \\
 & \texttt{poly\_interpolation}           & 150 & 0.420 & 0.327 & $-$0.093 & 0.713 & 0.500 & \textbf{$-$0.213} \\
 & \texttt{portfolio\_var\_missing\_c.}   & 150 & 0.920 & 0.920 &  0.000 & 0.993 & 0.987 & $-$0.007 \\
 & \texttt{steady\_state\_missing\_em.}   & 150 & 0.693 & 0.713 & +0.020 & 1.000 & 0.980 & $-$0.020 \\
\cmidrule(lr){2-9}
 & \textit{Type B average}               & 1650 & 0.558 & 0.541 & $-$0.018 & 0.586 & 0.515 & $-$0.071 \\
\midrule
 & \textbf{Overall}                    & 2310 & \textbf{0.503} & \textbf{0.483} & $-$0.020 & 0.587 & 0.555 & $-$0.032 \\
\bottomrule
\end{tabular}
\end{adjustbox}
\end{table}
 


\begin{table}[t]
\centering
\caption{Per-family accuracy on the 20/50 typed dataset (\texttt{Llama-3.1-8B-Instruct}, 2310 instances). Families sorted by Type then name. $n$ = instances per family.}
\label{tab:per-family-acc-llama31}
\begin{adjustbox}{max width=\textwidth}
\begin{tabular}{llr cc cc}
\toprule
& & & \multicolumn{2}{c}{\textbf{zero-shot}} & \multicolumn{2}{c}{\textbf{four-shot}} \\
\cmidrule(lr){4-5} \cmidrule(lr){6-7}
\textbf{Type} & \textbf{Family} & $n$ & \textbf{Acc} & \textbf{1st-Req} & \textbf{Acc} & \textbf{1st-Req} \\
\midrule
\multirow{11}{*}{A}
 & \texttt{bayes\_missing\_prior}       &  60 & 0.017 & 1.000 & 0.033 & 1.000 \\
 & \texttt{crt\_reconstruction}         &  60 & 0.000 & 0.983 & 0.000 & 1.000 \\
 & \texttt{geometry\_coordinates}       &  60 & 0.133 & 0.017 & 0.133 & 0.083 \\
 & \texttt{graph\_path\_sums}           &  60 & 0.000 & 0.000 & 0.000 & 0.000 \\
 & \texttt{linear\_system\_separator}   &  60 & 0.050 & 0.000 & 0.017 & 0.967 \\
 & \texttt{matrix\_completion}          &  60 & 0.183 & 0.000 & 0.000 & 0.000 \\
 & \texttt{moment\_problem}             &  60 & 0.000 & 0.000 & 0.000 & 0.000 \\
 & \texttt{phase\_retrieval}            &  60 & 0.000 & 0.000 & 0.067 & 1.000 \\
 & \texttt{piecewise\_missing\_thresh.} &  60 & 0.450 & 1.000 & 0.600 & 1.000 \\
 & \texttt{rankdef\_linear\_shared}     &  60 & 0.000 & 0.000 & 0.000 & 0.000 \\
 & \texttt{recurrence\_missing\_init}   &  60 & 0.017 & 1.000 & 0.017 & 0.967 \\
\midrule
\multirow{11}{*}{B}
 & \texttt{birth\_death\_missing\_rate}  & 150 & 0.000 & 0.200 & 0.007 & 0.140 \\
 & \texttt{circuit\_missing\_resistance} & 150 & 0.007 & 1.000 & 0.027 & 0.460 \\
 & \texttt{deconvolution}               & 150 & 0.000 & 0.000 & 0.000 & 0.000 \\
 & \texttt{discrete\_tomography}        & 150 & 0.093 & 0.000 & 0.180 & 0.113 \\
 & \texttt{eigenvector\_missing\_entry}  & 150 & 0.053 & 0.193 & 0.053 & 0.280 \\
 & \texttt{laplace\_grid}               & 150 & 0.000 & 0.040 & 0.000 & 0.013 \\
 & \texttt{linear\_sys\_missing\_coeff}  & 150 & 0.000 & 0.000 & 0.007 & 0.113 \\
 & \texttt{markov\_missing\_transition}  & 150 & 0.000 & 0.060 & 0.013 & 0.053 \\
 & \texttt{poly\_interpolation}         & 150 & 0.000 & 0.307 & 0.013 & 0.253 \\
 & \texttt{portfolio\_var\_missing\_c.}  & 150 & 0.007 & 0.733 & 0.027 & 0.627 \\
 & \texttt{steady\_state\_missing\_em.}  & 150 & 0.007 & 0.933 & 0.000 & 0.880 \\
\bottomrule
\end{tabular}
\end{adjustbox}
\end{table}


\begin{table}[p]
\centering
\caption{Full per-family metrics (\texttt{Llama-3.1-8B-Instruct}, zero-shot, 20/50 typed dataset).}
\label{tab:per-family-full-llama31}
\begin{adjustbox}{max width=\textwidth}
\begin{tabular}{llr ccccccc}
\toprule
\textbf{Type} & \textbf{Family} & $n$ & \textbf{Acc} & \textbf{Hit Rate} & \textbf{1st-Req} & \textbf{Avg Req} & \textbf{Avg Decl} & \textbf{Avg Hints} & \textbf{Avg Tok} \\
\midrule
\multirow{11}{*}{A}
 & \texttt{bayes\_missing\_prior}       &  60 & 0.017 & 1.000 & 1.000 & 1.00 & 0.00 & 1.00 & 154 \\
 & \texttt{crt\_reconstruction}         &  60 & 0.000 & 0.983 & 0.983 & 1.03 & 0.05 & 0.98 & 251 \\
 & \texttt{geometry\_coordinates}       &  60 & 0.133 & 0.447 & 0.017 & 2.35 & 1.35 & 1.00 & 209 \\
 & \texttt{graph\_path\_sums}           &  60 & 0.000 & 0.000 & 0.000 & 6.00 & 6.00 & 0.00 & 167 \\
 & \texttt{linear\_system\_separator}   &  60 & 0.050 & 0.406 & 0.000 & 2.57 & 1.57 & 1.00 & 235 \\
 & \texttt{matrix\_completion}          &  60 & 0.183 & 0.500 & 0.000 & 2.00 & 1.00 & 1.00 & 118 \\
 & \texttt{moment\_problem}             &  60 & 0.000 & 0.000 & 0.000 & 6.00 & 6.00 & 0.00 & 167 \\
 & \texttt{phase\_retrieval}            &  60 & 0.000 & 0.138 & 0.000 & 4.23 & 3.57 & 0.67 & 173 \\
 & \texttt{piecewise\_missing\_thresh.} &  60 & 0.450 & 1.000 & 1.000 & 1.00 & 0.00 & 1.00 & 179 \\
 & \texttt{rankdef\_linear\_shared}     &  60 & 0.000 & 0.002 & 0.000 & 6.00 & 5.98 & 0.02 & 169 \\
 & \texttt{recurrence\_missing\_init}   &  60 & 0.017 & 1.000 & 1.000 & 1.00 & 0.00 & 1.00 & 160 \\
\midrule
\multirow{11}{*}{B}
 & \texttt{birth\_death\_missing\_rate}  & 150 & 0.000 & 0.409 & 0.200 & 3.45 & 2.64 & 0.81 & 190 \\
 & \texttt{circuit\_missing\_resistance} & 150 & 0.007 & 1.000 & 1.000 & 1.00 & 0.00 & 1.00 &  92 \\
 & \texttt{deconvolution}               & 150 & 0.000 & 0.001 & 0.000 & 5.97 & 5.97 & 0.01 & 167 \\
 & \texttt{discrete\_tomography}        & 150 & 0.093 & 0.057 & 0.000 & 2.91 & 2.79 & 0.13 &  96 \\
 & \texttt{eigenvector\_missing\_entry}  & 150 & 0.053 & 0.375 & 0.193 & 3.50 & 2.79 & 0.71 & 184 \\
 & \texttt{laplace\_grid}               & 150 & 0.000 & 0.111 & 0.040 & 5.51 & 5.25 & 0.26 & 191 \\
 & \texttt{linear\_sys\_missing\_coeff}  & 150 & 0.000 & 0.086 & 0.000 & 5.41 & 5.17 & 0.23 & 203 \\
 & \texttt{markov\_missing\_transition}  & 150 & 0.000 & 0.146 & 0.060 & 4.94 & 4.55 & 0.39 & 190 \\
 & \texttt{poly\_interpolation}         & 150 & 0.000 & 0.491 & 0.307 & 2.97 & 2.22 & 0.75 & 156 \\
 & \texttt{portfolio\_var\_missing\_c.}  & 150 & 0.007 & 0.838 & 0.733 & 1.52 & 0.52 & 1.00 & 134 \\
 & \texttt{steady\_state\_missing\_em.}  & 150 & 0.007 & 0.966 & 0.933 & 1.07 & 0.07 & 1.00 & 145 \\
\bottomrule
\end{tabular}
\end{adjustbox}
\end{table}


\begin{table}[p]
\centering
\caption{Full per-family metrics (\texttt{Llama-3.1-8B-Instruct}, four-shot, 20/50 typed dataset).}
\label{tab:per-family-full-llama31-4s}
\begin{adjustbox}{max width=\textwidth}
\begin{tabular}{llr ccccccc}
\toprule
\textbf{Type} & \textbf{Family} & $n$ & \textbf{Acc} & \textbf{Hit Rate} & \textbf{1st-Req} & \textbf{Avg Req} & \textbf{Avg Decl} & \textbf{Avg Hints} & \textbf{Avg Tok} \\
\midrule
\multirow{11}{*}{A}
 & \texttt{bayes\_missing\_prior}       &  60 & 0.033 & 1.000 & 1.000 & 1.00 & 0.00 & 1.00 & 117 \\
 & \texttt{crt\_reconstruction}         &  60 & 0.000 & 1.000 & 1.000 & 1.00 & 0.00 & 1.00 & 192 \\
 & \texttt{geometry\_coordinates}       &  60 & 0.133 & 0.478 & 0.083 & 2.30 & 1.30 & 1.00 & 207 \\
 & \texttt{graph\_path\_sums}           &  60 & 0.000 & 0.000 & 0.000 & 6.00 & 6.00 & 0.00 & 197 \\
 & \texttt{linear\_system\_separator}   &  60 & 0.017 & 0.983 & 0.967 & 1.03 & 0.03 & 1.00 & 195 \\
 & \texttt{matrix\_completion}          &  60 & 0.000 & 0.000 & 0.000 & 6.00 & 6.00 & 0.00 & 167 \\
 & \texttt{moment\_problem}             &  60 & 0.000 & 0.000 & 0.000 & 6.00 & 6.00 & 0.00 & 167 \\
 & \texttt{phase\_retrieval}            &  60 & 0.067 & 1.000 & 1.000 & 1.00 & 0.00 & 1.00 & 175 \\
 & \texttt{piecewise\_missing\_thresh.} &  60 & 0.600 & 1.000 & 1.000 & 1.00 & 0.00 & 1.00 & 182 \\
 & \texttt{rankdef\_linear\_shared}     &  60 & 0.000 & 0.008 & 0.000 & 5.98 & 5.93 & 0.05 & 172 \\
 & \texttt{recurrence\_missing\_init}   &  60 & 0.017 & 0.983 & 0.967 & 1.03 & 0.03 & 1.00 & 172 \\
\midrule
\multirow{11}{*}{B}
 & \texttt{birth\_death\_missing\_rate}  & 150 & 0.007 & 0.337 & 0.140 & 3.83 & 3.05 & 0.79 & 212 \\
 & \texttt{circuit\_missing\_resistance} & 150 & 0.027 & 0.729 & 0.460 & 1.55 & 0.55 & 1.00 & 174 \\
 & \texttt{deconvolution}               & 150 & 0.000 & 0.000 & 0.000 & 6.00 & 6.00 & 0.00 & 166 \\
 & \texttt{discrete\_tomography}        & 150 & 0.180 & 0.183 & 0.113 & 2.70 & 2.41 & 0.29 & 108 \\
 & \texttt{eigenvector\_missing\_entry}  & 150 & 0.053 & 0.448 & 0.280 & 3.31 & 2.56 & 0.75 & 176 \\
 & \texttt{laplace\_grid}               & 150 & 0.000 & 0.101 & 0.013 & 5.33 & 5.06 & 0.27 & 193 \\
 & \texttt{linear\_sys\_missing\_coeff}  & 150 & 0.007 & 0.283 & 0.113 & 4.33 & 3.70 & 0.63 & 200 \\
 & \texttt{markov\_missing\_transition}  & 150 & 0.013 & 0.144 & 0.053 & 5.04 & 4.64 & 0.40 & 195 \\
 & \texttt{poly\_interpolation}         & 150 & 0.013 & 0.439 & 0.253 & 3.39 & 2.67 & 0.71 & 169 \\
 & \texttt{portfolio\_var\_missing\_c.}  & 150 & 0.027 & 0.765 & 0.627 & 1.85 & 0.86 & 0.99 & 167 \\
 & \texttt{steady\_state\_missing\_em.}  & 150 & 0.000 & 0.926 & 0.880 & 1.32 & 0.34 & 0.98 & 127 \\
\bottomrule
\end{tabular}
\end{adjustbox}
\end{table}


\begin{table}[p]
\centering
\caption{Four-shot effect: accuracy change per family (\texttt{Llama-3.1-8B-Instruct}, 20/50 typed dataset). $\Delta$ = four-shot accuracy $-$ baseline accuracy. Positive values indicate improvement.}
\label{tab:fourshot-effect-llama31}
\begin{adjustbox}{max width=\textwidth}
\begin{tabular}{llr ccc ccc}
\toprule
& & & \multicolumn{3}{c}{\textbf{Accuracy}} & \multicolumn{3}{c}{\textbf{1st-Request Success}} \\
\cmidrule(lr){4-6} \cmidrule(lr){7-9}
\textbf{Type} & \textbf{Family} & $n$ & \textbf{Base} & \textbf{4S} & $\Delta$ & \textbf{Base} & \textbf{4S} & $\Delta$ \\
\midrule
\multirow{11}{*}{A}
 & \texttt{bayes\_missing\_prior}       &  60 & 0.017 & 0.033 & +0.017 & 1.000 & 1.000 &  0.000 \\
 & \texttt{crt\_reconstruction}         &  60 & 0.000 & 0.000 &  0.000 & 0.983 & 1.000 & +0.017 \\
 & \texttt{geometry\_coordinates}       &  60 & 0.133 & 0.133 &  0.000 & 0.017 & 0.083 & +0.067 \\
 & \texttt{graph\_path\_sums}           &  60 & 0.000 & 0.000 &  0.000 & 0.000 & 0.000 &  0.000 \\
 & \texttt{linear\_system\_separator}   &  60 & 0.050 & 0.017 & $-$0.033 & 0.000 & 0.967 & \textbf{+0.967} \\
 & \texttt{matrix\_completion}          &  60 & 0.183 & 0.000 & $-$0.183 & 0.000 & 0.000 &  0.000 \\
 & \texttt{moment\_problem}             &  60 & 0.000 & 0.000 &  0.000 & 0.000 & 0.000 &  0.000 \\
 & \texttt{phase\_retrieval}            &  60 & 0.000 & 0.067 & +0.067 & 0.000 & 1.000 & \textbf{+1.000} \\
 & \texttt{piecewise\_missing\_thresh.} &  60 & 0.450 & 0.600 & \textbf{+0.150} & 1.000 & 1.000 &  0.000 \\
 & \texttt{rankdef\_linear\_shared}     &  60 & 0.000 & 0.000 &  0.000 & 0.000 & 0.000 &  0.000 \\
 & \texttt{recurrence\_missing\_init}   &  60 & 0.017 & 0.017 &  0.000 & 1.000 & 0.967 & $-$0.033 \\
\cmidrule(lr){2-9}
 & \textit{Type A average}              & 660 & 0.077 & 0.079 & +0.002 & 0.364 & 0.547 & +0.183 \\
\midrule
\multirow{11}{*}{B}
 & \texttt{birth\_death\_missing\_rate}  & 150 & 0.000 & 0.007 & +0.007 & 0.200 & 0.140 & $-$0.060 \\
 & \texttt{circuit\_missing\_resistance} & 150 & 0.007 & 0.027 & +0.020 & 1.000 & 0.460 & $-$0.540 \\
 & \texttt{deconvolution}               & 150 & 0.000 & 0.000 &  0.000 & 0.000 & 0.000 &  0.000 \\
 & \texttt{discrete\_tomography}        & 150 & 0.093 & 0.180 & \textbf{+0.087} & 0.000 & 0.113 & +0.113 \\
 & \texttt{eigenvector\_missing\_entry}  & 150 & 0.053 & 0.053 &  0.000 & 0.193 & 0.280 & +0.087 \\
 & \texttt{laplace\_grid}               & 150 & 0.000 & 0.000 &  0.000 & 0.040 & 0.013 & $-$0.027 \\
 & \texttt{linear\_sys\_missing\_coeff}  & 150 & 0.000 & 0.007 & +0.007 & 0.000 & 0.113 & +0.113 \\
 & \texttt{markov\_missing\_transition}  & 150 & 0.000 & 0.013 & +0.013 & 0.060 & 0.053 & $-$0.007 \\
 & \texttt{poly\_interpolation}         & 150 & 0.000 & 0.013 & +0.013 & 0.307 & 0.253 & $-$0.053 \\
 & \texttt{portfolio\_var\_missing\_c.}  & 150 & 0.007 & 0.027 & +0.020 & 0.733 & 0.627 & $-$0.107 \\
 & \texttt{steady\_state\_missing\_em.}  & 150 & 0.007 & 0.000 & $-$0.007 & 0.933 & 0.880 & $-$0.053 \\
\cmidrule(lr){2-9}
 & \textit{Type B average}             & 1650 & 0.015 & 0.030 & +0.015 & 0.315 & 0.267 & $-$0.048 \\
\midrule
 & \textbf{Overall}                    & 2310 & \textbf{0.033} & \textbf{0.044} & +0.011 & 0.329 & 0.347 & +0.018 \\
\bottomrule
\end{tabular}
\end{adjustbox}
\end{table}

\clearpage 

\section{Problem Family Generators} 
\label{app:families} 

Each family below defines a parameterised class of two-agent reasoning
problems. A \emph{generator} samples an instance at random given a
difficulty level $d \in \{1,2,3\}$ and a random seed. For each instance,
Agent~A and Agent~B each receive a disjoint subset of the constraints;
neither subset alone determines the answer, but together they do, and
exactly one atomic hint from B to A resolves A's ambiguity.

The 22 families are split into two types:
\begin{itemize}
  \item \textbf{Type~A --- Fixed hint slot (11 families).}
    The missing information is structurally determined by the family:
    it is always the same kind/position regardless of the specific instance.
    Agent~A can predict \emph{what} to request without scanning its constraints.
  \item \textbf{Type~B --- Variable hint slot (11 families).}
    The missing slot varies per instance --- it could be any one of
    multiple possible positions. Agent~A must first identify
    \emph{which} slot is absent before formulating its request.
\end{itemize}


\bigskip
\noindent\rule{\textwidth}{0.8pt}
\begin{center}
\Large\bfseries Type~A --- Fixed Hint Slot
\end{center}
\noindent\rule{\textwidth}{0.4pt}
\bigskip

\subsection*{A1.\quad Bayesian Inference with Missing Prior
  (\texttt{bayes\_missing\_prior})}

\paragraph{Setup.}
Three rational parameters are sampled:
\[
  P(H) = \tfrac{p_n}{p_d},\quad
  P(E \mid H) = \tfrac{s_n}{s_d},\quad
  P(E \mid \neg H) = \tfrac{f_n}{f_d},
\]
with denominators drawn from $\mathrm{Unif}[2,\,3+2d]$ and numerators
chosen so all three lie strictly in $(0,1)$ and $P(E\mid H) \ne P(E\mid
\neg H)$. The posterior is computed via Bayes' rule:
\[
  P(H \mid E) \;=\; \frac{P(E \mid H)\,P(H)}{P(E \mid H)\,P(H)
    \;+\; P(E \mid \neg H)\,(1 - P(H))}.
\]
Instances where the posterior denominator exceeds 200 are rejected.

\paragraph{Information split.}
\begin{itemize}
  \item \textbf{Agent A}: the likelihood $P(E \mid H)$ and the
    false-positive rate $P(E \mid \neg H)$.
  \item \textbf{Agent B}: the prior $P(H)$.
\end{itemize}
B's atomic hint is always the prior $P(H)$.

\paragraph{Answer.} The posterior $P(H \mid E)$ as a reduced fraction.

\paragraph{Difficulty scaling ($d \in \{1,2,3\}$).}
Fraction denominators drawn from $[2,\, 3+2d]$ (max 5, 7, 9), controlling
numerical complexity.

\subsection*{A2.\quad Chinese Remainder Theorem Reconstruction
  (\texttt{crt\_reconstruction})}

\paragraph{Setup.}
Three distinct primes $(m_1, m_2, m_3)$ are drawn uniformly at random
(without replacement) from the pool $\{3, 5, 7, 11, 13, 17, 19\}$.
An integer $x \sim \mathrm{Unif}[0, M)$ is sampled, where
$M = m_1 m_2 m_3$, and the three residues $r_i = x \bmod m_i$ are
computed. By the Chinese Remainder Theorem the solution in $[0,M)$ is unique.

\paragraph{Information split.}
\begin{itemize}
  \item \textbf{Agent A}: congruences $x \equiv r_1 \pmod{m_1}$ and
    $x \equiv r_2 \pmod{m_2}$.
  \item \textbf{Agent B}: congruence $x \equiv r_3 \pmod{m_3}$.
\end{itemize}
B's atomic hint is always the third congruence $(m_3, r_3)$.

\paragraph{Answer.} The integer $x$.

\paragraph{Difficulty scaling.}
This family does not scale with difficulty: the prime pool and problem
structure are identical for all $d$.  Accordingly, in \texttt{mix}
generation it is always run at $d = 1$ regardless of the requested
difficulty.

\subsection*{A3.\quad Coordinate Geometry with Missing Line
  (\texttt{geometry\_coordinates})}

\paragraph{Setup.}
An intersection point $(x, y)$ is drawn uniformly from
$[-(4+d),\,4+d]^2$. Two non-parallel lines through $(x,y)$ are
generated with integer coefficients in $[-(3+d),\,3+d]$:
\[
  a_1 x + b_1 y = c_1, \qquad a_2 x + b_2 y = c_2,
\]
subject to $a_1 b_2 - a_2 b_1 \ne 0$ (non-parallel). A reference point
$Q = (x + \Delta x,\; y + \Delta y)$ is placed at an integer Euclidean
distance from $(x,y)$ using a Pythagorean pair
$(\Delta x, \Delta y) \in \{(3,4),(4,3),(5,12),(12,5)\}$.
The target is $d((x,y), Q) = \sqrt{\Delta x^2 + \Delta y^2}$, guaranteed
to be an integer.

\paragraph{Information split.}
\begin{itemize}
  \item \textbf{Agent A}: the first line equation and the coordinates of $Q$.
  \item \textbf{Agent B}: the second line equation and the coordinates of $Q$.
\end{itemize}
B's atomic hint is always the second line equation.

\paragraph{Answer.} The integer distance $d((x,y), Q)$.

\paragraph{Difficulty scaling ($d \in \{1,2,3\}$).}
\begin{itemize}
  \item Intersection point drawn from $[-(4+d),\,4+d]^2$
    (range $\pm$5, $\pm$6, $\pm$7).
  \item Line coefficients drawn from $[-(3+d),\,3+d]$
    (range $\pm$4, $\pm$5, $\pm$6).
\end{itemize}

\subsection*{A4.\quad Triangle Path Sums
  (\texttt{graph\_path\_sums})}

\paragraph{Setup.}
A triangle graph has three edges with positive integer weights $e_{01}, e_{12}, e_{20} \sim \mathrm{Unif}[1,\,2+d]$. The three path sums (one per pair of edges sharing a vertex) are:
\[
  s_1 = e_{01} + e_{12}, \qquad
  s_2 = e_{12} + e_{20}, \qquad
  s_3 = e_{01} + e_{20}.
\]
Any two path sums leave one degree of freedom, while all three path sums determine the three edge weights uniquely.

\paragraph{Information split.}
\begin{itemize}
  \item \textbf{Agent A}: path sums $s_1$ and $s_2$.
  \item \textbf{Agent B}: path sum $s_3 = e_{01} + e_{20}$.
\end{itemize}
B's atomic hint is always the equation $e_{01} + e_{20} = s_3$.

\paragraph{Answer.} The triple $(e_{01}, e_{12}, e_{20})$.

\paragraph{Difficulty scaling ($d \in \{1,2,3\}$).}
Edge weights drawn from $[1,\, 2+d]$ (max 3, 4, 5).

\subsection*{A5.\quad Linear System with Separator Variable
  (\texttt{linear\_system\_separator})}

\paragraph{Setup.}
Unknowns $(a, b, c)$ are drawn uniformly from $[-(5+d),\,5+d]^3$.
Two linearly independent equations in $(a, b, c)$ are sampled with integer
coefficients in $[-(3+d),\,3+d]$, subject to the constraint that, when $c$
is substituted, the residual $2 \times 2$ system in $(a, b)$ is uniquely
solvable (i.e.\ the $2 \times 2$ submatrix on columns $a, b$ has rank 2).

\paragraph{Information split.}
\begin{itemize}
  \item \textbf{Agent A}: the two linear equations.
  \item \textbf{Agent B}: the exact value of $c$.
\end{itemize}
B's atomic hint is always $c = \text{value}$.

\paragraph{Answer.} The triple $(a, b, c)$.

\paragraph{Difficulty scaling ($d \in \{1,2,3\}$).}
\begin{itemize}
  \item Solution drawn from $[-(5+d),\, 5+d]^3$ (range $\pm$6, $\pm$7, $\pm$8).
  \item Coefficients drawn from $[-(3+d),\, 3+d]$ (range $\pm$4, $\pm$5, $\pm$6).
\end{itemize}

\subsection*{A6.\quad Rank-1 Matrix Completion
  (\texttt{matrix\_completion})}

\paragraph{Setup.}
A $2 \times 2$ rank-1 matrix $M = \mathbf{u}\mathbf{v}^\top$ is
constructed from vectors $\mathbf{u} = (u_0, u_1)^\top$ and
$\mathbf{v} = (v_0, v_1)^\top$ with $u_i, v_i \sim \mathrm{Unif}[1,\,2+d]$:
\[
  M = \begin{pmatrix} u_0 v_0 & u_0 v_1 \\ u_1 v_0 & u_1 v_1 \end{pmatrix}
    = \begin{pmatrix} a & b \\ c & d \end{pmatrix}.
\]
By the rank-1 identity, $m_{11} = m_{01}\,m_{10} / m_{00}$
(valid since $m_{00} = u_0 v_0 \ge 1$).

\paragraph{Information split.}
\begin{itemize}
  \item \textbf{Agent A}: entries $m_{00} = a$ and $m_{01} = b$ (first row).
  \item \textbf{Agent B}: entry $m_{10} = c$ (position $[1,0]$).
\end{itemize}
B's atomic hint is always $m_{10} = c$.

\paragraph{Answer.} The integer $m_{11} = bc/a$.

\paragraph{Difficulty scaling ($d \in \{1,2,3\}$).}
Factor entries $u_i, v_i$ drawn from $[1,\, 2+d]$ (max 3, 4, 5).

\subsection*{A7.\quad Moment Problem on Three-Point Support
  (\texttt{moment\_problem})}

\paragraph{Setup.}
A discrete probability distribution on support $\{0, 1, 2\}$ is specified
by unnormalised weights $(q_0, q_1, q_2)$ summing to a common denominator
$D = 6 + 2d$. Three linear constraints determine the distribution uniquely:
\[
  \text{(sum)}\;\; q_0 + q_1 + q_2 = D, \qquad
  \text{(1st moment)}\;\; q_1 + 2q_2 = m_1, \qquad
  \text{(2nd moment)}\;\; q_1 + 4q_2 = m_2.
\]
Values $q_0, q_1 \sim \mathrm{Unif}[1, D-2]$ are drawn subject to
$q_2 = D - q_0 - q_1 > 0$, and $m_1, m_2$ are derived accordingly.

\paragraph{Information split.}
\begin{itemize}
  \item \textbf{Agent A}: the sum constraint and the first-moment equation
    $q_1 + 2q_2 = m_1$.
  \item \textbf{Agent B}: the sum constraint and the second-moment equation
    $q_1 + 4q_2 = m_2$.
\end{itemize}
Each two-equation subsystem has rank 2 in three unknowns (ill-posed).
B's atomic hint is always the second-moment equation.

\paragraph{Answer.} The triple $(q_0, q_1, q_2)$.

\paragraph{Difficulty scaling ($d \in \{1,2,3\}$).}
Common denominator $D = 6+2d$ (values 8, 10, 12), controlling the
magnitude of the weights and moments.

\subsection*{A8.\quad Phase Retrieval via DFT Magnitudes
  (\texttt{phase\_retrieval})}

\paragraph{Setup.}
A signal $\mathbf{x} = (x_0, x_1, x_2, x_3) \in \mathbb{Z}^4$ is drawn
with $x_0 \ne 0$ and $|x_i| \le 2 + \min(d,2)$. The magnitude-squared
spectrum is computed using closed-form expressions for a
length-4 real DFT:
\begin{align*}
  |\hat{x}_0|^2 &= (x_0+x_1+x_2+x_3)^2, \\
  |\hat{x}_1|^2 &= (x_0-x_2)^2 + (x_3-x_1)^2, \\
  |\hat{x}_2|^2 &= (x_0-x_1+x_2-x_3)^2, \\
  |\hat{x}_3|^2 &= (x_0-x_2)^2 + (x_1-x_3)^2.
\end{align*}
Note that $|\hat{x}_1|^2 = |\hat{x}_3|^2$ in this formulation.
By rejection sampling, instances are kept only when
(i) the magnitude vector alone admits multiple solutions (i.e.\ A is
locally ill-posed), and (ii) the magnitude vector together with
$\mathrm{sign}(x_0)$ admits exactly one solution.

\paragraph{Information split.}
\begin{itemize}
  \item \textbf{Agent A}: the four magnitude-squared values
    $(|\hat{x}_0|^2, |\hat{x}_1|^2, |\hat{x}_2|^2, |\hat{x}_3|^2)$.
  \item \textbf{Agent B}: the sign bit $\mathrm{sign}(x_0) \in \{-1, +1\}$.
\end{itemize}
B's atomic hint is always $\mathrm{sign}(x_0)$.

\paragraph{Answer.} The full signal $(x_0, x_1, x_2, x_3)$.

\paragraph{Difficulty scaling ($d \in \{1,2,3\}$).}
Signal value range $|x_i| \le 2+\min(d,2)$ (bounds 3, 4, 4 ---
caps at $d = 2$, so $d = 3$ is identical to $d = 2$).

\subsection*{A9.\quad Piecewise Linear Function with Missing Threshold
  (\texttt{piecewise\_missing\_threshold})}

\paragraph{Setup.}
A two-branch piecewise linear function is defined:
\[
  f(x) = \begin{cases}
    a_0\,x + b_0 & \text{if } x < t, \\
    a_1\,x + b_1 & \text{if } x \ge t,
  \end{cases}
\]
with slopes $a_i \sim \mathrm{Unif}[-(2+d),\,2+d]$, intercepts
$b_i \sim \mathrm{Unif}[-(4+2d),\,4+2d]$, query point
$x \sim \mathrm{Unif}[-(5+3d),\,5+3d]$, and threshold
$t$ drawn uniformly from a symmetric interval, with $x \ne t$ enforced
and the two branch values at $x$ required to be distinct (so the active
branch is not degenerate).

\paragraph{Information split.}
\begin{itemize}
  \item \textbf{Agent A}: the query point $x$ and both branch formulas,
    but \emph{not} $t$.
  \item \textbf{Agent B}: the query point $x$ and the threshold $t$.
\end{itemize}
Without $t$, A cannot determine which branch is active. B's atomic hint
is always $t$.

\paragraph{Answer.} The integer $f(x)$.

\paragraph{Difficulty scaling ($d \in \{1,2,3\}$).}
\begin{itemize}
  \item Slope range $\pm(2+d)$ (max $\pm$3, $\pm$4, $\pm$5).
  \item Query and threshold range $\pm(5+3d)$ (max $\pm$8, $\pm$11, $\pm$14).
\end{itemize}

\subsection*{A10.\quad Rank-Deficient Linear System, Shared Variable
  (\texttt{rankdef\_linear\_shared})}

\paragraph{Setup.}
A solution $(x, y, z) \in \mathbb{Z}^3$ is drawn uniformly from
$[-(5+d),\,5+d]^3$. Four integer-coefficient equations are generated:
rows 1 and 2 form a rank-2 subsystem; rows 3 and 4 are each independently
sufficient to complete the system to rank 3. Formally, all four rows have
coefficients in $[-(3+d),\,3+d] \setminus \{0\}$ and are generated by
rejection to satisfy:
\[
  \mathrm{rank}(\text{rows }1,2) = 2, \quad
  \mathrm{rank}(\text{rows }1,2,3) = 3, \quad
  \mathrm{rank}(\text{rows }1,2,4) = 3.
\]

\paragraph{Information split.}
\begin{itemize}
  \item \textbf{Agent A}: equations 1 and 2 (rank 2 in 3 unknowns,
    ill-posed).
  \item \textbf{Agent B}: equations 3 and 4 (either one suffices for A).
\end{itemize}
B's atomic hint is always equation 3.

\paragraph{Answer.} The triple $(x, y, z)$.

\paragraph{Difficulty scaling ($d \in \{1,2,3\}$).}
\begin{itemize}
  \item Solution drawn from $[-(5+d),\, 5+d]^3$ (range $\pm$6, $\pm$7, $\pm$8).
  \item Coefficients drawn from $[-(3+d),\, 3+d] \setminus \{0\}$
    (range $\pm$4, $\pm$5, $\pm$6).
\end{itemize}

\subsection*{A11.\quad Recurrence Missing Initial Condition
  (\texttt{recurrence\_missing\_init})}

\paragraph{Setup.}
A second-order linear recurrence over the integers:
\[
  a(n+2) \;=\; r_1\,a(n+1) \;+\; r_2\,a(n), \qquad n \ge 0.
\]
Parameters are sampled as $r_1 \sim \mathrm{Unif}[1,\,3+d]$,
$r_2 \sim \mathrm{Unif}[-2,2]$, and initial conditions
$a(0), a(1) \sim \mathrm{Unif}[-3-d,\,3+d]$.
A target index $n^* \sim \mathrm{Unif}[4,\,7+d]$ is drawn, and the target
value $a(n^*)$ is computed by forward iteration.

\paragraph{Information split.}
\begin{itemize}
  \item \textbf{Agent A}: the recurrence coefficients $r_1, r_2$ and the
    initial condition $a(0)$.
  \item \textbf{Agent B}: the recurrence coefficients $r_1, r_2$ and the
    initial condition $a(1)$.
\end{itemize}
Neither agent can compute $a(n^*)$ alone. B's atomic hint is always $a(1)$.

\paragraph{Answer.} The integer $a(n^*)$.

\paragraph{Difficulty scaling ($d \in \{1,2,3\}$).}
\begin{itemize}
  \item Recurrence coefficient $r_1$ drawn from $[1,\, 3+d]$ (max 4, 5, 6).
  \item Initial conditions and coefficients drawn from $[-3-d,\, 3+d]$ (range $\pm$4, $\pm$5, $\pm$6).
  \item Target index $n^*$ drawn from $[4,\, 7+d]$ (max 8, 9, 10).
\end{itemize}


\bigskip
\noindent\rule{\textwidth}{0.8pt}
\begin{center}
\Large\bfseries Type~B --- Variable Hint Slot
\end{center}
\noindent\rule{\textwidth}{0.4pt}
\bigskip

\subsection*{B1.\quad Birth-Death Chain with Missing Rate
  (\texttt{birth\_death\_missing\_rate})}

\paragraph{Setup.}
A birth-death Markov chain on states $\{0, 1, \ldots, n-1\}$ with
$n = d + 2$ is defined by birth rates $\lambda_i$ (transitions
$i \to i+1$ for $i = 0, \ldots, n-2$) and death rates $\mu_i$
(transitions $i \to i-1$ for $i = 1, \ldots, n-1$), all drawn from
$\mathrm{Unif}[1,\,3+2d]$.

The unique stationary distribution satisfies detailed balance:
\[
  \pi_i \;=\; \pi_0 \cdot \prod_{k=0}^{i-1} \frac{\lambda_k}{\mu_{k+1}},
  \qquad i = 1, \ldots, n-1,
\]
normalised so $\sum_{i} \pi_i = 1$. One rate---either some $\lambda_i$ or
some $\mu_i$---is chosen uniformly at random to be hidden. A target state
$s^* \sim \mathrm{Unif}[0, n-1]$ is drawn. Instances where $\pi_{s^*}$
has denominator exceeding 300 are rejected.

\paragraph{Information split.}
\begin{itemize}
  \item \textbf{Agent A}: all birth and death rates \emph{except} the
    one missing rate, in natural-language form.
  \item \textbf{Agent B}: the single missing rate (which rate it is varies per instance).
\end{itemize}

\paragraph{Answer.} The stationary probability $\pi_{s^*}$ as a reduced fraction.

\paragraph{Difficulty scaling ($d \in \{1,2,3\}$).}
\begin{itemize}
  \item Number of states $n = d+2$ (3, 4, 5 states).
  \item Rates drawn from $[1,\, 3+2d]$ (max 5, 7, 9).
\end{itemize}

\subsection*{B2.\quad Series-Parallel Resistor Network
  (\texttt{circuit\_missing\_resistance})}

\paragraph{Setup.}
A random binary tree with $n = 2d + 1$ leaves is built by recursive
splitting. Each leaf is a named resistor $R_1, \ldots, R_n$ with a
positive integer resistance drawn from $\mathrm{Unif}[1,\,5+3(d-1)]$.
Each internal node is independently labelled \emph{series} or
\emph{parallel} with equal probability. The equivalent resistance is
evaluated recursively:
\[
  R_{\text{series}} = R_L + R_R, \qquad
  R_{\text{parallel}} = \frac{R_L\,R_R}{R_L + R_R}.
\]
One resistor $R_k$ (chosen uniformly) is withheld. The circuit topology
is described to both agents in natural language.

\paragraph{Information split.}
\begin{itemize}
  \item \textbf{Agent A}: the full circuit topology and all resistance
    values \emph{except} $R_k$ (which resistor varies per instance).
  \item \textbf{Agent B}: the value of $R_k$.
\end{itemize}

\paragraph{Answer.} The equivalent resistance $R_{\text{eq}}$ as a reduced fraction.

\paragraph{Difficulty scaling ($d \in \{1,2,3\}$).}
\begin{itemize}
  \item Number of resistors $n = 2d+1$ (3, 5, 7).
  \item Resistance values drawn from $[1,\, 5+3(d-1)]$ (max 5, 8, 11).
\end{itemize}

\subsection*{B3.\quad Deconvolution with Missing Measurement
  (\texttt{deconvolution})}

\paragraph{Setup.}
A signal $\mathbf{x} \in \mathbb{Z}^n$ and a kernel $\mathbf{h} \in \mathbb{Z}^m$
are sampled with entries in $[-(2+d),\,2+d]$ and $h_0 \ne 0$.
The full output vector $\mathbf{y} = \mathbf{x} * \mathbf{h} \in \mathbb{Z}^{n+m-1}$
is computed via linear convolution:
\[
  y_k = \sum_{i=0}^{m-1} h_i\, x_{k-i}, \qquad k = 0, \ldots, n+m-2.
\]
The $n+m-1$ equations in $\mathbf{x}$ form an overdetermined linear system.
By rejection, an index $k^*$ is found such that removing $y_{k^*}$ makes
the system rank-deficient, while the full system uniquely determines
$\mathbf{x}$.

\paragraph{Information split.}
\begin{itemize}
  \item \textbf{Agent A}: the kernel $\mathbf{h}$ and all output values
    $y_k$ \emph{except} $y_{k^*}$ (which index varies per instance).
  \item \textbf{Agent B}: the missing output value $y_{k^*}$.
\end{itemize}

\paragraph{Answer.} The signal $\mathbf{x} = (x_0, \ldots, x_{n-1})$.

\paragraph{Difficulty scaling ($d \in \{1,2,3\}$).}
\begin{itemize}
  \item Signal length $n = 3+\min(d,2)$ (4, 5, 5 --- caps at $d=2$).
  \item Kernel length $m = 2$ for $d \le 1$, else $m = 3$.
  \item Entry magnitudes bounded by $2+d$ (max 3, 4, 5).
\end{itemize}

\subsection*{B4.\quad Discrete Tomography with Ambiguous Cell
  (\texttt{discrete\_tomography})}

\paragraph{Setup.}
A $3 \times 3$ binary grid $G \in \{0,1\}^{3 \times 3}$ is sampled
uniformly at random. Its row sums $r_i = \sum_j G_{ij}$ and column sums
$c_j = \sum_i G_{ij}$ are computed. The row and column sums alone are
typically consistent with multiple binary grids. A \emph{target cell}
$(i_t, j_t)$ is chosen from those that are ambiguous (i.e.\ their value
differs across consistent completions). A \emph{hint cell}
$(i_h, j_h) \ne (i_t, j_t)$ is found by search: revealing its value must
reduce the consistent completions to exactly one.

\paragraph{Information split.}
\begin{itemize}
  \item \textbf{Agent A}: all three row sums $r_0, r_1, r_2$ and all
    three column sums $c_0, c_1, c_2$.
  \item \textbf{Agent B}: the value of the hint cell $G_{i_h j_h}$
    (which cell varies per instance).
\end{itemize}

\paragraph{Answer.} The binary value $G_{i_t j_t} \in \{0,1\}$.

\paragraph{Difficulty scaling.}
This family does not scale with difficulty: the grid is always $3 \times 3$
and the generation logic is identical for all $d$. Accordingly, in
\texttt{mix} generation it is always run at $d = 1$ regardless of the
requested difficulty.

\subsection*{B5.\quad Matrix Entry Hidden from Eigenvector
  (\texttt{eigenvector\_missing\_entry})}

\paragraph{Setup.}
An $n \times n$ integer matrix $M$ (with $n = d + 1$) is constructed such
that a known integer vector $\mathbf{v}$ is an eigenvector with eigenvalue
$\lambda$: $M\mathbf{v} = \lambda\mathbf{v}$.

The key construction: one component $v_{j^*} = 0$, so the eigenvector
equation for row $i^*$,
\[
  \sum_{c=0}^{n-1} M_{i^* c}\, v_c = \lambda\, v_{i^*},
\]
contains no term involving $M_{i^* j^*}$ (that term vanishes). Therefore
$M_{i^* j^*}$ is unconstrained by the eigenvector equations. A target
column $t^*$ is drawn ensuring $M_{j^* t^*} \ne 0$, so that A cannot
compute $(M^2)_{i^* t^*} = \sum_k M_{i^* k} M_{k t^*}$ without knowing
$M_{i^* j^*}$.

\paragraph{Information split.}
\begin{itemize}
  \item \textbf{Agent A}: all entries of $M$ \emph{except} $M_{i^* j^*}$,
    and all components of $\mathbf{v}$ (which entry is missing varies per instance).
  \item \textbf{Agent B}: the entry $M_{i^* j^*}$.
\end{itemize}

\paragraph{Answer.} The integer $(M^2)_{i^* t^*}$.

\paragraph{Difficulty scaling ($d \in \{1,2,3\}$).}
\begin{itemize}
  \item Matrix dimension $n = d+1$ ($2 \times 2$, $3 \times 3$, $4 \times 4$).
  \item Entry magnitudes bounded by $3+2(d-1)$ (max 3, 5, 7).
  \item Eigenvalue $\lambda$ drawn from $[1,\, 3+2(d-1)]$ (max 3, 5, 7).
\end{itemize}

\subsection*{B6.\quad Discrete Laplace Equation on a Grid
  (\texttt{laplace\_grid})}

\paragraph{Setup.}
An $n \times n$ grid has unknown values $u_{ij}$ on every cell. Interior
cells satisfy the discrete Laplace equation:
\[
  4\,u_{ij} \;=\; u_{i-1,j} + u_{i+1,j} + u_{i,j-1} + u_{i,j+1},
  \qquad (i,j) \text{ interior.}
\]
Boundary values are drawn as integer multiples of a scaling factor
(24 for $n=4$, 224 for $n=5$) chosen to ensure integer interior solutions.
One boundary cell $(i^*, j^*)$ is chosen uniformly at random and withheld;
it is verified that A's system is rank-deficient without this value.

\paragraph{Information split.}
\begin{itemize}
  \item \textbf{Agent A}: all discrete Laplace equations and all boundary
    values \emph{except} $u_{i^* j^*}$ (which cell varies per instance).
  \item \textbf{Agent B}: the missing boundary value $u_{i^* j^*}$.
\end{itemize}

\paragraph{Answer.} The integer value $u_{i_t j_t}$ at a target interior cell.

\paragraph{Difficulty scaling ($d \in \{1,2,3\}$).}
\begin{itemize}
  \item Grid size $n = 4$ for $d \le 1$, $n = 5$ for $d \ge 2$
    ($4\times4$ grid with 4 interior cells; $5\times5$ with 9 interior cells).
  \item Boundary value magnitudes scale linearly with $d$ within each grid size.
\end{itemize}

\subsection*{B7.\quad Linear System with Missing Coefficient
  (\texttt{linear\_system\_missing\_coeff})}

\paragraph{Setup.}
An $n \times n$ system $A\mathbf{x} = \mathbf{b}$ (with $n = d + 1$) is
generated with a random non-singular integer matrix $A$ and right-hand side
$\mathbf{b}$ (entries in $[-8,8]$). The unique solution $\mathbf{x}$ is
computed by Gaussian elimination. One coefficient $A_{i^* j^*}$ (chosen
uniformly) is withheld. Instances where $x_{k^*}$ has denominator
exceeding 300 are rejected.

\paragraph{Information split.}
\begin{itemize}
  \item \textbf{Agent A}: all coefficients $A_{ij}$ \emph{except}
    $A_{i^* j^*}$, and all right-hand-side values $b_i$, in natural language
    (which coefficient is missing varies per instance).
  \item \textbf{Agent B}: the missing coefficient $A_{i^* j^*}$.
\end{itemize}

\paragraph{Answer.} The solution component $x_{k^*}$ as a reduced fraction.

\paragraph{Difficulty scaling ($d \in \{1,2,3\}$).}
\begin{itemize}
  \item System size $n = d+1$ ($2\times2$, $3\times3$, $4\times4$).
  \item Coefficient magnitudes bounded by $4+3(d-1)$ (max 4, 7, 10).
\end{itemize}

\subsection*{B8.\quad Markov Chain with Missing Transition Probability
  (\texttt{markov\_missing\_transition})}

\paragraph{Setup.}
An $n$-state ergodic Markov chain is generated with a random
row-stochastic transition matrix $P$ (each row sampled by drawing positive
integer parts and normalising). One row index $r^*$ and a column partition
are chosen uniformly at random:
\begin{itemize}
  \item A knows $(n-2)$ entries of row $r^*$.
  \item B knows exactly one entry of row $r^*$ at hint column $c^*$.
  \item The remaining entry is derivable from $\sum_j P_{r^*j} = 1$.
\end{itemize}
All other rows are fully known to A. The unique stationary distribution
$\boldsymbol{\pi}$ satisfies $\boldsymbol{\pi}P = \boldsymbol{\pi}$,
$\sum_i \pi_i = 1$.

\paragraph{Information split.}
\begin{itemize}
  \item \textbf{Agent A}: all rows of $P$ completely, except row $r^*$
    where only $n-2$ entries are known (which row and column vary per instance).
  \item \textbf{Agent B}: entry $P_{r^* c^*}$.
\end{itemize}

\paragraph{Answer.} The stationary probability $\pi_{s^*}$ as a reduced fraction.

\paragraph{Difficulty scaling ($d \in \{1,2,3\}$).}
\begin{itemize}
  \item Number of states $n = \min(d+1, 4)$ (2, 3, 4 --- caps at $d=3$).
  \item Transition probability denominators up to $n \cdot (2+d)$, controlling fractional complexity.
\end{itemize}

\subsection*{B9.\quad Polynomial Interpolation with Missing Point
  (\texttt{poly\_interpolation})}

\paragraph{Setup.}
A polynomial $p$ of degree $\delta = 2 + \min(d, 2)$ is generated by
sampling $\delta + 1$ integer coefficients from $[-(3+d),\,3+d]$.
A set of $\delta + 1$ distinct integer evaluation points is drawn, and the
values $p(x_i)$ computed. By Lagrange interpolation, $\delta + 1$ distinct
points uniquely determine a degree-$\delta$ polynomial, so $\delta$ points
leave it underdetermined. A fresh query point $x^*$ (outside the
interpolation set) is sampled and $p(x^*)$ computed as the target.

\paragraph{Information split.}
\begin{itemize}
  \item \textbf{Agent A}: the $\delta$ evaluation pairs
    $(x_0, p(x_0)), \ldots, (x_{\delta-1}, p(x_{\delta-1}))$
    (which point is held back varies per instance).
  \item \textbf{Agent B}: the one remaining pair $(x_\delta, p(x_\delta))$.
\end{itemize}

\paragraph{Answer.} The integer $p(x^*)$.

\paragraph{Difficulty scaling ($d \in \{1,2,3\}$).}
\begin{itemize}
  \item Degree $\delta = 2+\min(d,2)$ (3, 4, 4 --- caps at $d=2$, so $d=3$
    is identical to $d=2$ in structure but with wider coefficient range).
  \item Coefficient magnitudes bounded by $3+d$ (max 4, 5, 6).
\end{itemize}

\subsection*{B10.\quad Portfolio Variance with Missing Correlation
  (\texttt{portfolio\_variance\_missing\_corr})}

\paragraph{Setup.}
A portfolio of $n = d + 1$ assets has weights $w_i$, standard deviations
$\sigma_i \sim \mathrm{Unif}[1, 5]$, and pairwise correlations
$\rho_{ij} \in (-1, 1)$ for $i < j$ (drawn as rational multiples of $1/D$
for $D \in \{2,3,4\}$). Portfolio variance is:
\[
  \sigma_p^2 = \sum_{i=0}^{n-1} \sum_{j=0}^{n-1}
    w_i\, w_j\, \sigma_i\, \sigma_j\, \rho_{ij}, \qquad \rho_{ii} = 1.
\]
One correlation pair $(a, b)$ with $a < b$ is chosen uniformly at random
and withheld. Instances where $\sigma_p^2$ has denominator exceeding 500
are rejected.

\paragraph{Information split.}
\begin{itemize}
  \item \textbf{Agent A}: all weights $w_i$, all standard deviations
    $\sigma_i$, and all $\binom{n}{2}$ pairwise correlations \emph{except}
    $\rho_{ab}$ (which pair varies per instance).
  \item \textbf{Agent B}: the correlation $\rho_{ab}$.
\end{itemize}

\paragraph{Answer.} The portfolio variance $\sigma_p^2$ as a reduced fraction.

\paragraph{Difficulty scaling ($d \in \{1,2,3\}$).}
Number of assets $n = d+1$ (2, 3, 4), giving $\binom{n}{2}$ correlation pairs
(1, 3, 6 --- of which one is always missing).

\subsection*{B11.\quad Hidden Markov Model with Missing Emission
  (\texttt{steady\_state\_missing\_emission})}

\paragraph{Setup.}
A Hidden Markov Model has $n = d + 1$ states and $m = \min(d+1, 3)$
observation symbols. The transition matrix $P$ is fully specified (each row
a random distribution over states), and the emission matrix has entries
$E[s][o]$ = probability of observing symbol $o$ from state $s$ (each row a
random distribution over symbols).

The stationary distribution $\boldsymbol{\pi}$ of $P$ is computed by
solving $\boldsymbol{\pi}P = \boldsymbol{\pi}$. The marginal observation
probability for a target symbol $o^*$ is:
\[
  P(\mathrm{observe}\; o^*) = \sum_{s=0}^{n-1} \pi_s\, E[s][o^*].
\]
One emission entry $E[s^*][o^*]$ (chosen uniformly) is withheld.

\paragraph{Information split.}
\begin{itemize}
  \item \textbf{Agent A}: the full transition matrix $P$ and all emission
    entries \emph{except} $E[s^*][o^*]$ (which state/symbol entry varies per instance).
  \item \textbf{Agent B}: the missing emission $E[s^*][o^*]$.
\end{itemize}

\paragraph{Answer.} The observation probability $P(\mathrm{observe}\; o^*)$
as a reduced fraction.

\paragraph{Difficulty scaling ($d \in \{1,2,3\}$).}
\begin{itemize}
  \item Number of states $n = d+1$ (2, 3, 4).
  \item Number of symbols $m = \min(d+1, 3)$ (2, 3, 3 --- caps at $d=2$).
\end{itemize}

\section{Illustrative Transcript Snapshots} 
\label{app:trace_snapshots}
This appendix gives compact snapshots of representative transcripts. The snapshots are included to make the qualitative failure modes in Section~7 concrete. They are not a sampling protocol and are not used to estimate prevalence. Aggregate conclusions should be drawn from the quantitative tables in the main text and Appendix~A. The full transcripts and raw JSON logs are included with the released artifacts. 

For each snapshot, Agent A is the solver and Agent B is the fixed constrained information holder. We report only the fields needed to interpret the trace: the instance, the canonical missing fact held by Agent B, the salient request/offer/decline sequence, and the exact-verifier outcome. 

We exclude noncanonical offers from the integration-failure examples. If the responder offers a constraint that does not match the canonical atomic hint, the trace is categorized separately as a responder-matching anomaly rather than as a solver integration failure.

\paragraph{T1: No canonical hint acquired.} This trace illustrates a request-acquisition failure. The solver identifies the general family and realizes that a birth--death rate is missing, but repeatedly asks for wrong rate slots. Agent B never issues an offer, so the solver exhausts its request budget without the resolving constraint. 

\begin{verbatim} 
Model/mode: gpt-4o-mini, zero-shot 
Instance: bir-000768 
Family: birth_death_missing_rate 
Target: pi_2 
Canonical missing fact held by B: mu_4 = 2 

A requests: "the birth rate lambda_2" 
B declines. 
A requests: "the death rate mu_0" 
B declines. 
A requests: "the death rate mu_2" 
B declines. 
A requests: "the birth rate lambda_2" 
B declines. 
A requests: "the birth rate lambda_1" 
B declines. 
A requests: "the birth rate lambda_3" 
B declines. 
A requests: "the birth rate lambda_0" 
B declines. 
A requests: "the death rate mu_3" 
B declines. 
A requests: "the birth rate lambda_4" 
B declines. 

Outcome: no answer submitted. 
Ground truth: 1/9. 
Failure type: request-acquisition failure. 
\end{verbatim} 

This example shows why average request count and declination count are informative: the solver is not merely failing the final arithmetic, but failing to localize the hidden slot under the request budget. \paragraph{T2: Canonical hint acquired, but final answer wrong.} This trace illustrates an integration failure. The solver asks for the correct missing prior and receives it, but then submits an incorrect posterior. 

\begin{verbatim} 
Model/mode: gpt-4o-mini, zero-shot 
Instance: bay-000044 
Family: bayes_missing_prior 
Target: P_H_given_E 
A's private facts: P(E|H) = 1/9, P(E|not_H) = 1/2 
Canonical missing fact held by B: P(H) = 5/8 

A requests: "the prior P(H)" 
B offers: P(H) = 5/8 

A final answer: 5/29 
Ground truth: 10/37. 
Failure type: integration failure after successful hint acquisition. 
\end{verbatim} 

The request was successful, so this transcript should not be counted as a gap-identification failure. The incorrect score comes from the downstream computation or simplification step. \paragraph{T3: Canonical hint acquired and final answer correct.} This trace illustrates the intended successful loop. The solver identifies the absent resistor, receives the missing value, and computes the exact equivalent resistance. 

\begin{verbatim} 
Model/mode: gpt-4o-mini, zero-shot 
Instance: cir-000924 
Family: circuit_missing_resistance 
Target: R_eq 
Canonical missing fact held by B: R3 = 7 ohms 

A requests: "the resistance of R3" 
B offers: R3 = 7 ohms 
A computes: 
    G1 = R3 + R7 = 7 + 6 = 13. 
    G2 = R2 + R5 = 11 + 4 = 15. 
    G3 = R1 + G2 = 8 + 15 = 23. 
    G4 = G3 + R6 = 23 + 9 = 32. 
    G5 = G4 + R4 = 32 + 1 = 33. 
    R_eq = (13 * 33) / (13 + 33) = 429/46. 

A final answer: 429/46 
Ground truth: 429/46. 
Failure type: none; full request--resolve--solve success. 
\end{verbatim} 

\paragraph{T4: Optional prompt-anchoring snapshot for four-shot analysis.} This trace is included only to illustrate a prompt-sensitivity hypothesis. It should not be interpreted as a prevalence estimate. In this moment-recovery instance, the solver repeatedly asks for generic named moments rather than asking for the concrete missing equation held by Agent B. 

\begin{verbatim} 
Model/mode: gemini-2.5-flash, four-shot 
Instance: mom-000404 
Family: moment_problem 
Target: q0, q1, q2 
A's private equations: q0 + q1 + q2 = 12 q1 + 2q2 = 3 
Canonical missing fact held by B: q1 + 4q2 = 5 

A requests: "the 2nd moment" 
B declines. 
A requests: "the 1st moment" 
B declines. 
A requests: "the 0th moment" 
B declines. 
A requests: "the value of moment_0" 
B declines. 
A requests: "the value of moment_1" 
B declines. 
A requests: "the value of moment_2" 
B declines. 
A requests: "the 0th moment" 
B declines. 
A requests: "the 2nd moment" 
B declines. 
A requests: "the 3rd moment" 
B declines. 

Outcome: no answer submitted. 
Ground truth: q0 = 10, q1 = 1, q2 = 1. 
Interpretation: possible exemplar anchoring or mismatch between generic 
moment labels and the concrete constraint held by the responder. 
\end{verbatim} 

This example supports the qualitative interpretation in Section~6: four-shot demonstrations may encourage protocol-shaped requests without guaranteeing instance-specific slot localization. It is not used as evidence that this behavior is frequent.

\end{document}


\maketitle

\section{Dataset Summary}
\bench is a synthetic benchmark for minimal information requesting and mathematical reasoning under partial information. Each instance is generated from a complete latent mathematical state with a unique answer. The solver receives a locally underdetermined private view that is missing exactly one necessary atomic fact. To solve the instance, the solver must request the missing fact in natural language under a fixed request budget and then integrate the returned fact into an exact final answer.

The benchmark is designed to separate two capabilities that are conflated by standard full-information mathematical benchmarks: identifying and requesting the missing mathematical information, and performing the downstream computation after receiving it. Instance generation, typed atomic-hint specifications, validation, and final-answer verification are deterministic. Request acceptance in the reported reference protocol is measured under a fixed constrained LLM-mediated responder channel.

\section{Access}
The release artifacts are available at the following locations:
\begin{center}
Code and documentation: \releaseurl\\
Generated benchmark data: \dataurl
\end{center}

The generated benchmark files are hosted as downloadable \jsonl artifacts. The code repository contains the generators, verifiers, prompts, runner, scoring utilities, documentation, and reproduction scripts. The submitted release is identified by the repository commit hash, dataset version, and file checksums recorded with the artifact release. Subsequent updates will use version tags or archived releases so that the version corresponding to this paper remains recoverable.

\section{Released Artifacts}
The release consists of the following artifacts:
\begin{itemize}[leftmargin=*]
  \item Frozen generated benchmark instances in \jsonl format.
  \item Family-specific Python generators and seed lists for regenerating the benchmark.
  \item Instance validators that check global uniqueness, local underdetermination, and sufficiency of the canonical atomic hint.
  \item Exact answer verifiers and normalization utilities.
  \item Machine-readable canonical atomic-hint specifications and typed hint identifiers.
  \item Prompt templates for zero-shot and four-shot evaluation.
  \item Structured-output schemas and prompts for the fixed constrained information-holder responder.
  \item Reference runner and scoring scripts.
  \item Raw interaction transcripts, run metadata, and aggregate result files for the reported baselines.
  \item Documentation, changelog, license files, and checksums.
\end{itemize}

\section{Dataset Composition}
The current release is the \emph{20/50 typed} split. It contains 2{,}310 instances across 22 typed mathematical families. The families are divided into two groups:
\begin{itemize}[leftmargin=*]
  \item \textbf{Type A: fixed missing-information slots.} There are 11 Type A families, with 20 instances per difficulty level for each family, for 660 Type A instances in total.
  \item \textbf{Type B: variable missing-information slots.} There are 11 Type B families, with 50 instances per difficulty level for each family, for 1{,}650 Type B instances in total.
\end{itemize}
Each family is generated at three difficulty levels unless the family definition is intentionally non-scaling. Difficulty can affect coefficient ranges, dimensions, denominators, target-index depth, grid size, or other family-specific parameters.

The 22 families cover Bayesian inference, Chinese remainder reconstruction, coordinate geometry, triangle path-sum inversion, linear systems with a separator variable, rank-one matrix completion, moment recovery, phase retrieval, piecewise functions, rank-deficient linear systems with a shared variable, recurrences, birth-death chains, series-parallel resistor networks, deconvolution, discrete tomography, eigenvector-constrained matrix questions, discrete Laplace equations, linear systems with a missing coefficient, Markov chains, polynomial interpolation, portfolio variance, and hidden Markov models.

\section{Instance Fields and Data Format}
Each generated instance records the information needed for validation, evaluation, and regeneration. The exact field names are defined by the released schema, but conceptually each instance includes:
\begin{itemize}[leftmargin=*]
  \item family name, family type, difficulty level, and generation seed;
  \item solver-facing private view and target question;
  \item responder-facing private constraint containing the missing atomic fact;
  \item machine-readable canonical atomic-hint identifier and value;
  \item exact target answer and answer type;
  \item verifier metadata and normalization information;
  \item prompt-regime metadata where applicable.
\end{itemize}
The frozen \jsonl files are intended for direct evaluation. The generator code and fixed seed lists are provided so that users can regenerate the same release and audit individual instances from their seeds.

\section{Generation Procedure}
Each instance is sampled from a family-specific generator using recorded random seeds. Generators use exact arithmetic and rejection sampling to enforce nondegeneracy, local underdetermination of the solver view, global uniqueness after adding the canonical atomic hint, and verifier-compatible answers. Family-specific checks include rank tests, uniqueness tests, exact rational normalization, finite completion enumeration, denominator thresholds, and structural constraints such as nonsingularity or nonparallelism.

The generation process enforces the following high-level invariant: the solver's private view alone is insufficient to determine the target answer, while the solver's private view plus the dataset-provided atomic hint is sufficient for a unique answer. This construction makes the benchmark a diagnostic of missing-information requesting rather than a collection of ordinary fully specified math problems.

\section{Labels, Hints, and Verifiers}
Each instance contains an exact answer represented as an integer, reduced rational, tuple, vector, or family-specific structured object. Final answers are evaluated by family-specific exact verifiers after normalization rather than by free-form string matching.

The missing information is represented by a typed canonical atomic-hint specification. Examples include a missing prior, congruence, coefficient, boundary value, transition probability, resistor value, interpolation point, or emission entry. The solver may ask for this information in free-form natural language, but successful acquisition requires the request to identify the held atomic fact precisely enough to be accepted by the fixed constrained responder.

\section{Interaction and Evaluation Protocol}
In the reference evaluation protocol, the solver is shown its private view, the target question, and a request budget. At each turn, the solver may request information or submit a final answer. The information holder is a fixed constrained LLM responder. It receives only its private constraint and the solver's current request. It must return a structured \textsc{offer} when it judges the request to match the held fact, or a structured \textsc{decline} otherwise. A valid offer includes \texttt{has\_exact\_match=true}, the quoted private constraint, and the extracted hint string. A decline must not reveal alternative information.

This responder channel is controlled but not deterministic: semantic matching is LLM-mediated. The deterministic components are the family generators, canonical hint specifications, validators, stored answers, and exact final-answer verifiers. Request metrics should therefore be interpreted as solver performance under this fixed constrained responder channel, not as performance under a deterministic symbolic request-matching oracle.

\section{Reported Metrics}
The benchmark supports both final-answer and request-level metrics:
\begin{itemize}[leftmargin=*]
  \item \textbf{Accuracy}: fraction of instances whose final answer is accepted by the exact verifier.
  \item \textbf{Request hit rate}: accepted requests divided by responder messages, averaged per instance.
  \item \textbf{First-request success}: fraction of instances where the first request is accepted.
  \item \textbf{Average requests}: mean number of solver requests per instance.
  \item \textbf{Average declinations}: mean number of fixed-responder declinations per instance.
  \item \textbf{Average hints}: mean number of accepted offers per instance.
  \item \textbf{Trace-level decomposition}: no canonical hint acquired, canonical hint acquired but final answer wrong, and final answer correct.
\end{itemize}
The trace-level decomposition is computed from released transcripts and offline hint audits, without additional model calls.

\section{Quality Controls}
Quality controls include deterministic seeding, exact arithmetic, family-specific nondegeneracy filters, schema validation, instance validators, verifier tests, and release checksums. For finite discrete families, validators can enumerate consistent completions. For algebraic families, validators use rank, uniqueness, or constraint-satisfaction tests where appropriate.

The release includes checksums for the frozen JSONL files and released transcript archives. Future releases will add continuous-integration tests that regenerate benchmark subsets, validate schemas, and run exact verifiers on sampled instances.

\section{Intended Uses}
Appropriate uses include:
\begin{itemize}[leftmargin=*]
  \item evaluating whether a model can recognize a missing mathematical fact;
  \item measuring how precisely a solver can request a necessary atomic hint;
  \item testing whether a model can integrate a returned hint into an exact mathematical answer;
  \item comparing prompts, solver loops, request-budget strategies, and communication-efficiency methods;
  \item auditing the gap between request acquisition and downstream mathematical integration;
  \item regression testing mathematical reasoning systems under controlled partial information.
\end{itemize}

\section{Out-of-Scope Uses}
\bench should not be used as evidence that a model can collaborate with an autonomous teammate. The second role is a fixed constrained information holder, not a strategic collaborator. The benchmark should also not be treated as a general measure of tool use, web navigation, retrieval, social reasoning, human dialogue, or real-world decision-making. Success on \bench should not be used to claim broad agentic ability without additional evidence from settings designed for those capabilities.

\section{Limitations}
The benchmark is synthetic and mathematically structured. This improves control, auditability, and exact verification, but limits realism. Each current instance requires exactly one atomic hint; many real tasks require multiple facts, uncertain evidence, or iterative human clarification. The typed hint schema and constrained responder protocol make the benchmark reproducible but may understate the difficulty of communicating with humans or open-ended systems.

The reported request metrics depend on a fixed LLM-mediated responder channel. Although the responder is constrained and auditable, semantic offer-or-decline decisions can still reflect occasional responder matching errors. Model rankings may also become stale as API-backed models and open-weight checkpoints change. The release is therefore intended to support reproducible reruns and future baselines rather than a permanent leaderboard.

\section{Licensing}
The generated benchmark data are released under the CC BY 4.0 license. The generator, verifier, runner, and evaluation code are released under the MIT License. Users should cite the MIRA-Math paper and dataset when using the benchmark.

\section{Hosting and Maintenance Plan}
The maintainers will preserve the 20/50 typed split as a stable regression suite. Future releases may add larger public splits, private held-out seeds, multi-hint variants, alternative prompt protocols, or alternative responder implementations. Each release will include:
\begin{itemize}[leftmargin=*]
  \item a versioned dataset identifier;
  \item generator version and code commit;
  \item seed lists and prompt-protocol version;
  \item dataset and transcript checksums;
  \item a changelog covering changes to generators, prompts, matching rules, verifiers, and scoring scripts;
  \item documentation describing backward compatibility with the original 20/50 typed split.
\end{itemize}
The Hugging Face dataset page should remain the primary location for downloadable generated \jsonl files, while the code repository should remain the primary location for generation, validation, and evaluation code.

\section{Ethical Considerations}
\bench uses synthetically generated mathematical problem text and synthetic numeric values. It does not contain human-subject data, private user data, or scraped personal information. Privacy and consent risks are therefore minimal. The main ethical risk is overclaiming: benchmark performance should not be presented as evidence of broad collaboration, human communication skill, open-ended tool use, or deployment readiness. The paper and release documentation mitigate this risk by specifying intended uses, out-of-scope uses, and limitations.

\section{Author Responsibility Statement}
The authors are responsible for the benchmark design, generated instances, code, documentation, evaluation protocol, and released metadata. The authors have verified that the release does not violate third-party rights, that released model-output transcripts are shared in accordance with applicable provider terms, and that the selected data and code licenses are stated in the repository and dataset page.